\documentclass[letterpaper]{article} 
\usepackage{aaai2026}  
\usepackage{times}  
\usepackage{helvet}  
\usepackage{courier}  
\usepackage[hyphens]{url}  
\usepackage{graphicx} 
\usepackage{natbib}  
\usepackage{caption} 
\usepackage{multirow} 
\usepackage{algorithmic}
\usepackage[linesnumbered,ruled,vlined]{algorithm2e}
\usepackage{xcolor}

\usepackage{xurl}
\usepackage{newfloat}
\usepackage{listings}
\DeclareCaptionStyle{ruled}{labelfont=normalfont,labelsep=colon,strut=off} 
\usepackage{booktabs}
\usepackage{xspace}
\usepackage{amsfonts} 
\usepackage{subcaption} 
\usepackage{pifont}
\usepackage{appendix}
\usepackage{cleveref}

\newcommand{\smart}{SMART\xspace}
\newcommand{\lsmartpro}{LSMART\xspace}
\newcommand{\noaction}{event-based\xspace}
\newcommand{\Noaction}{Event-based\xspace}
\newcommand{\lrgw}{LRGW\xspace}
\newcommand{\LRGW}{Local Repair Guided Waits\xspace}
\newcommand{\distinctOneGoal}{distinct-one-goal\xspace}
\newcommand{\oneGoal}{one-goal\xspace}
\newcommand{\windowedMultiGoal}{windowed-multi-goals\xspace}
\newcommand{\DistinctOneGoal}{Distinct-One-Goal\xspace}
\newcommand{\OneGoal}{One-Goal\xspace}
\newcommand{\WindowedMultiGoal}{Windowed-Multi-Goals\xspace}
\newcommand{\warehouseSmall}{\texttt{warehouse-33-36}\xspace}
\newcommand{\mazeSmall}{\texttt{maze-32-32-4}\xspace}
\newcommand{\emptySmall}{\texttt{empty-32-32}\xspace}
\newcommand{\randomMid}{\texttt{random-64-64-10}\xspace}
\newcommand{\roomMid}{\texttt{room-64-64-16}\xspace}
\newcommand{\denMid}{\texttt{den312d}\xspace}
\newcommand{\warehouseLarge}{\texttt{warehouse-10-20-10-2-1\xspace}}
\newcommand{\cmark}{\ding{51}}%
\newcommand{\xmark}{\ding{55}}%

\newtheorem{definition}{Definition}
\Crefname{equation}{Eq.}{Eqs.}
\Crefname{figure}{Fig.}{Figs.}
\Crefname{tabular}{Tab.}{Tabs.}

\title{Lifelong Scalable Multi-Agent Realistic Testbed and A Comprehensive Study on Design Choices in Lifelong AGV Fleet Management Systems}
\author{
    Jingtian Yan\textsuperscript{\rm 1}\thanks{These authors contributed equally and are listed alphabetically.},
    Yulun Zhang\textsuperscript{\rm 1}\footnotemark[1],
    Zhenting Liu\textsuperscript{\rm 1},
    Han Zhang\textsuperscript{\rm 2},
    He Jiang\textsuperscript{\rm 1},\\
    Jingkai Chen\textsuperscript{\rm 2},
    Stephen F. Smith\textsuperscript{\rm 1},
    Jiaoyang Li\textsuperscript{\rm 1}
}
\affiliations{
    \textsuperscript{\rm 1}Robotics Institute, Carnegie Mellon University \hspace{1cm}
    \textsuperscript{\rm 2}Symbotic \\
    \{jingtianyan,yulunzhang\}@cmu.edu

}

\begin{document}

\maketitle

\begin{abstract}


We present \textbf{L}ifelong \textbf{S}calable \textbf{M}ulti-\textbf{A}gent \textbf{R}ealistic \textbf{T}estbed (\textbf{\lsmartpro}), an open-source simulator to evaluate any Multi-Agent Path Finding (MAPF) algorithm in a Fleet Management System (FMS) with Automated Guided Vehicles (AGVs).
MAPF aims to move a group of agents from their corresponding starting locations to their goals. Lifelong MAPF (LMAPF) is a variant of MAPF that continuously assigns new goals for agents to reach. 
LMAPF applications, such as autonomous warehouses, often require a centralized, lifelong system to coordinate the movement of a fleet of robots, typically AGVs.
However, existing works on MAPF and LMAPF often assume simplified kinodynamic models, such as pebble motion, as well as perfect execution and communication for AGVs.
Prior work has presented \smart, a software capable of evaluating any MAPF algorithms while considering agent kinodynamics, communication delays, and execution uncertainties.
However, \smart is designed for MAPF, not LMAPF. 
Generalizing \smart to an FMS requires many more design choices. 
First, an FMS parallelizes planning and execution, raising the question of \emph{when to plan}. 
Second, given planners with varying optimality and differing agent-model assumptions, one must decide \emph{how to plan}. 
Third, when the planner fails to return valid solutions, the system must determine \emph{how to recover}.
In this paper, we first present \lsmartpro, an open-source simulator that incorporate all these considerations to evaluate any MAPF algorithms in an FMS.
We then provide experiment results based on state-of-the-art methods for each design choice, offering guidance on how to effectively design centralized lifelong AGV Fleet Management Systems. LSMART is available at \url{https://smart-mapf.github.io/lifelong-smart}.

\end{abstract}

\section{Introduction} \label{sec:intro}

We present \lsmartpro, the first open-source software capable of evaluating any Multi-Agent Path Finding (MAPF) algorithm in a centralized, lifelong Fleet Management System (FMS) with Automated Guided Vehicles (AGVs). MAPF~\cite{SternSoCS19} aims to move agents from their starts to goals without collisions, and lifelong MAPF (LMAPF) is a variant of MAPF that continuously assigns new goals to agents.
LMAPF has extensive applications, such as autonomous fulfillment warehouses~\cite{Li2020LifelongMP} and robotic sorting systems~\cite{zhang2025tmo}, where a centralized, lifelong system is required to coordinate the movement of a fleet of AGVs.
Many prior works have studied LMAPF by planning paths~\cite{Li2020LifelongMP}, optimizing task assignments~\cite{KouAAAI20}, optimizing physical layouts~\cite{zhangLayout23}, or optimizing traffic~\cite{zhang2024ggo}.
Although MAPF and LMAPF research is often motivated by real-world applications, very few works are evaluated in a realistic setting that considers kinodynamic constraints, communication delays, and execution uncertainties.
In fact, most works predominantly assume the pebble motion model for AGVs' movement, where they move in a discretized space (typically a 4-connected grid) and at discretized timesteps. 
While a prior work has proposed \smart~\cite{yan2025smart}, an open-source software tool to evaluate any MAPF algorithm while respecting these factors, \smart is designed for MAPF, instead of LMAPF.

Generalizing \smart to an FMS in LMAPF settings requires non-trivial design choices.
First, we need to decide \emph{when to plan}. In LMAPF, agents are continuously assigned new goals, which requires invoking the planner to plan new paths.
We also need to resynchronize agents’ plans through replanning based on execution feedback to maintain efficiency.
Therefore, an \emph{invocation policy} is needed to decide \emph{when} to invoke the planner.
Second, we must decide \emph{how to plan}. Prior work has proposed MAPF algorithms with different optimality and agent models. Using an algorithm with better optimality and a more accurate agent model improves solution quality, but also increases runtime~\cite{yan2025bridging}. In LMAPF, the MAPF planner often has a time limit. Thus, it is critical to determine \emph{how} to plan. For example, is an optimal planner with a simplified agent model better than a suboptimal planner with an accurate agent model? Moreover, MAPF planners often assume that agents remain at their goals upon arrival and therefore require distinct goals, whereas in LMAPF it is reasonable to assume that the same goal can be assigned to multiple agents. Therefore, an \emph{instance generator} is needed to generate and refine MAPF problem instances that satisfy such assumptions.
Third, we must decide \emph{how to recover from failure}. In practice, MAPF planners may fail to return a solution, either due to algorithmic incompleteness~\cite{MaAAAI19} or time limits despite theoretical completeness~\cite{SharonAIJ15}. Thus, a \emph{fail policy} is required to recover from planning failures~\cite{morag_adapting_2023}.

In this paper, we present \lsmartpro, the first open-source simulation software that allow MAPF researchers to evaluate any MAPF algorithm in an FMS.
\lsmartpro \textbf{encapsulates all} the design choices mentioned above in separate modules. These include (1) a MAPF planner, (2) a MAPF problem instance generator, (3) an invocation policy, and (4) a fail policy. We provide documentation so that users can customize each module to conduct empirical experiments relevant to their research. 
We also implement state-of-the-art solutions in each module.
Although some of these solutions have been studied in prior LMAPF work, they are almost all evaluated under highly simplified assumptions, most commonly the pebble motion model with perfect execution and communication. Thus, such evaluations could systematically overestimate performance or even lead to qualitatively different conclusions when applied to realistic settings.
LSMART enables these design choices to be evaluated under realistic execution conditions at scale, supporting experiments with thousands of AGVs in physics-based simulation.
Our empirical study shows that system-level design choices may behave differently when evaluated in realistic FMS settings compared to idealized MAPF models, underscoring the need for execution-aware evaluation tools.

\section{Background}
LMAPF is often solved by decomposing the problem into a sequence of MAPF problems and exploiting an existing MAPF algorithm to solve them. However, deploying MAPF algorithms in an FMS requires more than a sequential decomposition. In this section, we first formally define the \emph{standard} MAPF and LMAPF, along with the \emph{pebble motion agent}, the agent model predominantly used in prior MAPF and LMAPF works. We then define the AGV and the FMS that are simulated in \lsmartpro. Finally, we discuss related work on LMAPF and FMS. 

\subsection{Lifelong MAPF and Its Design Choices} \label{sec:back:sub-design}

\begin{definition} [Pebble Motion Agent]
    Given a 2D grid graph $G(V,E)$, a pebble motion agent must be placed at a vertex. It can move to an adjacent vertex or wait at its current vertex in each discretized timestep. Two pebble motion agents collide if they reach the same vertex or swap vertices at the same timestep.
\end{definition}

\begin{definition} [Standard MAPF]
Given a 2D grid graph $G(V,E)$ and $k$ pebble motion agents, each with a start and a goal, the standard MAPF aims to search for collision-free paths from the starts to the corresponding goals. Standard MAPF minimizes the sum-of-costs, defined as the sum of the path lengths of all agents, where a path length equals the number of timesteps it takes for an agent to arrive at its goal.
\end{definition}

\begin{definition} [Standard Lifelong MAPF]
Standard lifelong MAPF (LMAPF) is a variant of standard MAPF where agents are assigned new goals upon reaching their current goals. Standard LMAPF aims to maximize throughput, defined as the number of goals reached per timestep.
\end{definition}

\begin{definition} [Automated Guided Vehicles (AGV)]
Given a 2D workspace, an AGV is modeled as a differential drive robot that can move forward or rotate in place in continuous time.
Each AGV has an onboard controller that executes motions subject to speed and acceleration limits. The state of an AGV is determined by $(x,y,\theta,t)$, where $x,y \in \mathbb{R}$ determines its planar location, $\theta \in [0, 2\pi)$ determines the orientation, and $t \in \mathbb{R}_{\geq 0}$ determines the time.
\end{definition}

\begin{definition} [AGV Fleet Management System (FMS)]
An FMS is a centralized lifelong system in which a fleet of AGVs is coordinated to solve a lifelong MAPF problem in a 2D rectangular workspace. The space is evenly divided into a grid consisting of $M \times N$ cells, some of which are non-traversable obstacles. Each AGV can occupy at most one cell. It can either wait or rotate in a cell, or move to a neighboring cell. Two AGVs are considered to collide if they occupy the same cell or swap cells during overlapping time intervals.
During execution, AGVs are subject to kinodynamic constraints and receive their planned paths through network communication. Each AGV uses an onboard controller to track its assigned path.
The system considers: (1) communication delays among the AGVs, (2) execution uncertainties in action completion times, (3) concurrent planning and execution, and (4) planning failure recovery.
\end{definition}


\subsubsection{Agent Models in Planning}
Prior works in MAPF predominantly follow the standard MAPF definition, where pebble motion agents are assumed in planning~\cite{SternSoCS19,SharonAIJ15,Ma2018SearchingWC,okumura2019priority}. However, the pebble motion model fails to account for communication delays, execution uncertainties, and the kinodynamics of AGVs. Therefore, to tackle execution uncertainties, \citet{Atzmon2020krobust} proposes the $k$-robust delay model, where the MAPF solution is guaranteed to be collision-free when each agent is delayed for at most $k$ timesteps. To account for kinodynamics, some works incorporate more realistic robot models, such as rotation and kinematic limits~\cite{cohen2019optimal,zhang2023efficient,yan2025mass}. However, these works always assume that the agent models in planning are always exactly the same as the agent models in execution, resulting in inaccurate experimental evaluations.

\subsubsection{Execution Policy}
Although it is possible to consider \emph{some} real-world factors during planning, it is impractical to build perfect agent models for planning that account for \emph{all} of such factors.
Therefore, another line of research attempts to develop execution policies capable of executing plans found using simplified agent models. A pioneering work proposes the Action Dependency Graph (ADG)~\cite{Hnig2019PersistentAR}, where AGVs are modeled as pebble motion agents during planning. The path of each agent is converted into a sequence of actions. If two agents visit the same location, ADG constructs a dependency between the corresponding actions of the agents. During execution, the AGVs can then robustly execute their paths following the action dependencies according to the ADG. 
More recent works~\cite{SuAAAI24,FengICAPS24,JiangAAAI25} seek to optimize ADG by switching inter-agent dependencies. However, they are developed on the basis of the pebble motion model and are non-trivial to extend to FMS.
Another work~\cite{zhang_concurrent_2025} proposed different execution policies for pebble motion agents based on PIBT~\cite{okumura2019priority}, a one-step rule-based algorithm. At each timestep, each agent attempts to move to a location closer to its goal. If agents collide, a simple rule based on priority inheritance and backtracking is applied to resolve the collision. Since PIBT is designed for pebble motion agents, it is non-trivial to extend them to FMS.

\begin{table*}[!t]
    \centering
    \resizebox{1\linewidth}{!}{
    \begin{tabular}{l|l|ccccccc}
        \toprule
          &                                    & Agent Model (Planning) & Agent Model (Execution) & Execution Policy & Planner Invocation Policy    & Instance Generator & Fail Policy & Open-source \\
        \midrule
        1 & \citet{Hnig2019PersistentAR}       & Pebble Motion & Differential Drive & ADG     & \Noaction                    & \DistinctOneGoal           & N/A & \xmark \\
        2 & \citet{VaramballySoCS22}           & Pebble Motion & Differential Drive & ADG     & \Noaction                    & \WindowedMultiGoal           & N/A & \xmark \\
        3 & \citet{Li2020LifelongMP}           & Pebble Motion & Pebble Motion & N/A          & Periodic                     & \WindowedMultiGoal           & \lrgw & \cmark \\
        4 & \citet{zhang_planning_2024}        & Pebble Motion & Pebble Motion & N/A          & Periodic                     & \WindowedMultiGoal           & Generalized ADG & \cmark \\
        5 & \citet{chan2024lorr}               & Rotation Motion & Rotation Motion & N/A          & Periodic                     & \WindowedMultiGoal           & N/A & \cmark \\
        6 & \citet{zhang_concurrent_2025}      & Pebble Motion & Pebble Motion & ADG, PIBT-Based & Periodic               & \WindowedMultiGoal           & N/A & \cmark \\
        7 & \citet{Morag2025TMAPF}             & Pebble Motion & Pebble Motion & N/A            & Periodic                   & \OneGoal + MAPF4L                      & Rule-based & \cmark \\
        8 & \lsmartpro (ours)                  & Any           & Differential Drive & ADG              & Multiple          & Multiple                         & Multiple & \cmark \\
        \bottomrule
    \end{tabular}
    }
    \caption{Overview of previous FMS software tools.}
    \label{tab:lmapf_tools}
\end{table*}

\subsubsection{Planner Invocation Policy}

Previous LMAPF algorithms employ primarily a \emph{periodic} invocation policy, in which MAPF planners are invoked periodically after a few timesteps~\cite{Li2020LifelongMP,LiuAAMAS19} or once per timestep~\cite{okumura2019priority}.
Meanwhile, works studying ADG-based execution policies~\cite{Hnig2019PersistentAR,VaramballySoCS22} use an \emph{\noaction} invocation policy in which the planner is only invoked when one of the AGVs is expected to finish executing all its actions in the ADG before the planner returns a new batch of paths. To the best of the authors' knowledge, no prior work has systematically studied invocation policy or compared the proposed invocation policies.

\subsubsection{Duplicate Goals Resolution in Instance Generator}
In FMS, it is common for two agents to have duplicate goals.
However, duplicate goals may cause standard MAPF solvers to fail, since they assume agents remain at their goals indefinitely.
To resolve duplicate goals, one can assign a temporary goal to an agent and allow it to return to its original goal after the other agent releases that goal, namely the \emph{\distinctOneGoal} instance generator. However, this could result in unnecessary detours for AGVs because traveling to temporary goals does not contribute to throughput.
\citet{LiuAAMAS19} assigns a distinct home location for each agent and appends them as the last goals to avoid duplicate goals. The paths to the home locations are never executed. This method avoids taking detours in temporary goals, but requires additional planning efforts.
\citet{Li2020LifelongMP} leverages a \emph{\windowedMultiGoal} instance generator that assigns multiple goals to each agent and plans collision-free paths within a pre-defined time window, but it requires a windowed MAPF algorithm.
\citet{Morag2025TMAPF} reformulates the MAPF problem as \emph{MAPF for Lifelong} (MAPF4L), which allows agents to move freely after reaching their goals. MAPF4L planners can solve standard MAPF problems regardless of duplicate goals. 
Notably, all of these works evaluate their methods using the unrealistic pebble motion model, and none of the previous works have compared these different methods on resolving duplicate goals.

\subsubsection{Fail Policy}
In case the MAPF planner fails to return collision-free paths, an FMS must adapt to the failure. One approach is to replan using a separate, fast, but suboptimal planner, such as PIBT~\cite{okumura2019priority}, the runtime of which can be neglected. 
If the MAPF planner returns colliding paths, which can be intermediate solutions found by the planner when the timeout is reached, it can be helpful to exploit such results rather than abolish them. 
\citet{morag_adapting_2023} propose a simple rule-based fail policy that (1) allows agents whose paths have no collisions to move and (2) instructs agents that are conflicting with others to wait at their start locations.
If a waiting agent collides with a moving agent, the waiting agent can make a single move to avoid the collision if possible. Otherwise, the moving agent must stop. However, this strategy might instruct too many agents to wait. 
The open-source code repository of \citet{Li2020LifelongMP} implements \LRGW (\lrgw), which guides agents along colliding paths and inserts as few wait actions as possible to generate collision-free paths.
\citet{zhang_planning_2024} proposes a similar fail policy based on a generalized ADG which inserts wait actions to avoid collisions and preserve passing orders of the agents in each location.
Both generalized ADG and \lrgw resolve collisions only by adding wait actions, which could cause too many agents to wait indefinitely.
A more recent work has proposed Guided PIBT. which instructs agents to follow a pre-defined set of paths, known as guide paths, while resolving collisions~\cite{ChenAAAI24} using PIBT. 
Guided PIBT was originally proposed as an LMAPF algorithm, but it can also be an effective fail policy. 
All the aforementioned fail policies are studied while assuming the pebble motion model. In this work, we select representative policies and conduct a systematic comparison among them.

\subsection{FMS Software}
\Cref{tab:lmapf_tools} summarizes the existing FMS software.
Works 1 and 2 use differential drive robots with speed and acceleration limits during execution, but none of them are open-source. Work 5 considers the rotation motion model, where agents can move forward or rotate in discretized time and space. All other works use the pebble motion model.
For the execution policy, works 1, 2, and 6 consider ADG~\cite{Hnig2019PersistentAR}. 
Since the PIBT-based execution policy cannot be trivially extended to FMS, \lsmartpro uses ADG.
For instance generators, all previous works support only one method, while \lsmartpro supports multiple. 
For planner invocation policies, previous work supports either periodic or \noaction policy, while \lsmartpro supports both. 
For fail policies, work 3 supports \lrgw, work 4 supports a generalized ADG, and work 7 supports several rule-based fail policies. All avoid collisions by adding wait actions to the colliding paths returned by the planner. \lsmartpro supports a wider variety, including fail policies that depend on colliding paths and those that do not.

\section{\lsmartpro}

\begin{figure*}[t]
    \centering
    \includegraphics[width=1\linewidth]{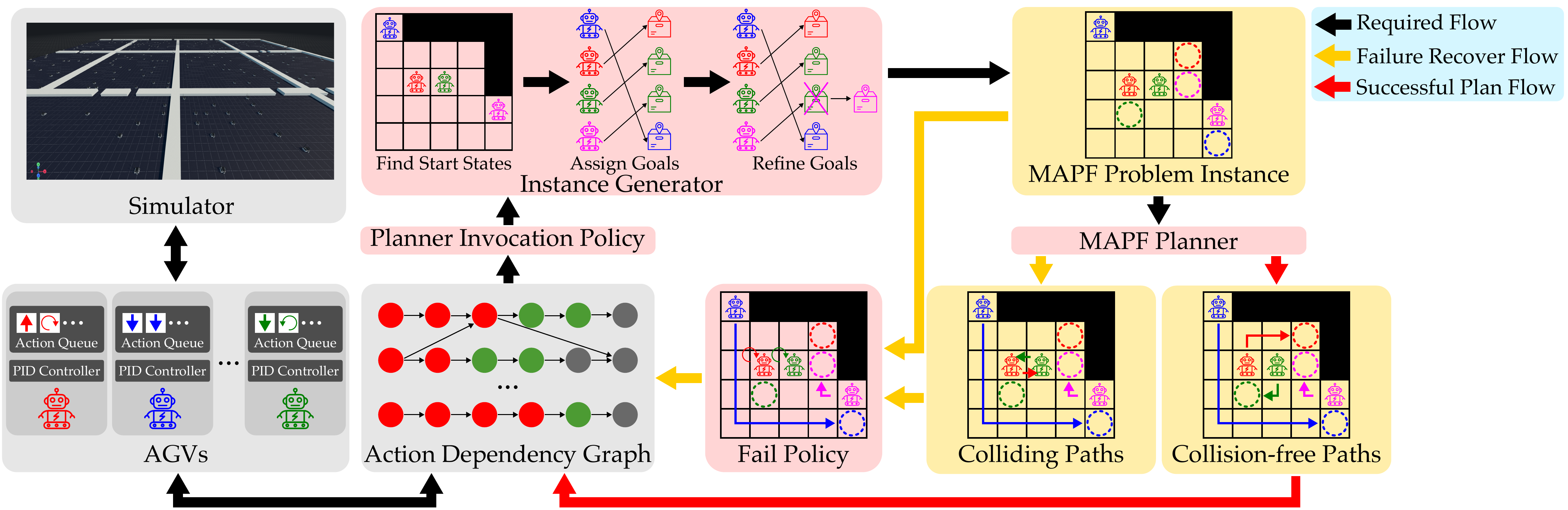}
    \caption{Detailed architecture of \lsmartpro. Red boxes are user customizable modules, gray boxes are non-customizable modules, and yellow boxes are data structures used for communicating between modules. Black arrows indicates required flow, red arrows indicate flow when the MAPF planner successfully finds collision-free paths, and yellow arrows indicate flow when the planner fails and the system need to recover from failure.}
    \label{fig:lsmart2}
\end{figure*}

\Cref{fig:lsmart2} shows the architecture of \lsmartpro. It consists of 7 modules: (1) a planner invocation policy, (2) an instance generator, (3) a MAPF planner, (4) a fail policy, (5) an ADG, (6) a fleet of AGVs, and (7) a physics-based simulator.
When generalizing \smart to \lsmartpro, we use the same simulator, ADG, and AGV fleet, while adding or updating other modules.
In this section, we first provide an overview of \lsmartpro and discuss each added or updated module in detail.

\subsubsection{System Overview}
A simulation starts by using the planner invocation policy to determine whether the MAPF planner should be invoked. If so, \lsmartpro first uses the instance generator to generate the next MAPF problem instance by computing a commit cut in the ADG to find the start states, assigning goals to AGVs, and refining the goals if necessary. At the start of the simulation, the start locations of the AGVs are randomly generated with all agents facing north. The instance is passed to the MAPF planner. If the planner solves it successfully, the collision-free paths are returned directly to the ADG. If it fails, the fail policy replans or resolves the collisions. Depending on the fail policy, the MAPF planner may not be required to send colliding paths to the fail policy. The paths are then converted and added to the ADG, which records a sequence of actions for each AGV with their passing orders in each location.
The AGVs and the simulator run in parallel with the planner. Each AGV is equipped with a PID controller and an action queue. Each agent periodically obtains actions from the ADG following the action dependencies, stores them in the action queue, and executes them in the simulator. If no actions are left, it waits in place. The simulation stops after a pre-defined amount of time.



\subsubsection{Planner Invocation Policy}
\smart is designed for standard MAPF, and therefore does not need an invocation policy.
\lsmartpro supports two planner invocation policies, namely the periodic policy, where the planner is invoked periodically every $P$ seconds, and the \noaction policy, where the planner is invoked when at least one AGV is expected to complete all its assigned actions in the ADG before the planner can return the next plan. To determine whether an AGV is finishing its actions, we pre-define a time limit of $T$ seconds for the planner and compute the minimal amount of time for an AGV to finish one action as $\epsilon$ seconds based on the kinodynamic model of the AGV. If an AGV has fewer than $\frac{T}{\epsilon}$ actions left in the ADG, which is the maximum number of actions the AGV can finish in $T$ seconds, the \noaction policy invokes the planner. If the planner is to be invoked, the invocation policy awakens the instance generator to generate the next MAPF problem instance.


\subsubsection{Instance Generator}
The instance generator is responsible for generating the next MAPF problem instance by (1) computing a commit cut on the ADG to determine the start states of the AGVs, (2) assigning goals to the AGVs, and (3) refining the goals to satisfy planner-specific assumptions if necessary.
To compute the commit cut, we use the same algorithm from ~\citet{Hnig2019PersistentAR}, which looks into the future for $\frac{T}{\epsilon}$ actions for each AGV. Since \lsmartpro focuses on evaluating MAPF algorithms, we use a random goal assigner.
Depending on the MAPF planner, \lsmartpro provides three types of instance generator: (1) \distinctOneGoal, (2) \oneGoal, and (3) \windowedMultiGoal. 
The \distinctOneGoal instance generator first assigns one goal to each AGV. If duplicate goals are detected, a temporary goal closest to the original goal to which no other agents are going is generated and assigned.
The \distinctOneGoal instance generator is compatible with all standard MAPF algorithms, which expect distinct goals. 
The \oneGoal instance generator simply assigns one goal to each AGV and does not perform any refinements. Therefore, it is only compatible with MAPF algorithms that can handle duplicate goals. In this paper, we present experiment results with the MAPF4L model~\cite{Morag2025TMAPF} for its state-of-the-art performance in this category. The \windowedMultiGoal instance generator assigns at least one goal to each AGV, given a window size, and expects a windowed MAPF planner to plan collision-free paths within the window. It also does not perform refinements. It guarantees that no AGVs can finish their goals within the pre-defined window. Notably, while these three instance generators have been studied in various prior LMAPF works, we are the first to present a comprehensive comparative study of them in an FMS.

\subsubsection{Fail Policy}
\lsmartpro supports four fail policies. Since the MAPF planner may not return colliding paths, \lsmartpro supports (1) replanning from scratch using PIBT~\cite{okumura2019priority} and (2) asking all AGVs to wait in place~\cite{morag_adapting_2023}. If the MAPF planner is capable of returning colliding paths in the case of failure and we want to exploit such paths, \lsmartpro supports (1) Guided PIBT~\cite{ChenAAAI24} and (2) \lrgw~\cite{Li2020LifelongMP}.


\subsubsection{MAPF Planner}
The MAPF planner takes in a MAPF problem instance with a time limit and returns collision-free paths within that time limit. In \smart, the MAPF planner is only invoked once and does not need to return colliding paths in case of failure. However, in \lsmartpro, the MAPF planner should return colliding paths in case of failure if the fail policy expects them.
\lsmartpro supports MAPF planners of different AGV models, including the pebble motion model, the rotation model, the differential drive model, and the $k$-robust model~\cite{Atzmon2020krobust}. \lsmartpro also supports planners that plan in continuous time. 



\section{Experimental Evaluation} \label{sec:exp}

\begin{table*}[!t]
    \centering
    \small
    \resizebox{1.0\linewidth}{!}{
    \begin{tabular}{c|c|c|c|c|c|c|c|c}
        \toprule
        Setup & Agent Model (Planning) & Planner            & W (s)    & P (s)                    & T (s) & Planner Invocation Policy & Instance Generator & Fail Policy \\
        \midrule
        1     & Pebble Motion          & Standard PBS       & $\infty$ & 1                        & 1     & Periodic                   & \DistinctOneGoal   & PIBT \\
        2     & Pebble Motion          & Standard PBS       & $\infty$ & N/A                      & 1     & \Noaction                  & \DistinctOneGoal   & PIBT \\
        3     & Pebble Motion          & Windowed PBS       & 1        & 1                        & 1     & Periodic                   & \WindowedMultiGoal & PIBT \\
        4     & Pebble Motion          & Windowed PBS       & 1        & N/A                      & 1     & \Noaction                  & \WindowedMultiGoal & PIBT \\
        5     & Pebble Motion          & MAPF4L PBS         & $\infty$ & 1                        & 1     & Periodic                   & \OneGoal           & PIBT \\
        6     & Pebble Motion          & MAPF4L PBS         & $\infty$ & N/A                      & 1     & \Noaction                  & \OneGoal           & PIBT \\
        7     & Pebble Motion          & Windowed PBS       & 10       & 10                       & 10    & Periodic                   & \WindowedMultiGoal & PIBT \\
        8     & Pebble Motion          & Windowed PBS       & 10       & N/A                      & 10    & \Noaction                  & \WindowedMultiGoal & PIBT \\
        9     & Pebble Motion          & Windowed PBS       & 2        & 2                        & 2     & Periodic                   & \WindowedMultiGoal & PIBT \\
        10    & Pebble Motion          & Windowed PBS       & 5        & 5                        & 5     & Periodic                   & \WindowedMultiGoal & PIBT \\
        11    & Pebble Motion          & MAPF-LNS2 & $\infty$ & \multicolumn{2}{c|}{$\{0.1,0.5,1,2,...,20\}$} & Periodic                   & \DistinctOneGoal   & All Wait \\
        12    & Pebble Motion          & Windowed PBS       & 1        & 1                        & 1     & Periodic                   & \WindowedMultiGoal & \lrgw \\
        13    & Pebble Motion          & Windowed PBS       & 1        & 1                        & 1     & Periodic                   & \WindowedMultiGoal & Guided PIBT \\
        14    & Pebble Motion          & Windowed PBS       & 10       & 10                       & 10    & Periodic                   & \WindowedMultiGoal & \lrgw \\
        15    & Pebble Motion          & Windowed PBS       & 10       & 10                       & 10    & Periodic                   & \WindowedMultiGoal & Guided PIBT \\
        16    & Pebble Motion          & Standard CBS       & $\infty$ & 20                       & 20    & Periodic                   & \DistinctOneGoal   & All Wait \\
        17    & Rotation Motion        & Standard CBS       & $\infty$ & 20                       & 20    & Periodic                   & \DistinctOneGoal   & All Wait \\
        18    & Pebble Motion          & Standard PP        & $\infty$ & 20                       & 20    & Periodic                   & \DistinctOneGoal   & All Wait \\
        19    & Rotation Motion        & Standard PP        & $\infty$ & 20                       & 20    & Periodic                   & \DistinctOneGoal   & All Wait \\
        \bottomrule
    \end{tabular}
    }
    \caption{Summary of \lsmartpro setups used in the experiments. $W$ is the planning window for windowed MAPF solvers, $P$ is the period to invoke the planner for periodic invocation policy, and $T$ is the runtime limit for one invocation of the MAPF planner, all given in seconds. The $\infty$ sign indicates that the planner plans collision-free paths to all the given goals.}
    \label{tab:exp-config}
\end{table*}

\begin{table}[!t]
    \centering
    \small
    \resizebox{0.7\linewidth}{!}{
    \begin{tabular}{c|c|c}
        \toprule
        Section                                       & Experiment & Setups  \\
        \midrule
        \Cref{sec:exp:subsec:IG}                      & 1          & 1, 2, 3, 4, 5, 6  \\
        \midrule
        \multirow{3}{*}{\Cref{sec:exp:subsec:invoke}} & 2          & 3, 4, 7, 8     \\
                                                      & 3          & 3, 7, 9, 10     \\
                                                      & 4          & 11     \\
        \midrule
        \Cref{sec:exp:subsec:fail}                    & 5          & 3, 7, 12, 13, 14, 15 \\
        \midrule
        \Cref{sec:exp:subsec:planner}                 & 6          & 16, 17, 18, 19 \\
        \bottomrule
    \end{tabular}
    }
    \caption{Summary of experiments and the setups being used.}
    \label{tab:exp}
\end{table}

In this section, we conduct empirical evaluations of key design choices in \lsmartpro. We focus on user-customizable modules, including (1) instance generators, (2) planner invocation policies, (3) failure policies, and (4) agent models and theoretical optimality in the MAPF planner.

\subsection{General Experiment Setup}
\Cref{tab:exp-config} summarizes all \lsmartpro setups used in all experiments and \Cref{tab:exp} summarizes all experiments.
\subsubsection{Map}
We conduct all experiments on six maps, including \warehouseSmall, \warehouseLarge, \mazeSmall, \emptySmall, \randomMid, and \roomMid, shown in the corners of \Cref{fig:ta-ab}. Due to constraints in space, we show part of the results in Appendix~\ref{appen:exp-add}.
The \warehouseSmall is used extensively in previous lifelong MAPF works~\cite{Li2020LifelongMP,zhangLayout23}, and the other five maps are selected from the MAPF benchmark~\cite{SternSoCS19}. On the two warehouse maps, agents move indefinitely between randomly selected workstations (pink) and endpoints (blue). In other maps, agents move between randomly selected empty spaces (white). In all maps, black represents obstacles. During execution, each map is converted to continuous grids, where each grid has a size of $1$ × $1$ m.

\subsubsection{AGV Model in Execution}
During execution, we use differential drive robots with maximum speed of $2\,m/s$, acceleration $2\,m/s^2$ and angular speed of $45^{\circ}/s$.

\subsubsection{Metrics}
For all experiments, we show two metrics: (1) throughput and (2) the ratio of fail policy calls, computed as the ratio between the number of fail policy calls to the number of planner invocations.
In each map, we run simulations with various numbers of agents. For each number of agents, we run 10 simulations, each lasting for 600 simulation seconds. We plot the average as solid lines and the 95\% confidence intervals as shaded areas in the figures.


\subsubsection{Compute Resource}
We conduct all experiments on an HPC with numerous 64-core AMD EPYC 7742 CPUs, each with 256 GB of RAM~\cite{PSCBridgeTwo2021}.

\subsection{Instance Generator} \label{sec:exp:subsec:IG}
\begin{figure*}[!t]
    \centering
    \includegraphics[width=0.9\textwidth]{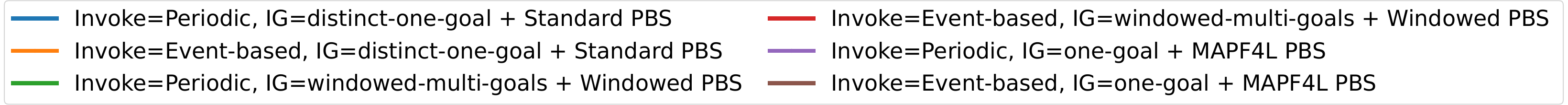}

    \begin{subfigure}{0.33\textwidth}
        \centering
        \includegraphics[width=0.5\textwidth]{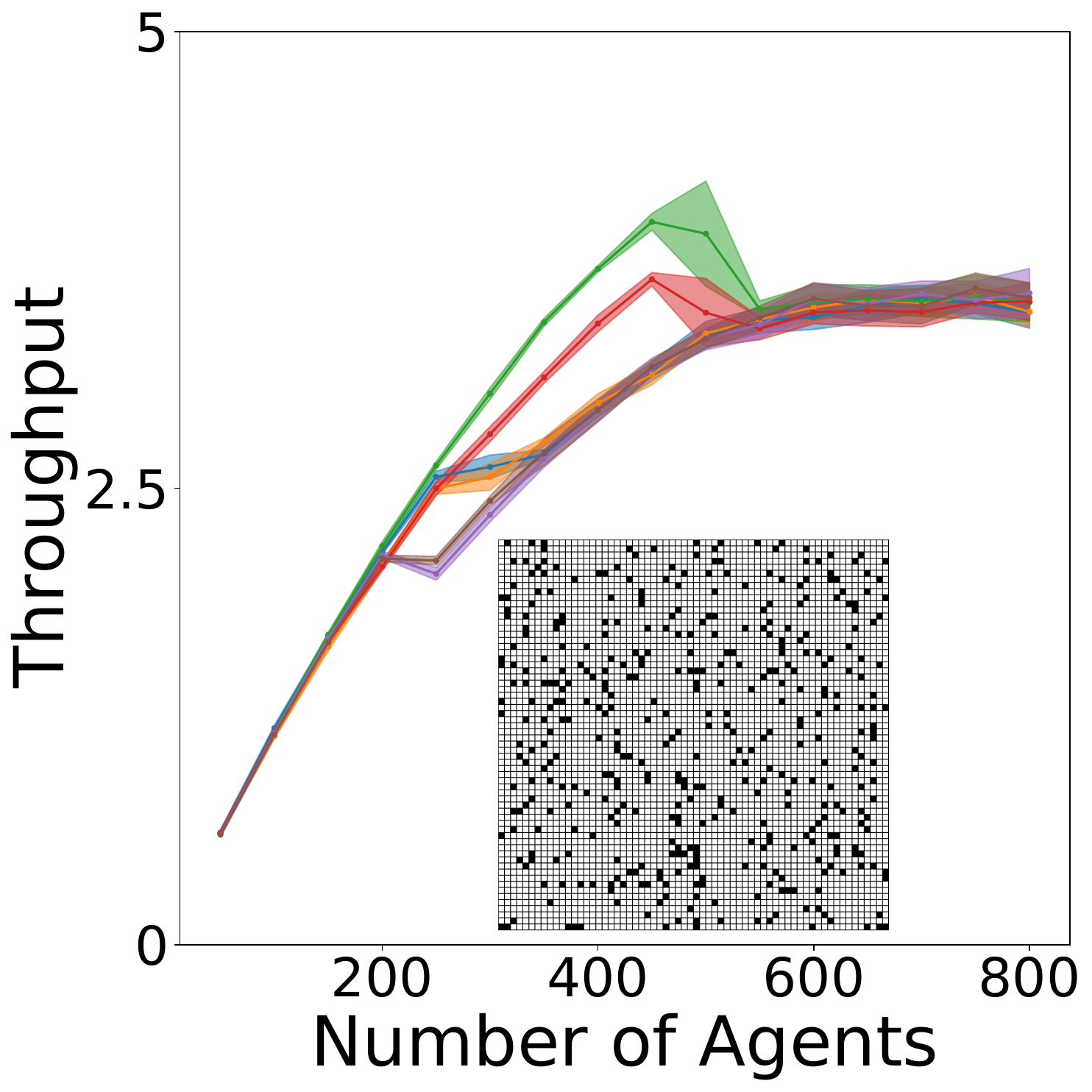}%
        \includegraphics[width=0.5\textwidth]{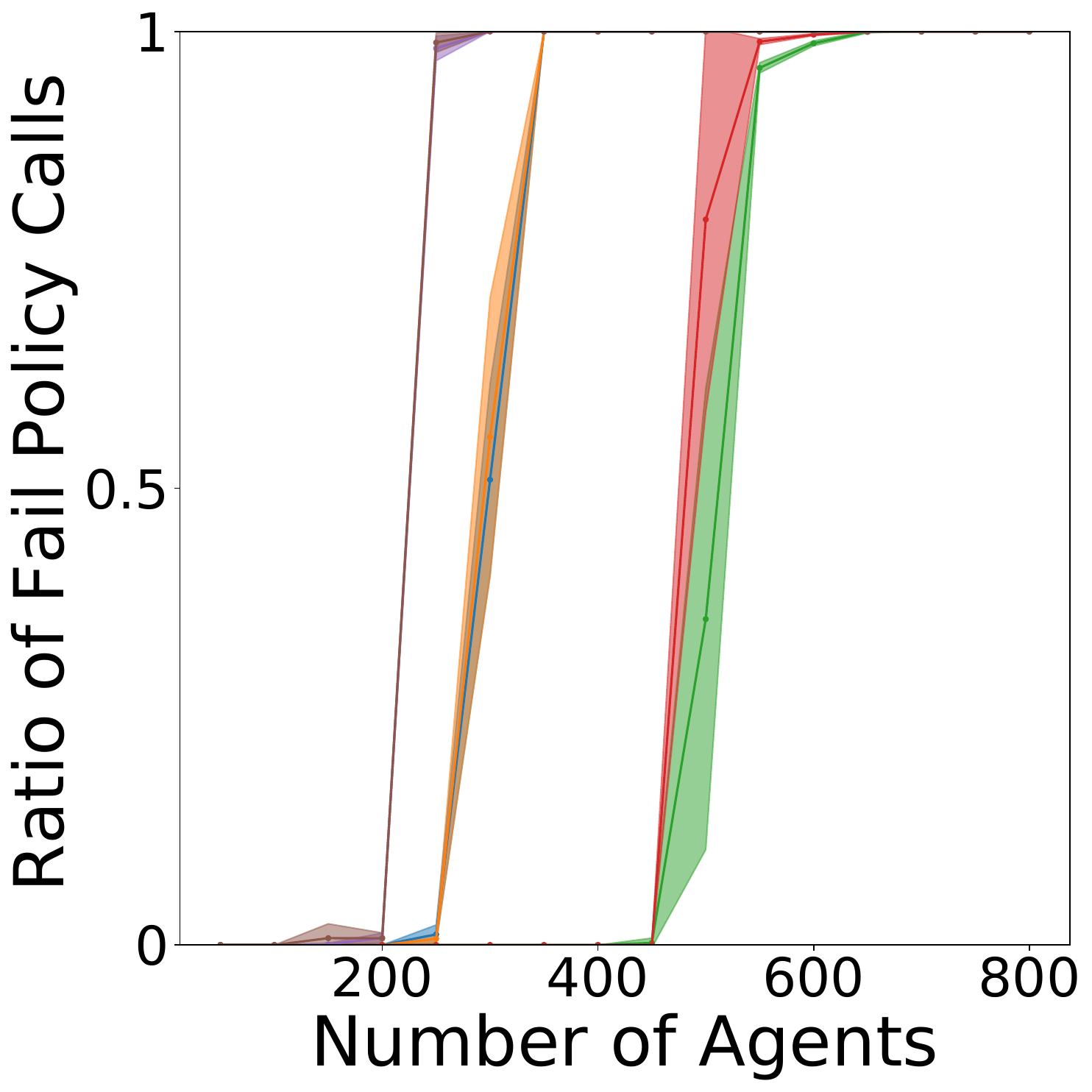}
        \caption{\randomMid}
        \label{fig:ta-ab:random-64-64-10}
    \end{subfigure}%
    \hfill
    \begin{subfigure}{0.33\textwidth}
        \centering
        \includegraphics[width=0.5\textwidth]{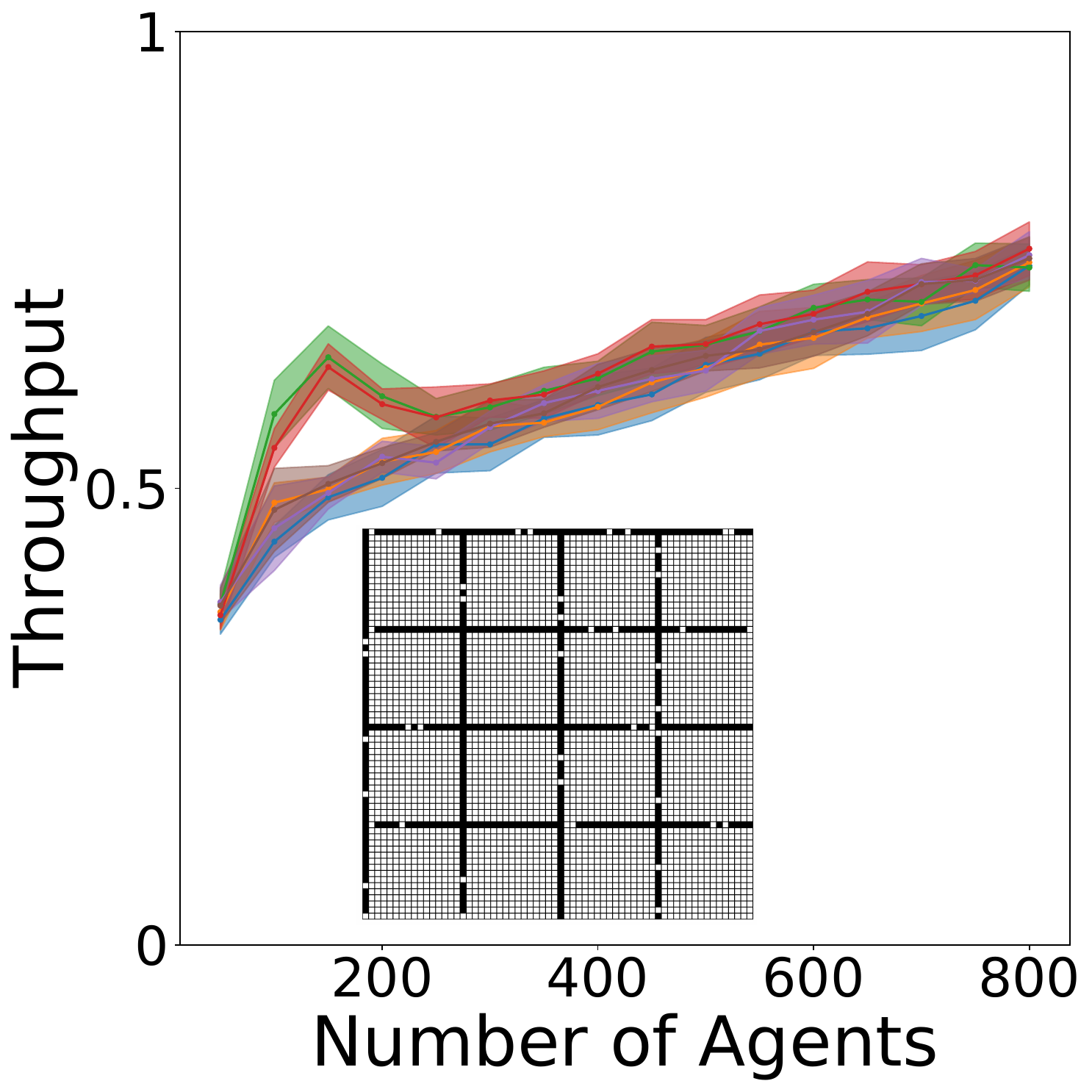}%
        \includegraphics[width=0.5\textwidth]{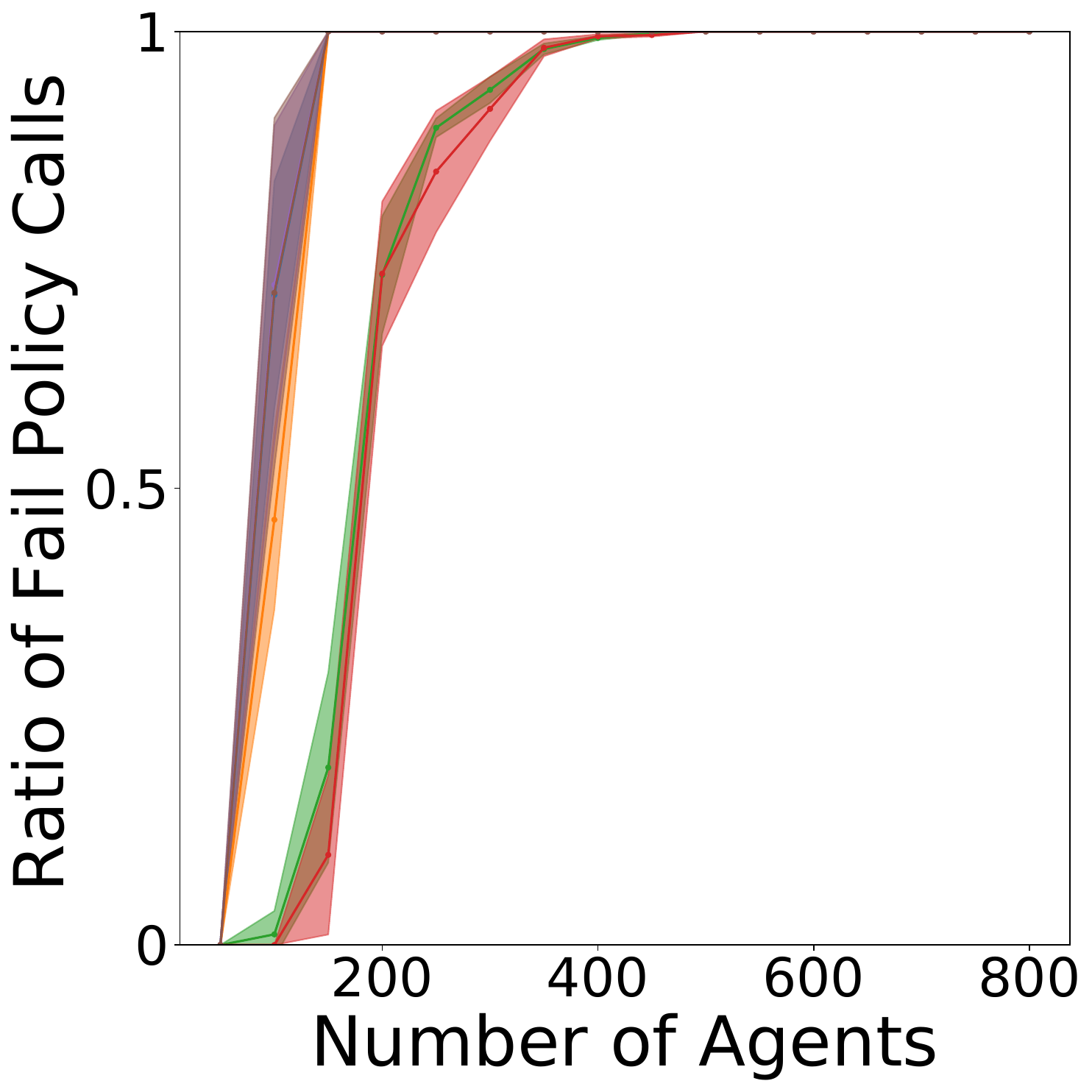}
        \caption{\roomMid}
        \label{fig:ta-ab:room-64-64-16}
    \end{subfigure}
    \hfill
    \begin{subfigure}{0.33\textwidth}
        \centering
        \includegraphics[width=0.5\textwidth]{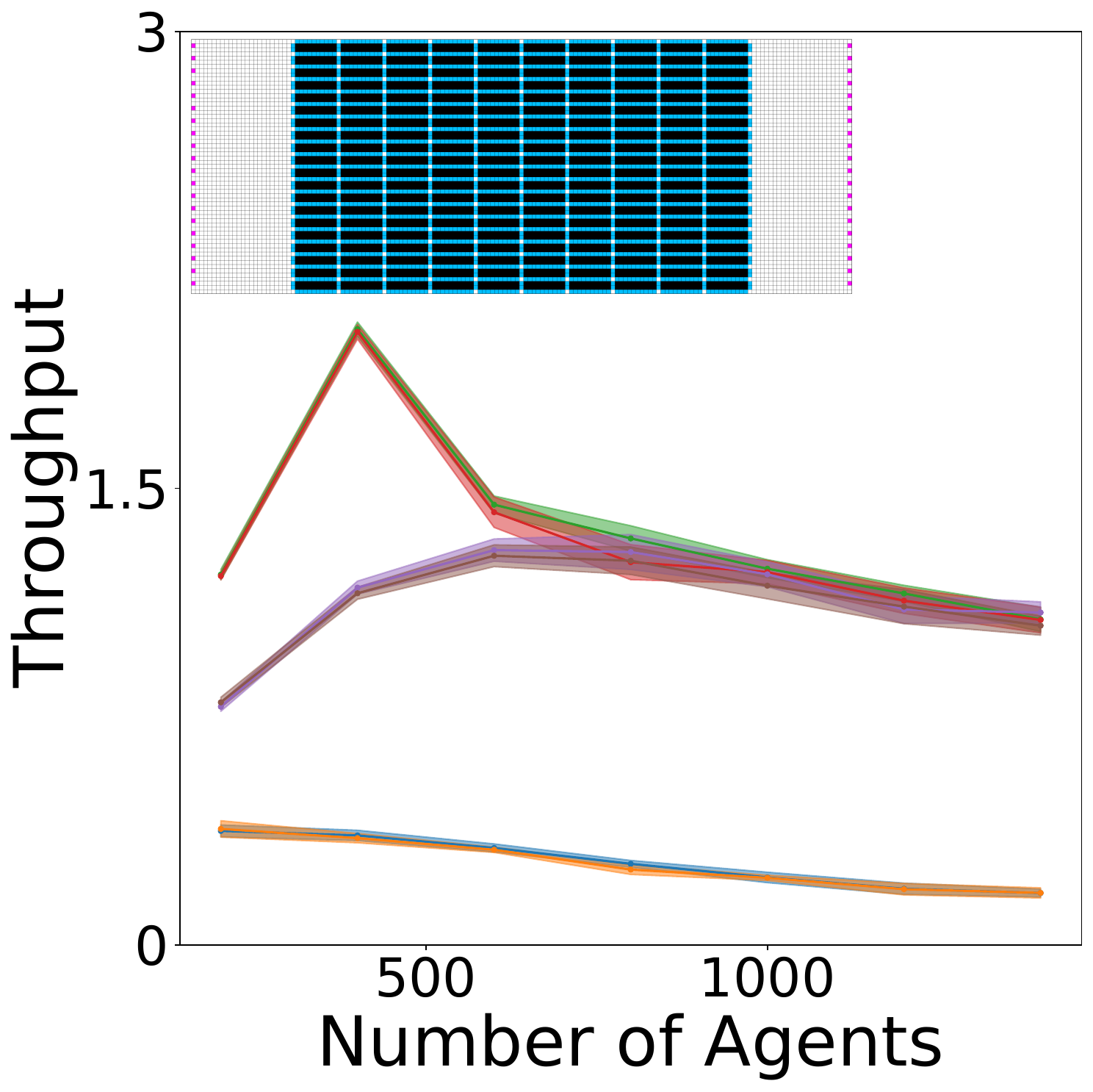}%
        \includegraphics[width=0.5\textwidth]{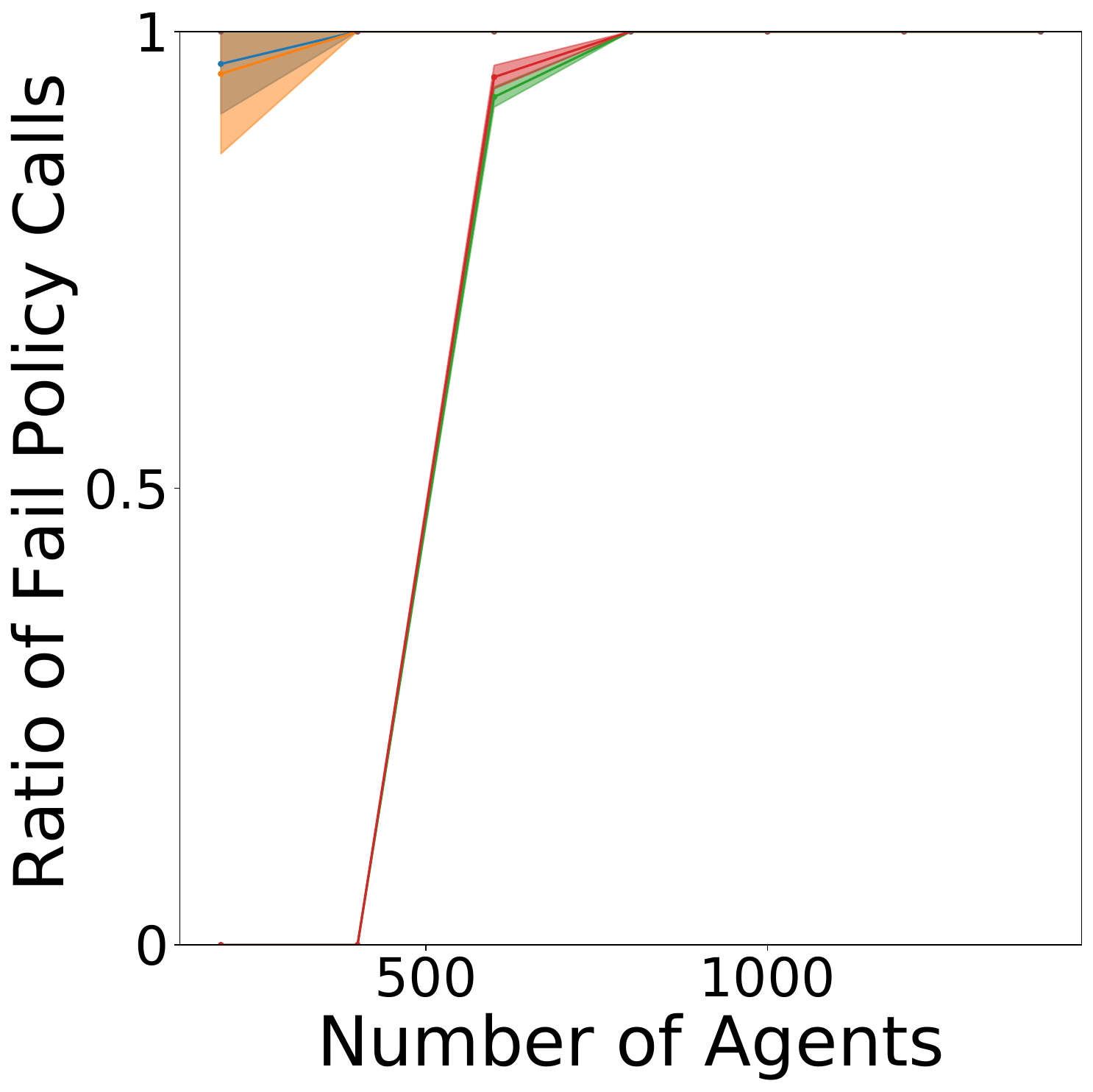}
        \caption{\warehouseLarge}
        \label{fig:ta-ab:warehouse-10-20-10-2-1}
    \end{subfigure}

    \caption{Experiment results of instance generators (Experiment 1 in \Cref{tab:exp}).}
    \label{fig:ta-ab}
\end{figure*}


\subsubsection{Experiment Setup} We perform experiment 1 as described in \Cref{tab:exp}.
We compare three instance generators (IG) each paired with a corresponding MAPF planner. For a fair comparison, we use variants of PBS~\cite{MaAAAI19} as the MAPF planners, resulting in three setups of IGs and planners: (1) \textbf{standard:} \distinctOneGoal IG with standard PBS, (2) \textbf{transient:} \oneGoal IG with MAPF4L PBS~\cite{Morag2025TMAPF}, and (3) \textbf{windowed:} \windowedMultiGoal IG with windowed PBS~\cite{Li2020LifelongMP}. 

\subsubsection{Experiment Result} \Cref{fig:ta-ab} presents the results of experiment 1. 
The windowed setup consistently achieves equal or highest throughput.
This is because the windowed planning mechanism reduces planning effort, allowing the planner to fail less and reducing the ratio of fail policy calls. The comparison between standard and transient setups is in accordance with \citet{Morag2025TMAPF}. When there are a limited number of possible goals, such as in \warehouseLarge, the transient setup outperforms the standard setup. Otherwise, either the two setups are similar (\roomMid) or the standard setup slightly outperforms the transient setup (\randomMid). Additional results are shown in \Cref{fig:ta-ab-add} of Appendix~\ref{appen:exp-add:IG}.

\subsection{Planner Invocation Policy} \label{sec:exp:subsec:invoke}

\subsubsection{Experiment Setup}
We perform experiments 2, 3, and 4 in \Cref{tab:exp}.
For experiment 2, we compare periodic and \noaction invocation policies using the windowed PBS planner with $W=1$ seconds and $W=10$ seconds.
In Experiment 3, we compare different configurations with $P = W$, pairing the periodic invocation policy with planning windows of the same size.
Although \citet{VaramballySoCS22} conducted a similar study with the \noaction policy, their experiments were conducted in a single setup with low agent density and a small map ($50$ robots in a $60 \times 20$ map), whereas our experiments scale to as many as $2000$ robots on maps as large as $61 \times 159$.
In experiment 4, we isolate the impact of $W$ and $P$, performing an ablation on different values of $P$ with a standard MAPF solver.
Intuitively, more frequent replanning can improve the system’s ability to adapt to delays during execution.
To study this effect without confounding it with fail policy, we use MAPF-LNS2~\cite{li2022mapf}, a scalable MAPF solver that consistently produces solutions across all tested settings. We adopt the standard MAPF model with the \distinctOneGoal instance generator.

\begin{figure*}[!t]
    \centering
    \includegraphics[width=0.8\textwidth]{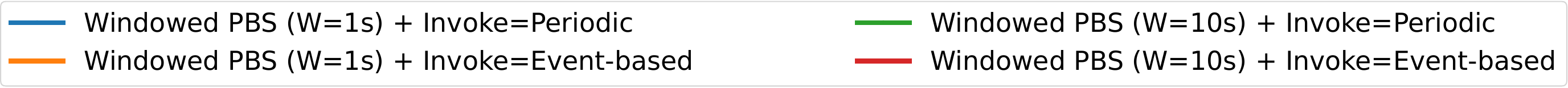}

    \begin{subfigure}{0.33\textwidth}
        \centering
        \includegraphics[width=0.5\textwidth]{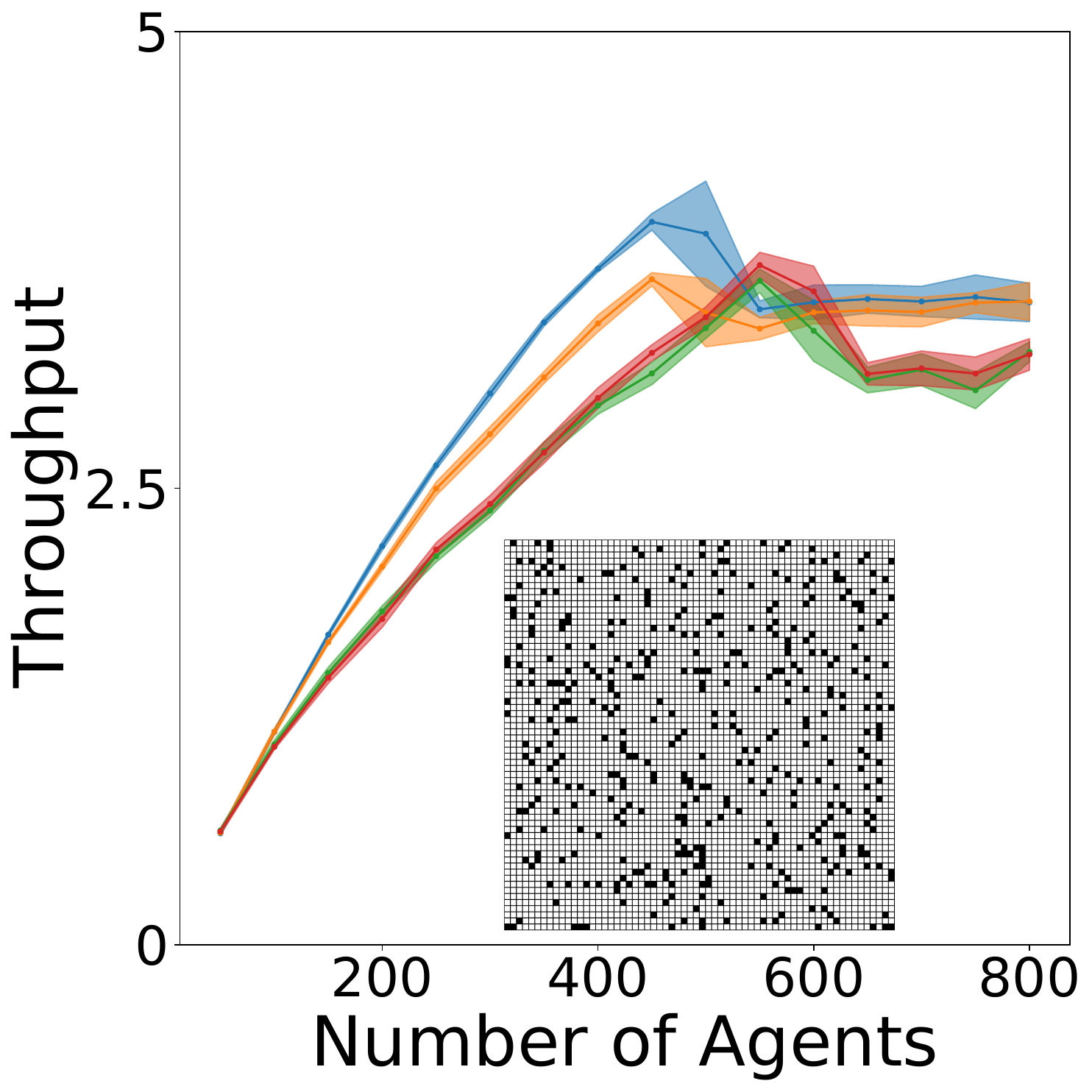}%
        \includegraphics[width=0.5\textwidth]{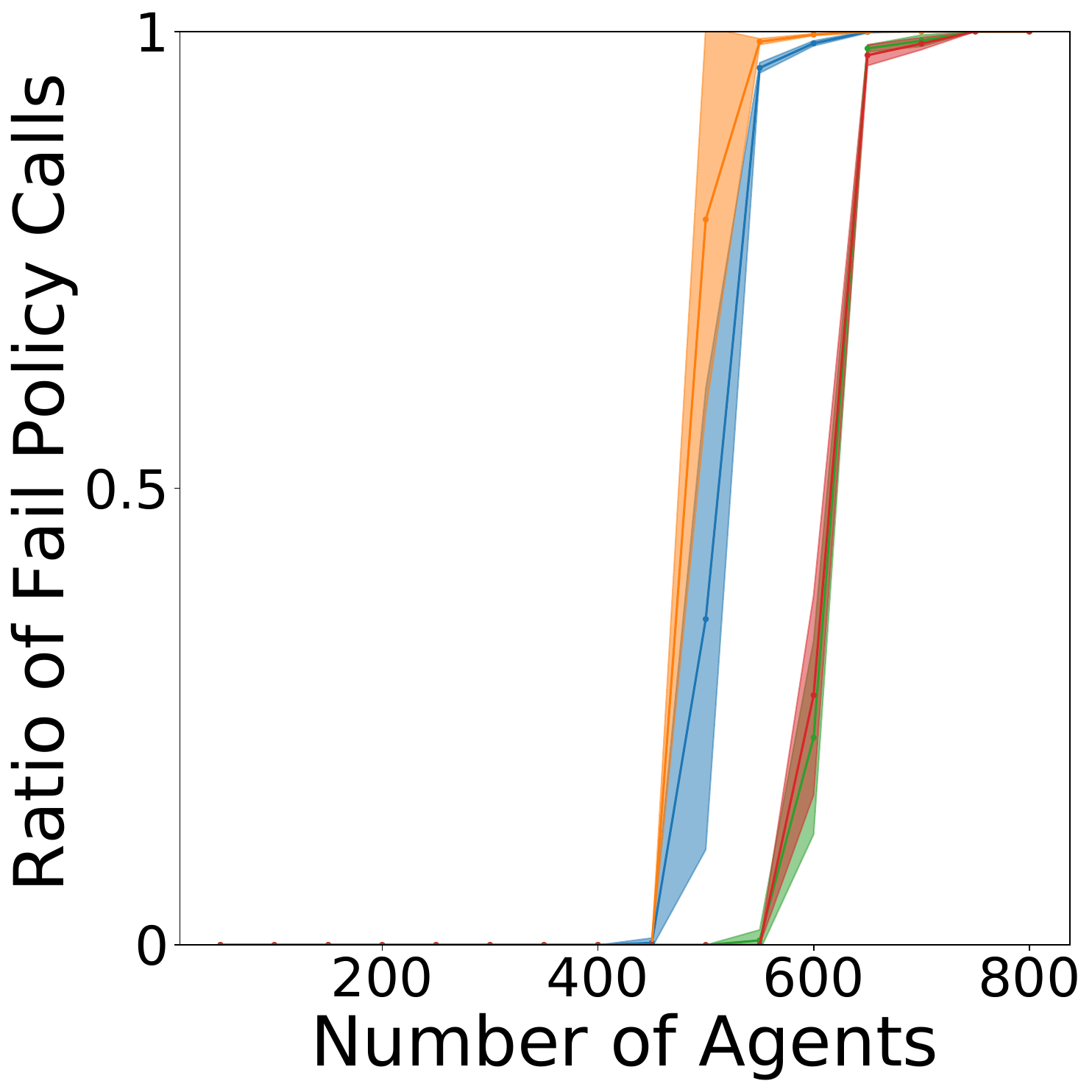}
        \caption{\randomMid}
        \label{fig:plan-invoke-ab:random-64-64-10}
    \end{subfigure}%
    \hfill
    \begin{subfigure}{0.33\textwidth}
        \centering
        \includegraphics[width=0.5\textwidth]{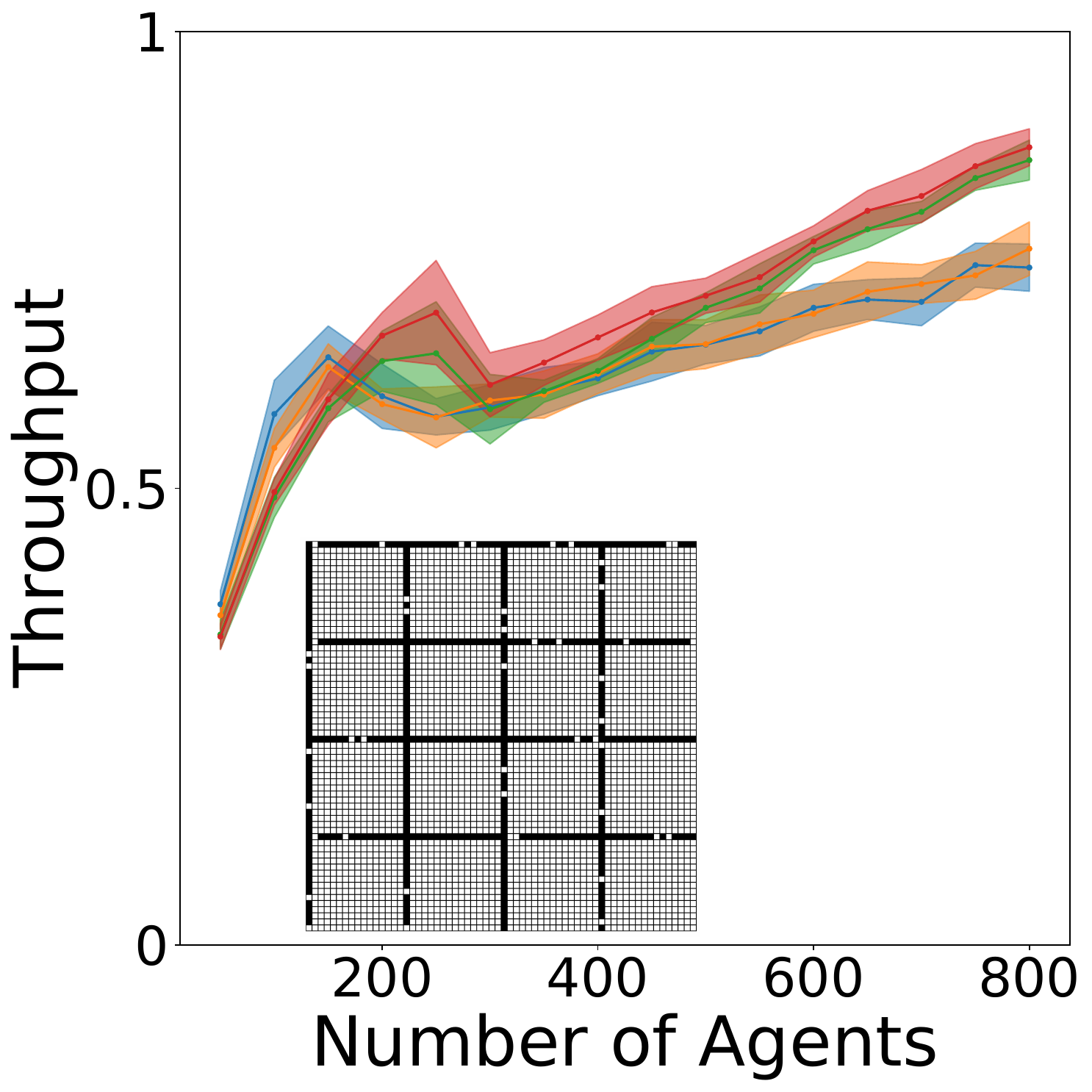}%
        \includegraphics[width=0.5\textwidth]{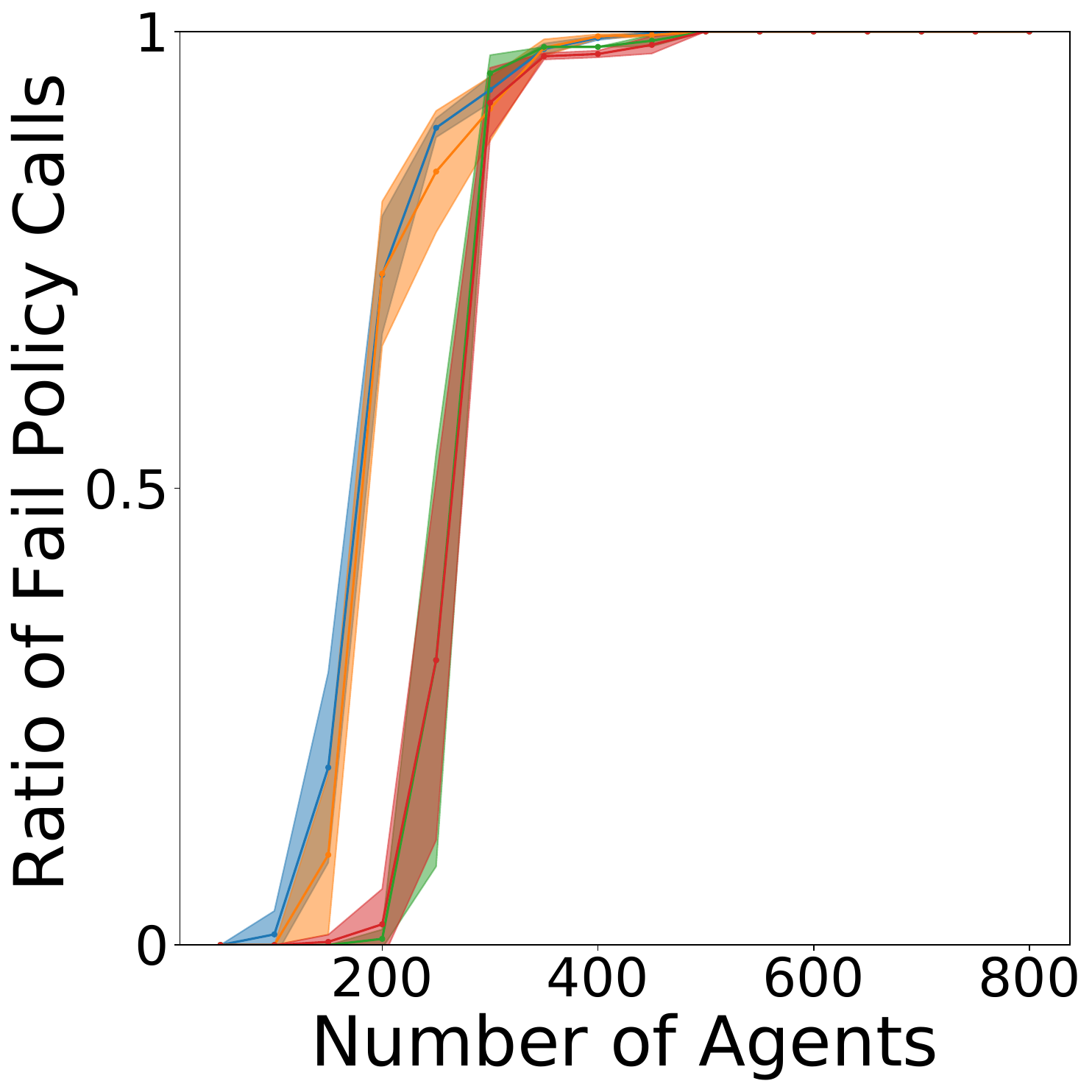}
        \caption{\roomMid}
        \label{fig:plan-invoke-ab:room-64-64-16}
    \end{subfigure}
    \hfill
    \begin{subfigure}{0.33\textwidth}
        \centering
        \includegraphics[width=0.5\textwidth]{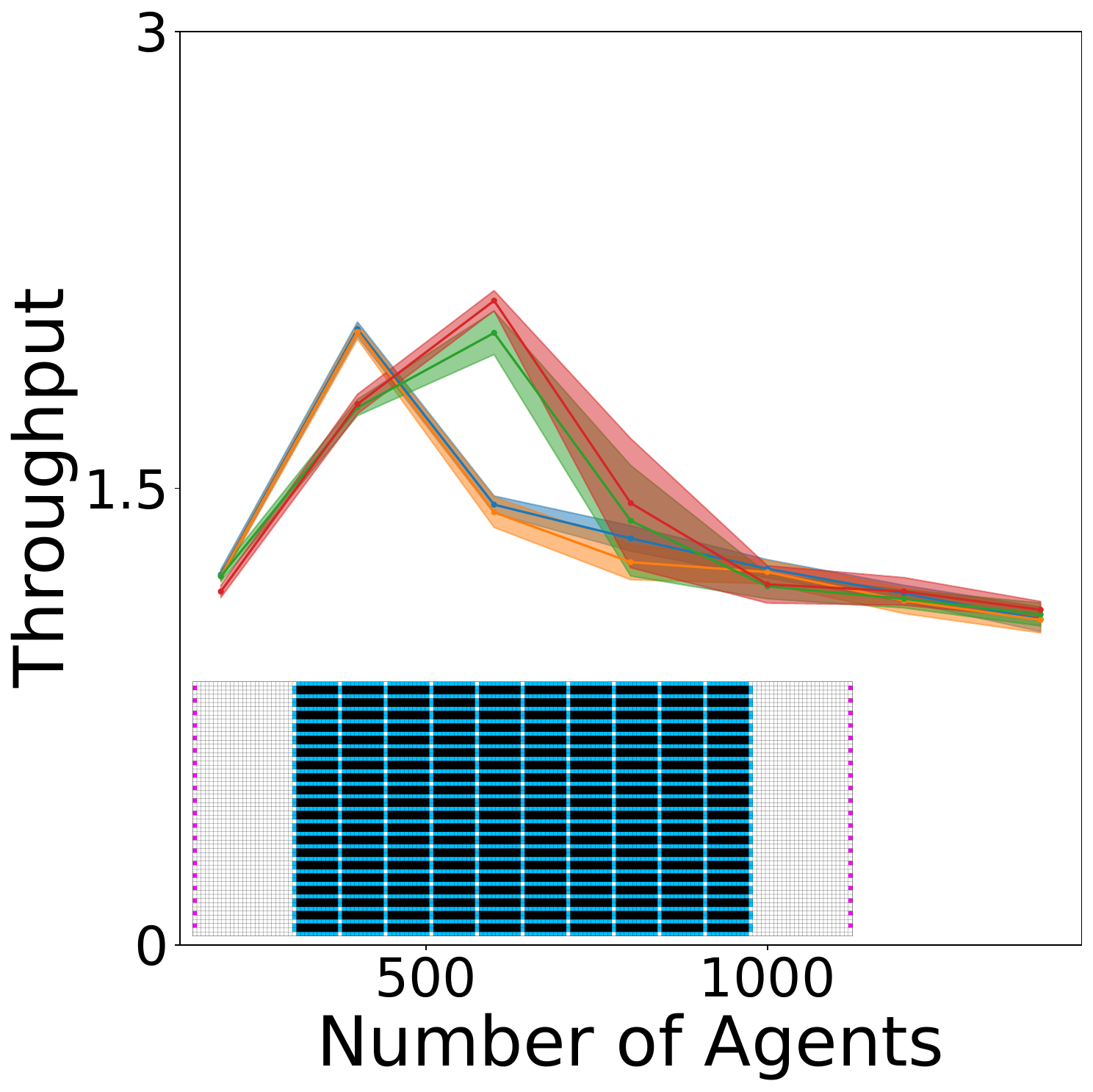}%
        \includegraphics[width=0.5\textwidth]{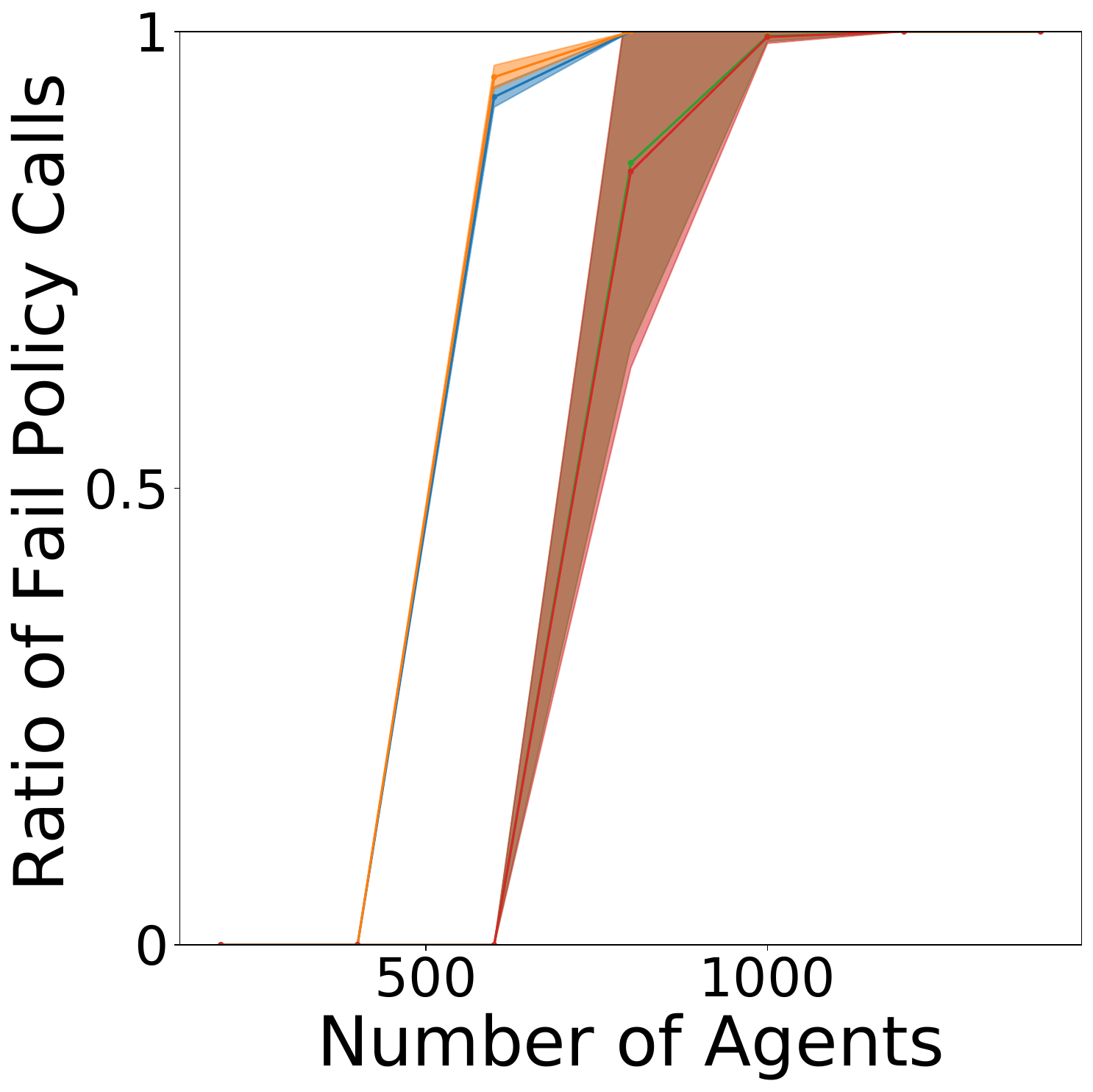}
        \caption{\warehouseLarge}
        \label{fig:plan-invoke-ab:warehouse-10-20-10-2-1}
    \end{subfigure}

    \caption{Experiment results of planner invocation policies (Experiment 2 in \Cref{tab:exp}).}
    \label{fig:plan-invoke-ab}
\end{figure*}

\begin{figure*}[!t]
    \centering
    \includegraphics[width=1\textwidth]{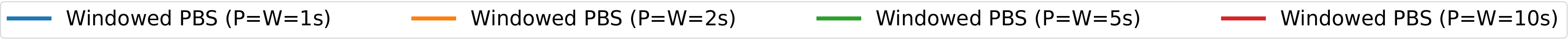}

    \begin{subfigure}{0.33\textwidth}
        \centering
        \includegraphics[width=0.5\textwidth]{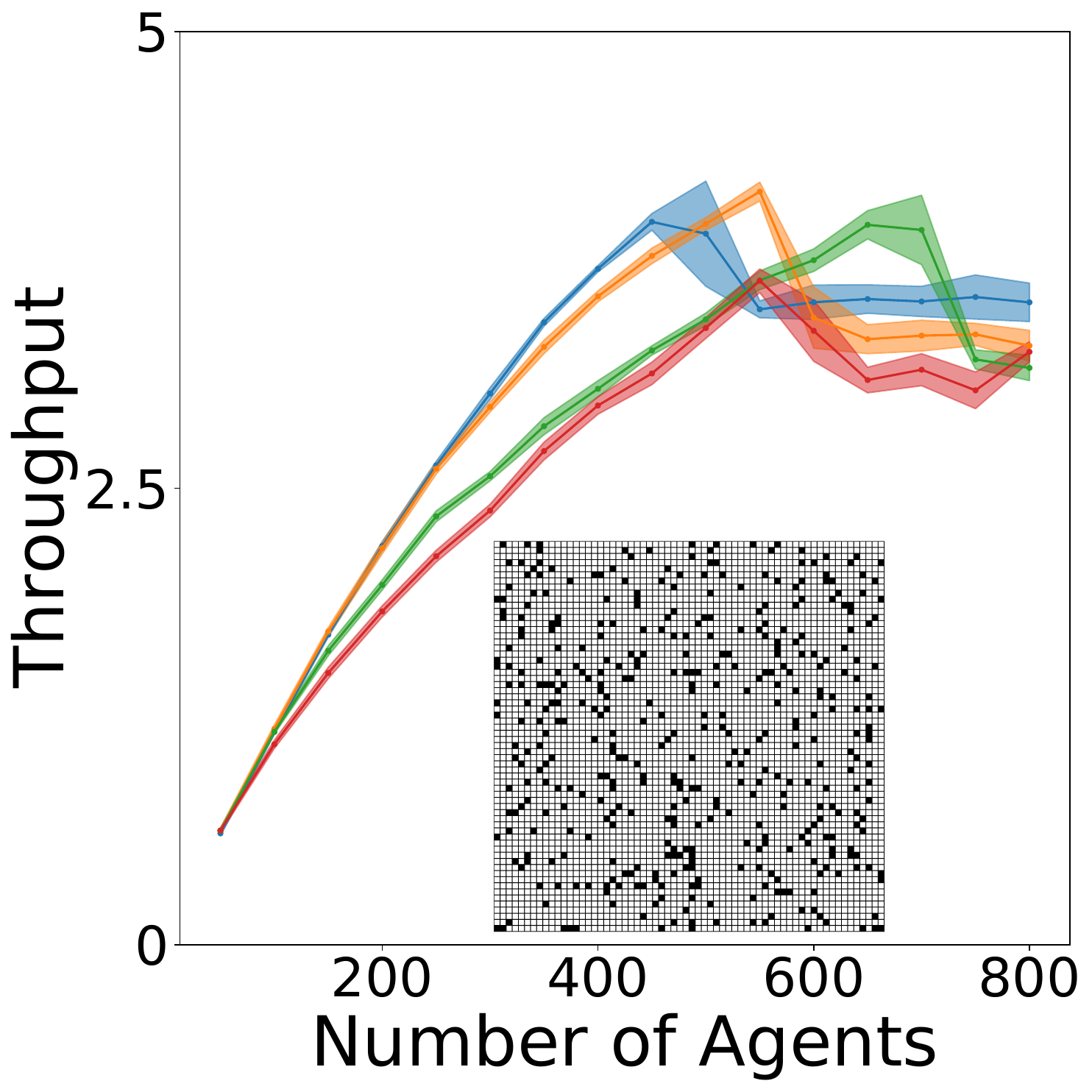}%
        \includegraphics[width=0.5\textwidth]{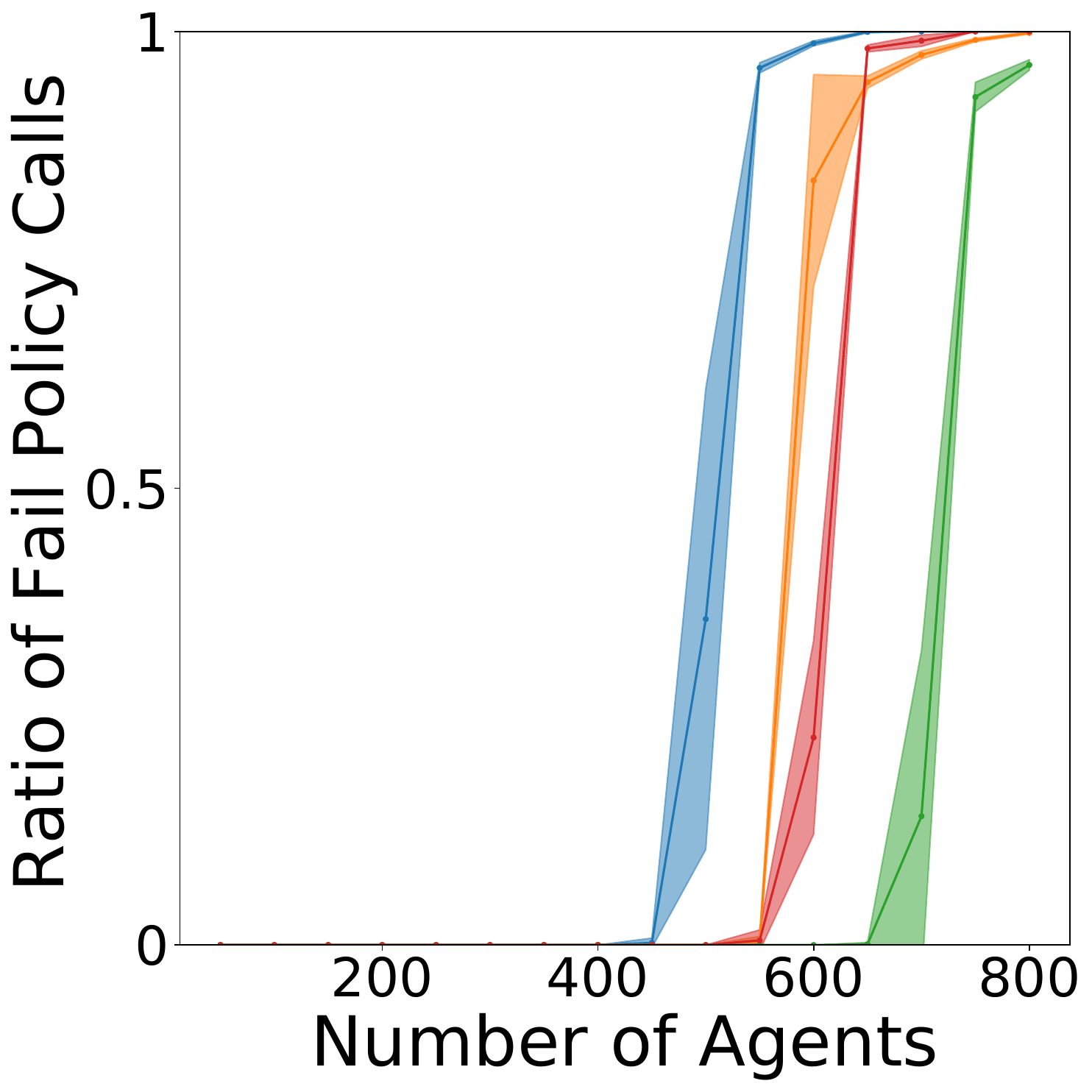}
        \caption{\randomMid}
        \label{fig:sim-w-ab:random-64-64-10}
    \end{subfigure}%
    \hfill
    \begin{subfigure}{0.33\textwidth}
        \centering
        \includegraphics[width=0.5\textwidth]{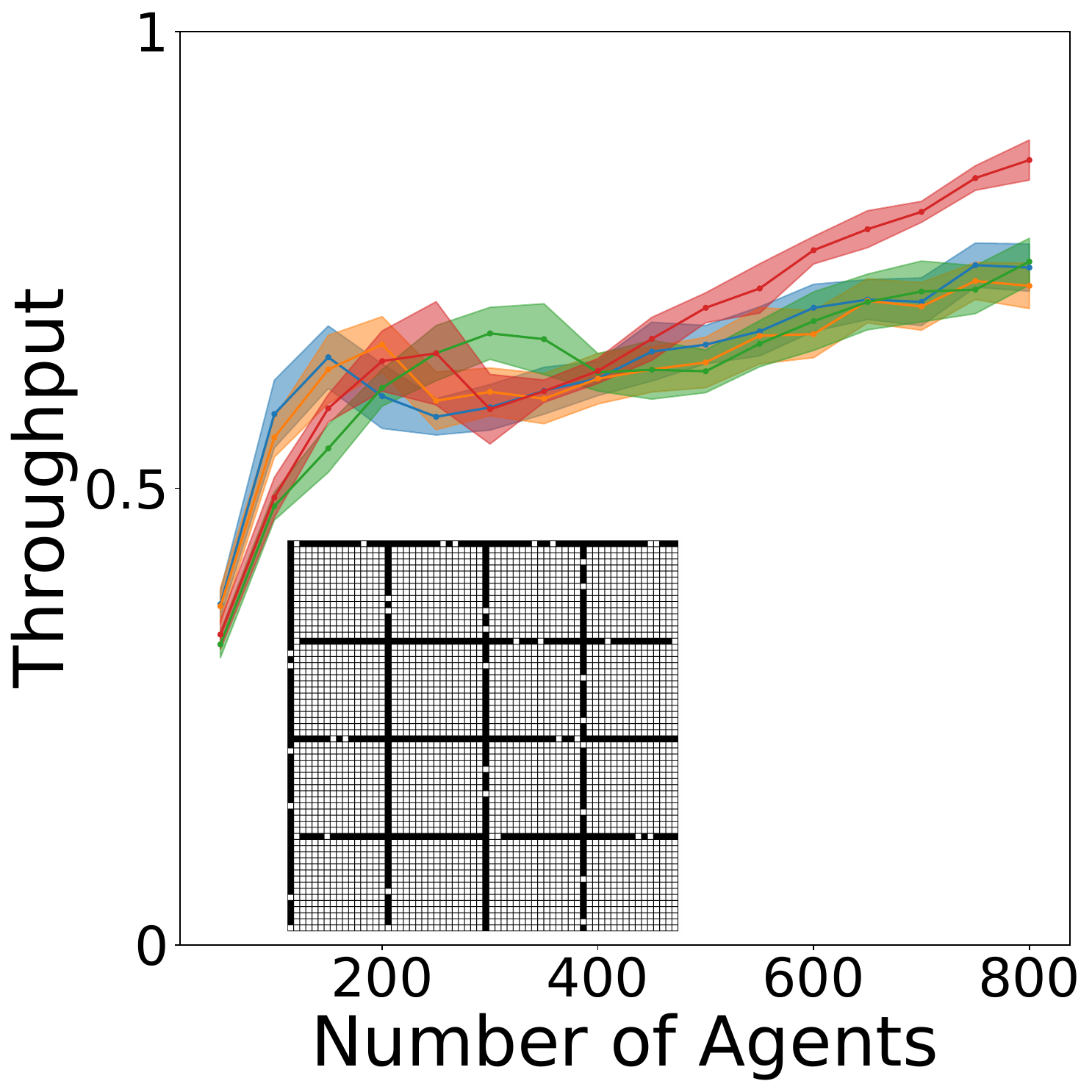}%
        \includegraphics[width=0.5\textwidth]{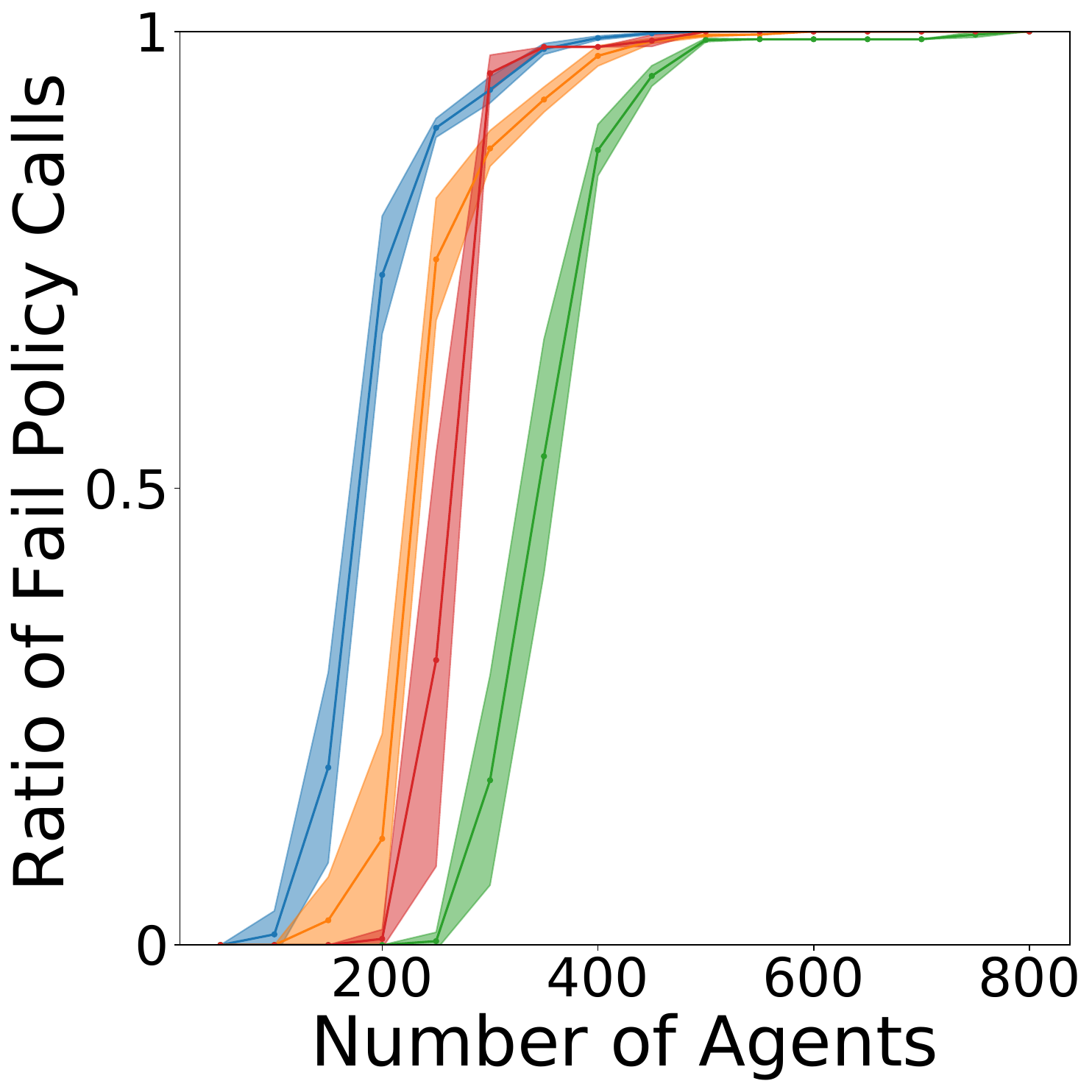}
        \caption{\roomMid}
        \label{fig:sim-w-ab:room-64-64-16}
    \end{subfigure}
    \hfill
    \begin{subfigure}{0.33\textwidth}
        \centering
        \includegraphics[width=0.5\textwidth]{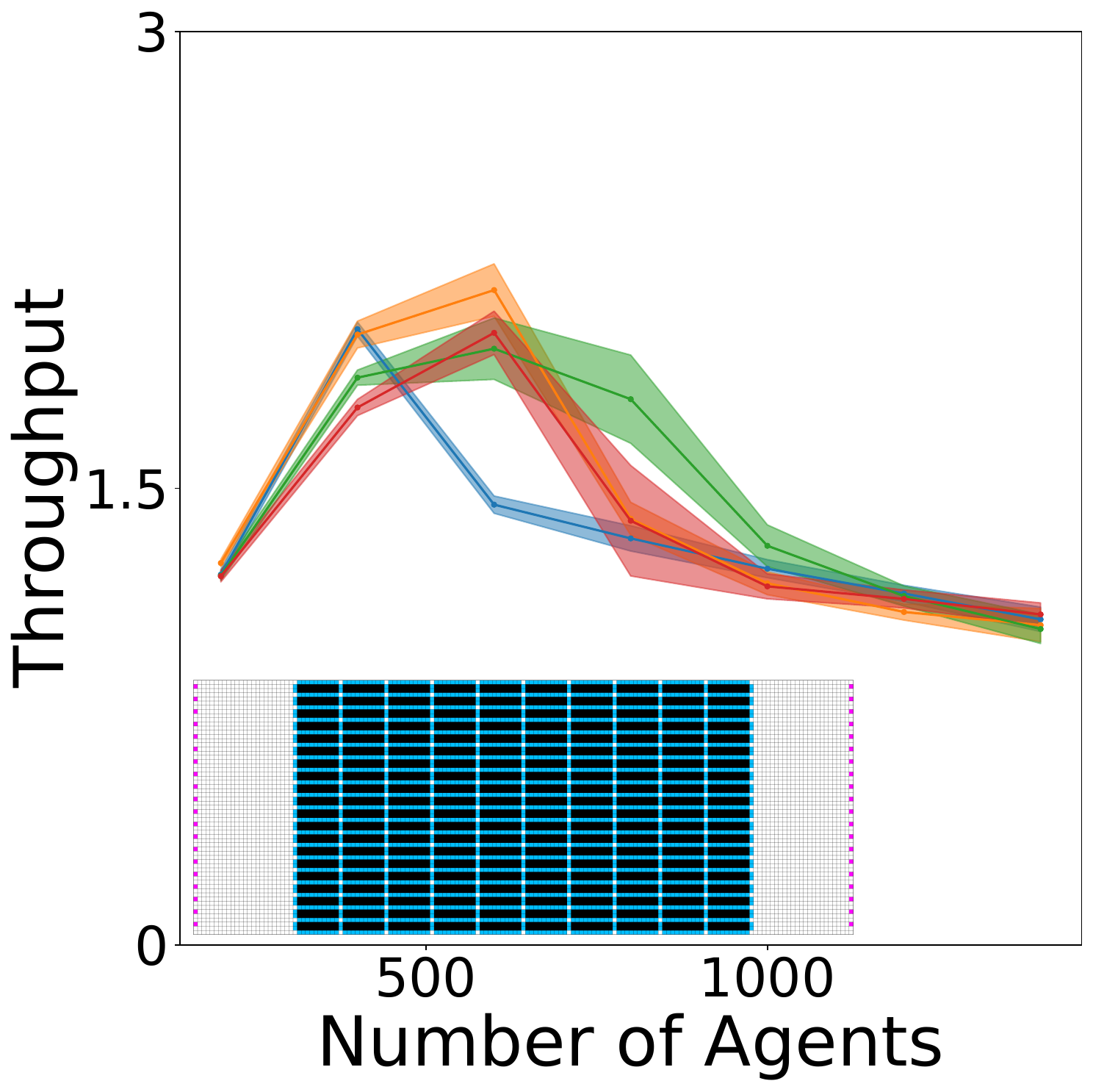}%
        \includegraphics[width=0.5\textwidth]{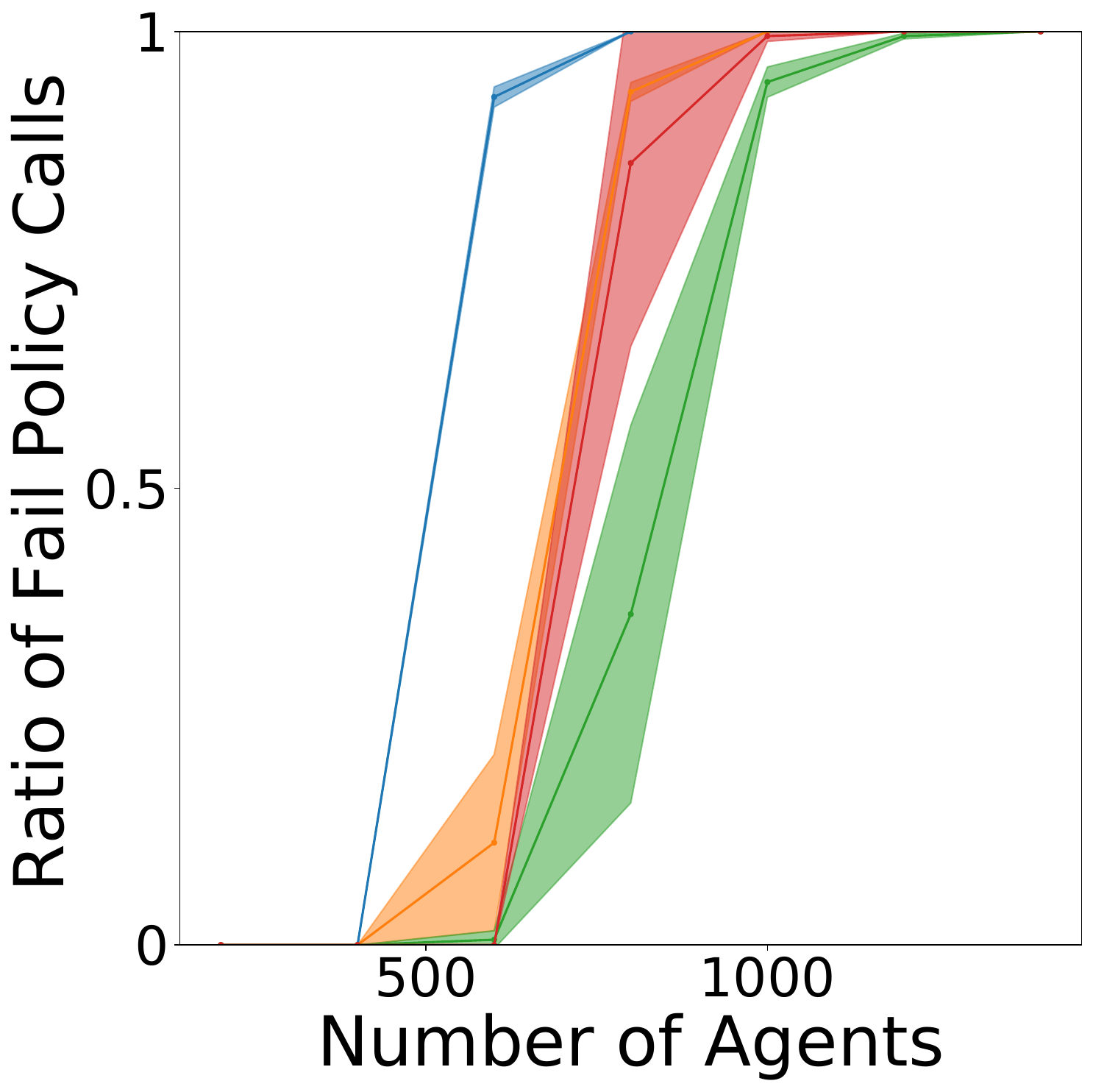}
        \caption{\warehouseLarge}
        \label{fig:sim-w-ab:warehouse-10-20-10-2-1}
    \end{subfigure}


    \caption{Experiment results of $P$ and $W$ (Experiment 3 in \Cref{tab:exp}).}
    \label{fig:sim-w-ab}
\end{figure*}

\begin{figure*}[!t]
    \centering
    \includegraphics[width=.4\textwidth]{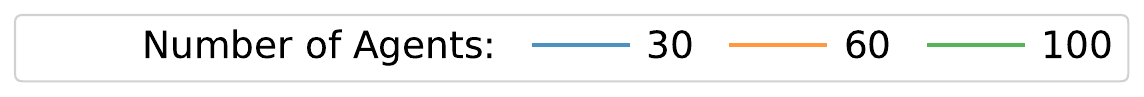}\\
    \begin{subfigure}{0.33\textwidth}
        \centering
        \includegraphics[width=1\textwidth]{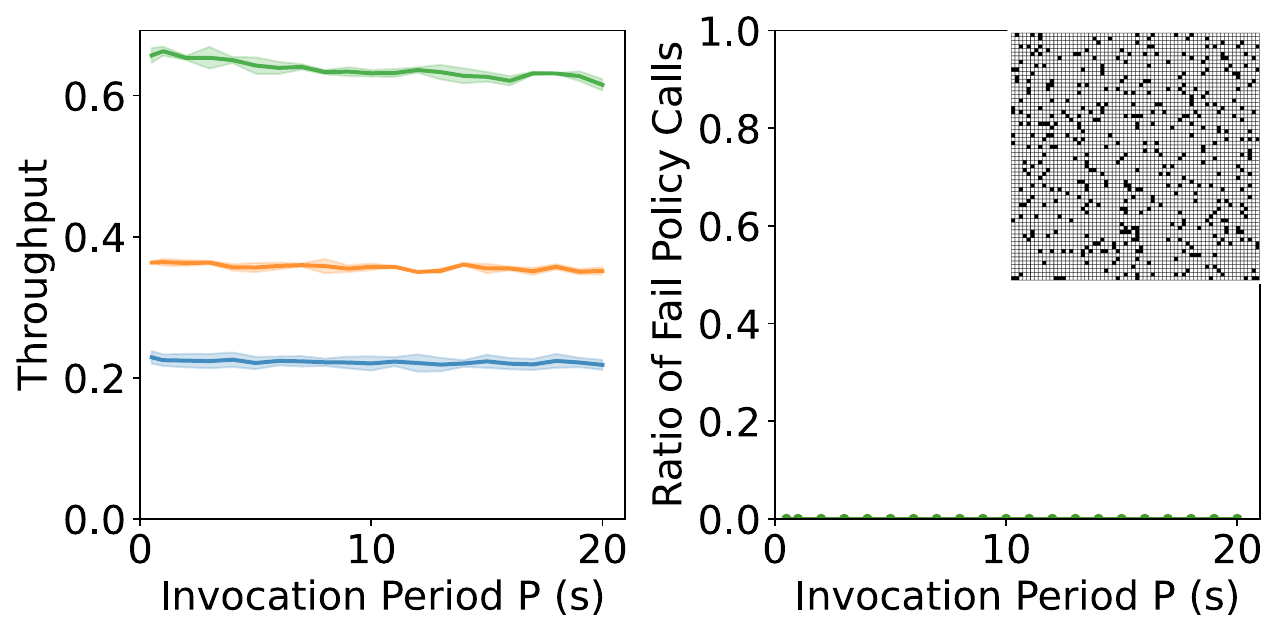}
        \caption{\randomMid}
        \label{fig:planner-replanFreq-ab:random-64-64-10}
    \end{subfigure}%
    \hfill
    \begin{subfigure}{0.33\textwidth}
        \centering
        \includegraphics[width=1\textwidth]{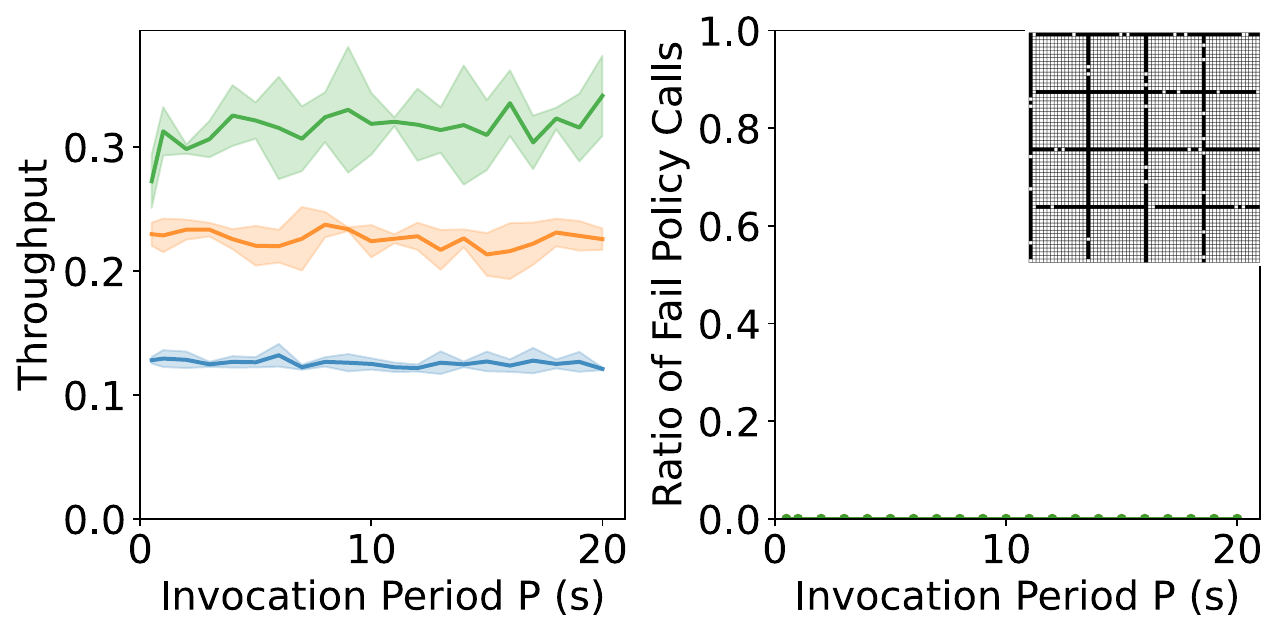}
        \caption{\roomMid}
        \label{fig:planner-replanFreq-ab:room-64-64-16}
    \end{subfigure}%
    \hfill
    \begin{subfigure}{0.33\textwidth}
        \centering
        \includegraphics[width=1\textwidth]{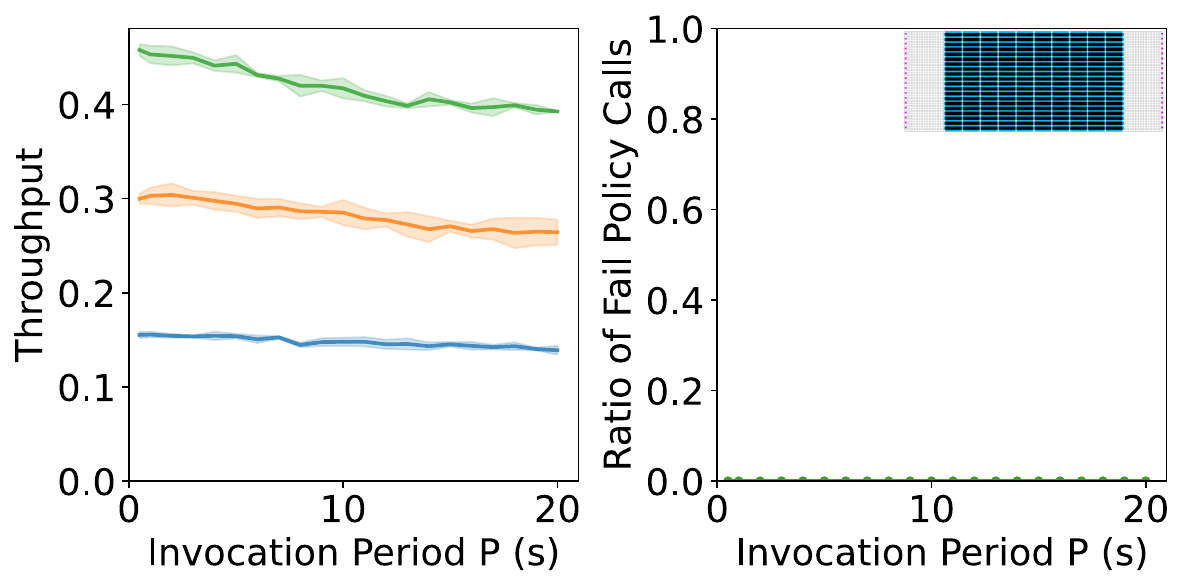}
        \caption{\warehouseLarge}
        \label{fig:planner-replanFreq-ab:den312d}
    \end{subfigure}%




    \caption{Experiment results of different $P$ with standard MAPF (Experiment 4 in \Cref{tab:exp}).}
    \label{fig:planner-replanFreq-ab}
\end{figure*}

\subsubsection{Experiment Result}

\Cref{fig:plan-invoke-ab} shows the results of experiment 2 in \Cref{tab:exp}. In \roomMid and \warehouseLarge, both policies have similar performance with the same values of $W$, potentially due to the topology of the maps.
\roomMid contains large open areas connected by narrow entry points or corridors. In contrast, \warehouseLarge is sufficiently spacious that congestion rarely occurs, allowing agents to follow their shortest paths regardless of the invocation policy, which results in little performance difference among the policies.
In \randomMid, interestingly, the advantage varies among different values of $W$. When $W = 1$ second, the planner makes myopic decisions 1 second in the future, so planning more frequently with the periodic policy results in slightly better throughput. However, when $W=10$, the planner makes long-term decisions, making replanning frequently less desirable. Therefore, sticking to the planned paths with the \noaction policy results in better throughput than the periodic policy. \Cref{fig:plan-invoke-ab-add} in Appendix~\ref{appen:exp-add:invoke} shows additional results.

\Cref{fig:sim-w-ab} shows the result of experiment 3. Across all maps, invoking the planner more frequently and planning for shorter horizons results in better throughput with a smaller number of agents. This conforms with the results obtained by \citet{VaramballySoCS22}. However, with a larger number of agents, invoking the planner less frequently while planning for longer horizons is more beneficial. This serves as an updated result of \citet{VaramballySoCS22}. \Cref{fig:sim-w-ab-add} in Appendix~\ref{appen:exp-add:invoke} shows additional results.

\Cref{fig:planner-replanFreq-ab} shows the result of experiment 4.
Decreasing $P$ improves throughput on \randomMid and \warehouseLarge at high agent densities. However, the advantage of a higher replanning frequency does not hold consistently in \roomMid or at lower densities.
This behavior stems from the execution–planning mismatch inherent in FMS. Before each replanning step, the system computes a commit cut to determine the expected start states, and the planner assumes synchronous planning from these states. In practice, however, the committed cuts estimated from the ADG are never perfectly aligned with the agents’ actual progress. As a result, more frequent replanning does not necessarily resynchronize the system and may even accumulate timing inconsistencies over time.
This highlights a key challenge in deploying MAPF within FMS: accurately estimating the commit cut to ensure proper resynchronization at each replanning step. Additional results are provided in \Cref{fig:planner-replanFreq-ab-add} in Appendix~\ref{appen:exp-add:invoke}.

\subsection{Fail Policy} \label{sec:exp:subsec:fail}
\subsubsection{Experiment Setup}
We perform experiment 5 in \Cref{tab:exp} to compare the following fail policies: PIBT, Guided PIBT, and \lrgw. We do not include the all wait fail policy because experiments in \citet{morag_adapting_2023} have proven its bad performance.

\begin{figure*}[!t]
    \centering
    \includegraphics[width=1\textwidth]{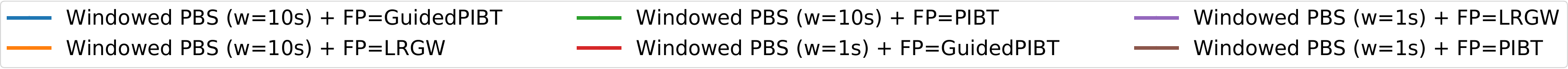}

    \begin{subfigure}{0.33\textwidth}
        \centering
        \includegraphics[width=0.5\textwidth]{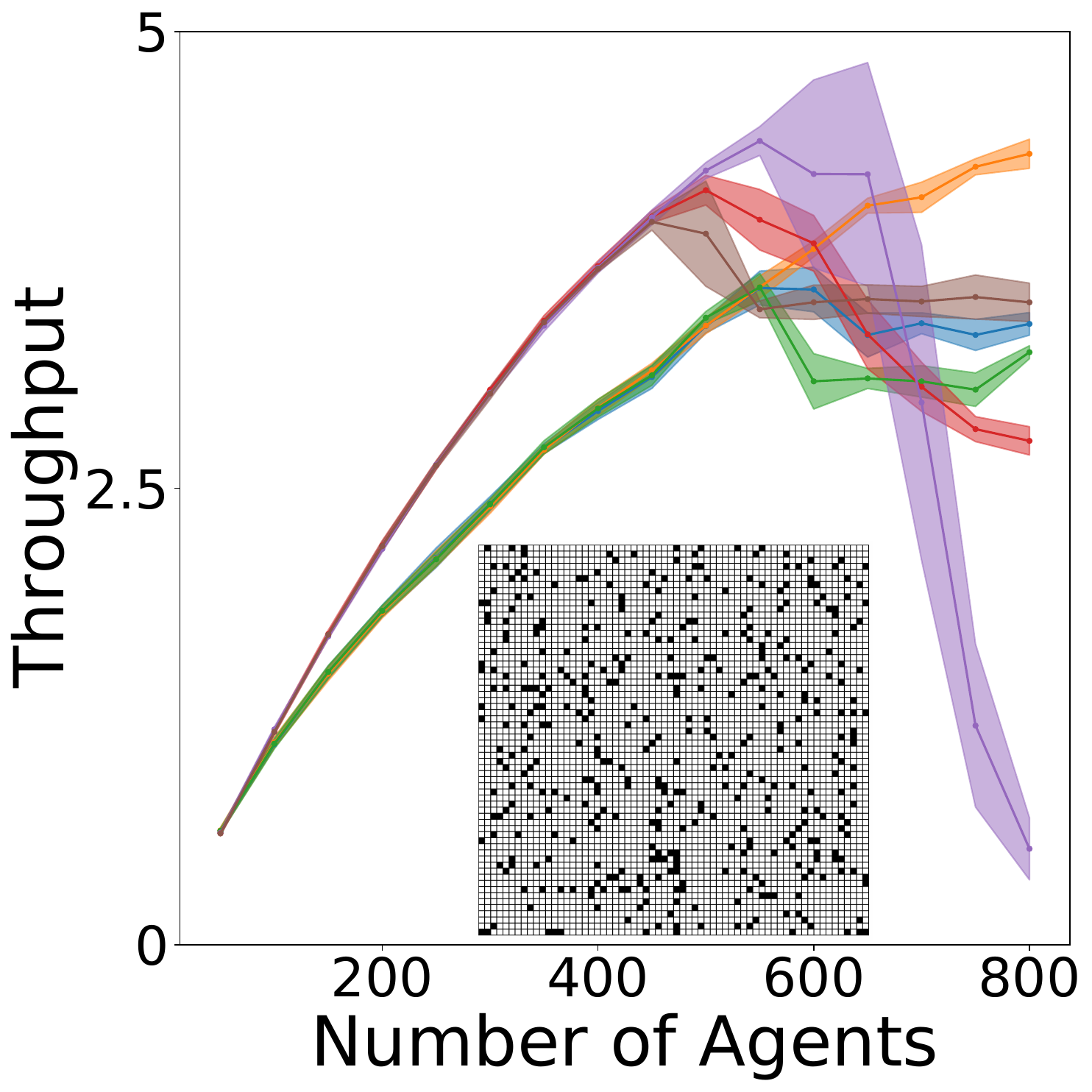}%
        \includegraphics[width=0.5\textwidth]{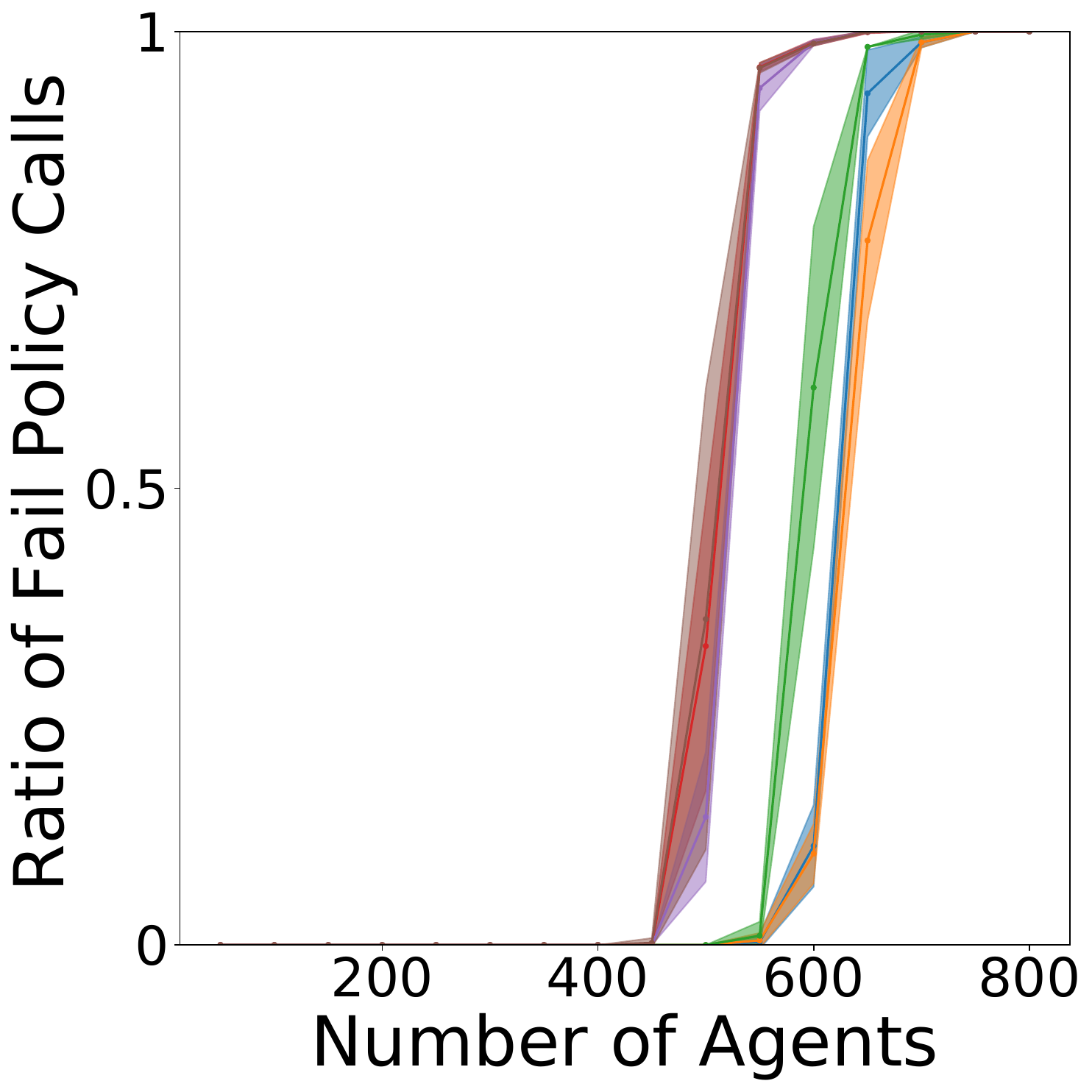}
        \caption{\randomMid}
        \label{fig:fail-policy-ab:random-64-64-10}
    \end{subfigure}%
    \hfill
    \begin{subfigure}{0.33\textwidth}
        \centering
        \includegraphics[width=0.5\textwidth]{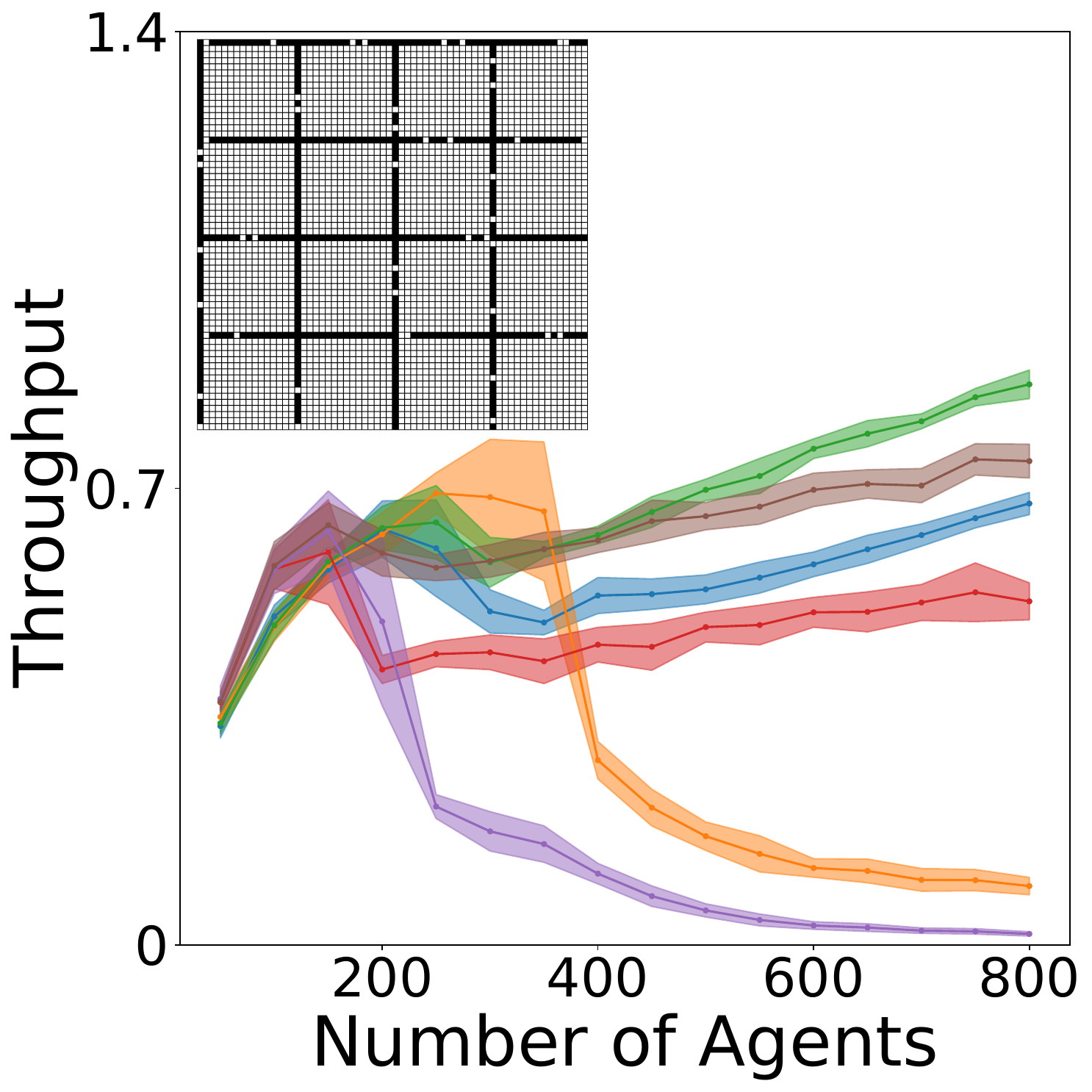}%
        \includegraphics[width=0.5\textwidth]{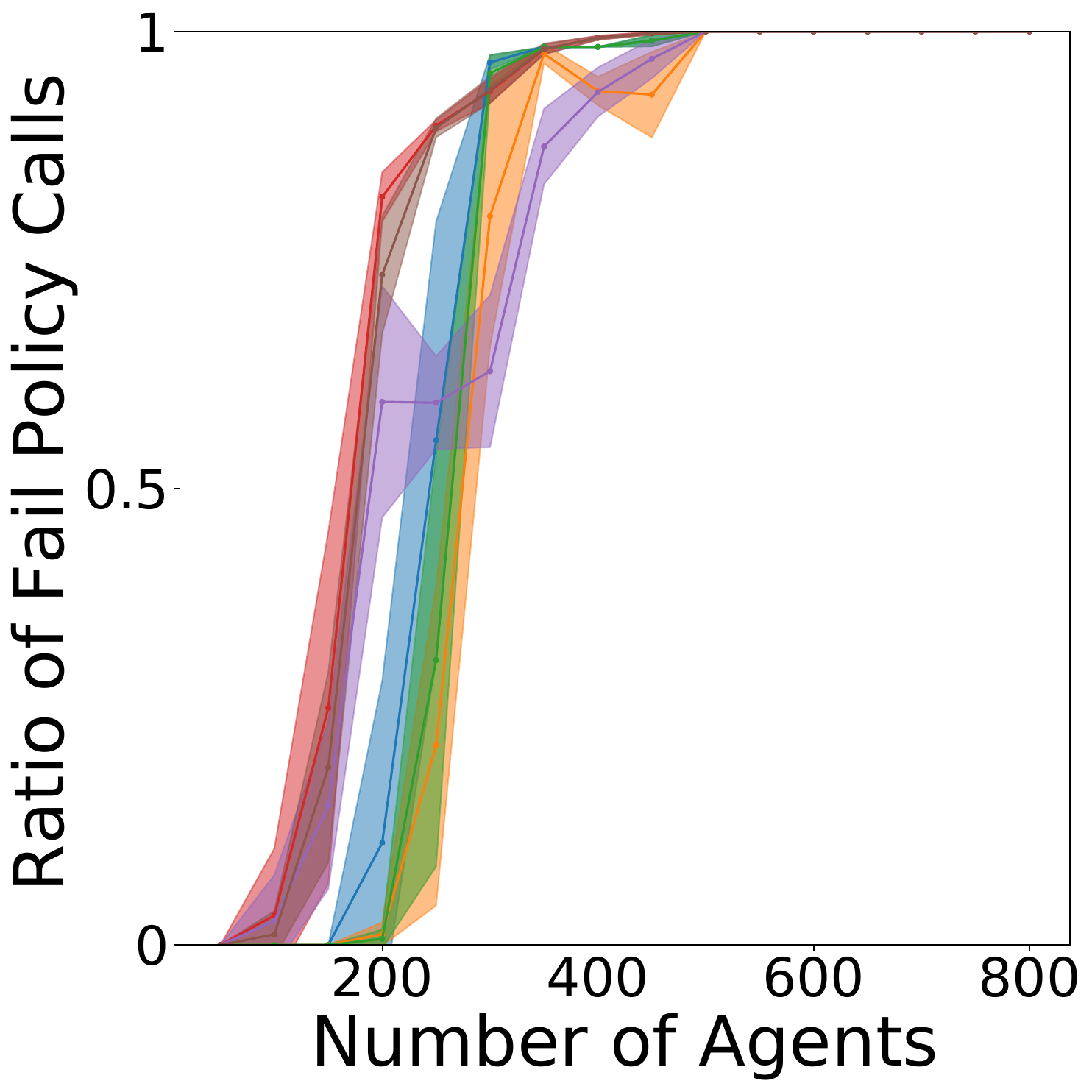}
        \caption{\roomMid}
        \label{fig:fail-policy-ab:room-64-64-16}
    \end{subfigure}
    \hfill
    \begin{subfigure}{0.33\textwidth}
        \centering
        \includegraphics[width=0.5\textwidth]{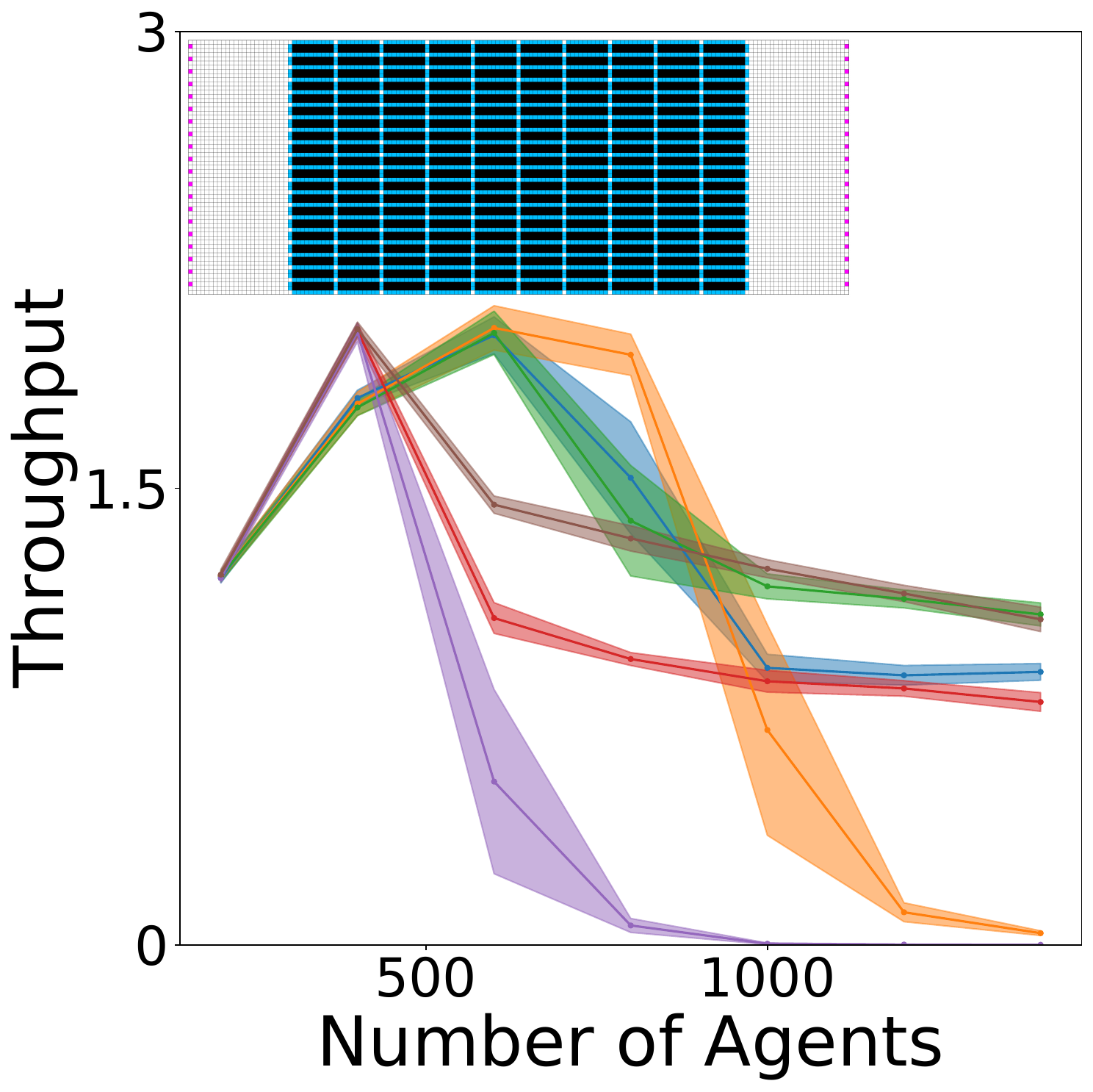}%
        \includegraphics[width=0.5\textwidth]{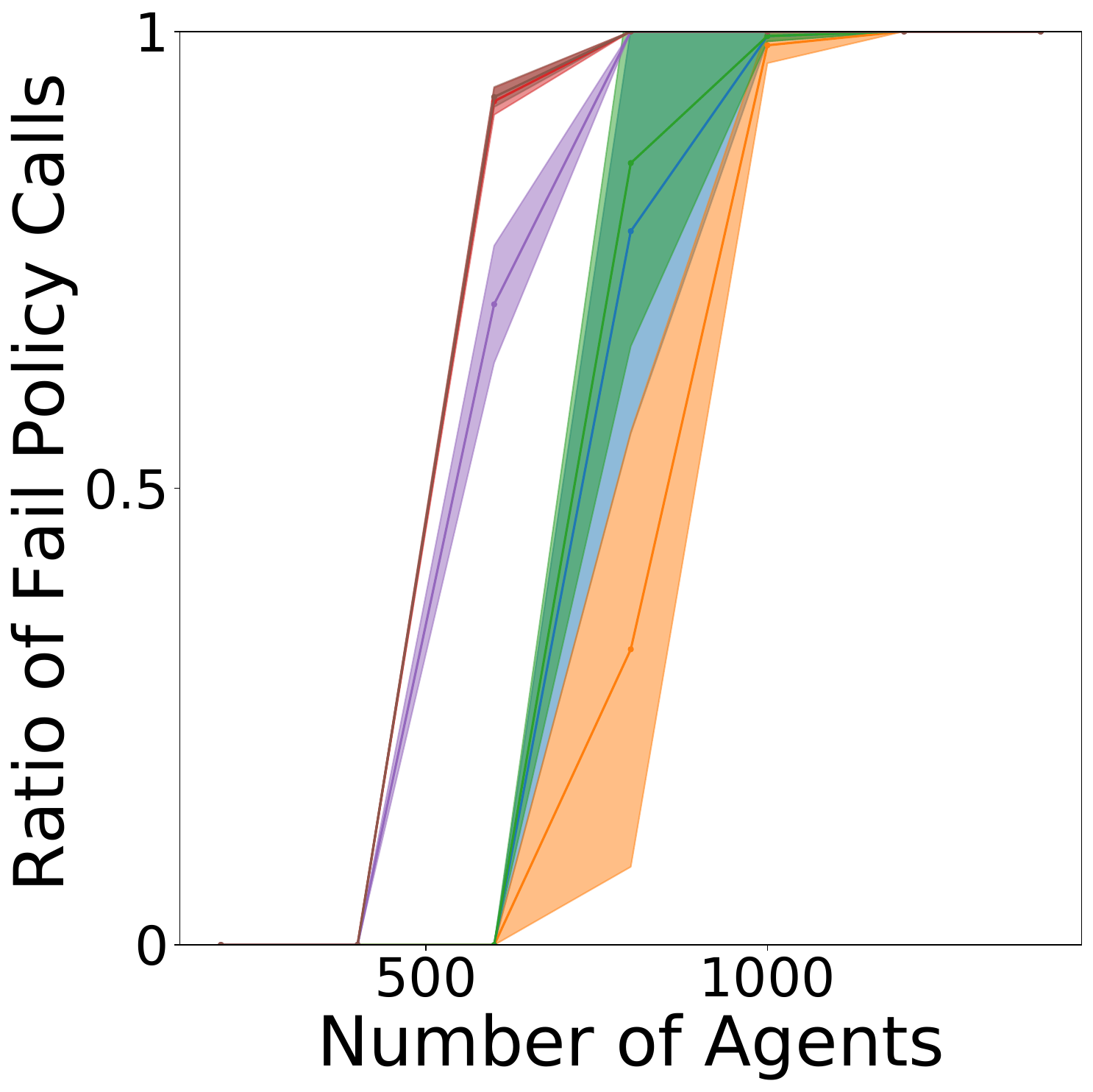}
        \caption{\warehouseLarge}
        \label{fig:fail-policy-ab:warehouse-10-20-10-2-1}
    \end{subfigure}

    \caption{Experiment results of fail policies (Experiment 5 in \Cref{tab:exp}).}
    \label{fig:fail-policy-ab}
\end{figure*}

\begin{figure*}[!t]
    \centering
    \includegraphics[width=.5\textwidth]{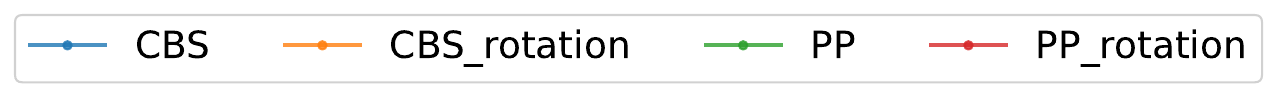}


    \begin{subfigure}{0.33\textwidth}
        \centering
        \includegraphics[width=1\textwidth]{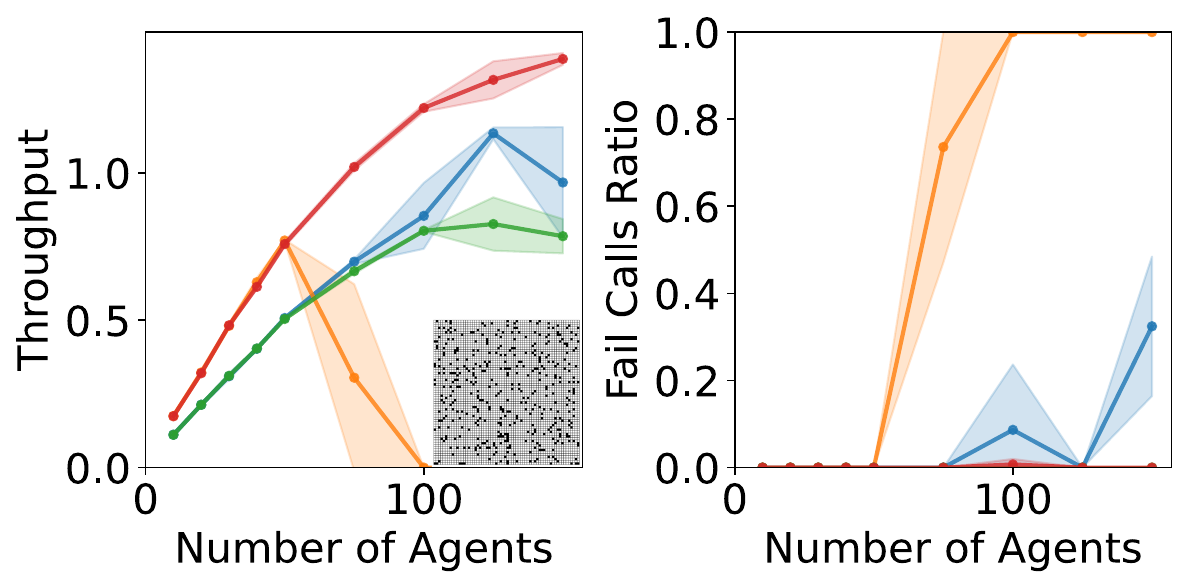}
        \caption{\randomMid}
        \label{fig:planner-modelAccOptimality-ab:random-64-64-10}
    \end{subfigure}%
    \hfill
    \begin{subfigure}{0.33\textwidth}
        \centering
        \includegraphics[width=1\textwidth]{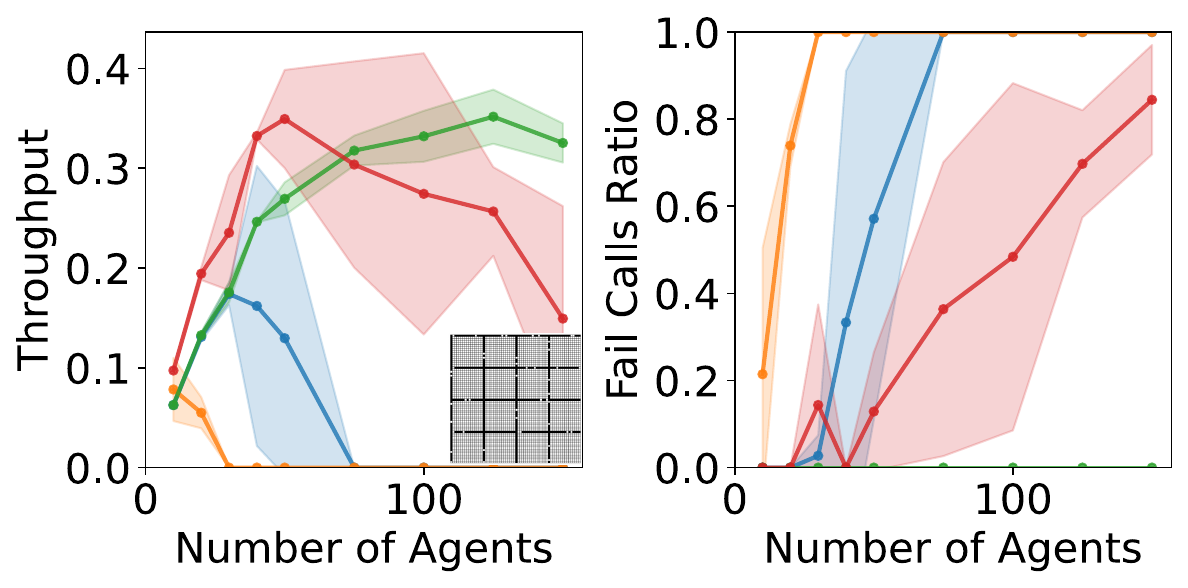}
        \caption{\roomMid}
        \label{fig:planner-modelAccOptimality-ab:room-64-64-16}
    \end{subfigure}%
    \hfill
    \begin{subfigure}{0.33\textwidth}
        \centering
        \includegraphics[width=1\textwidth]{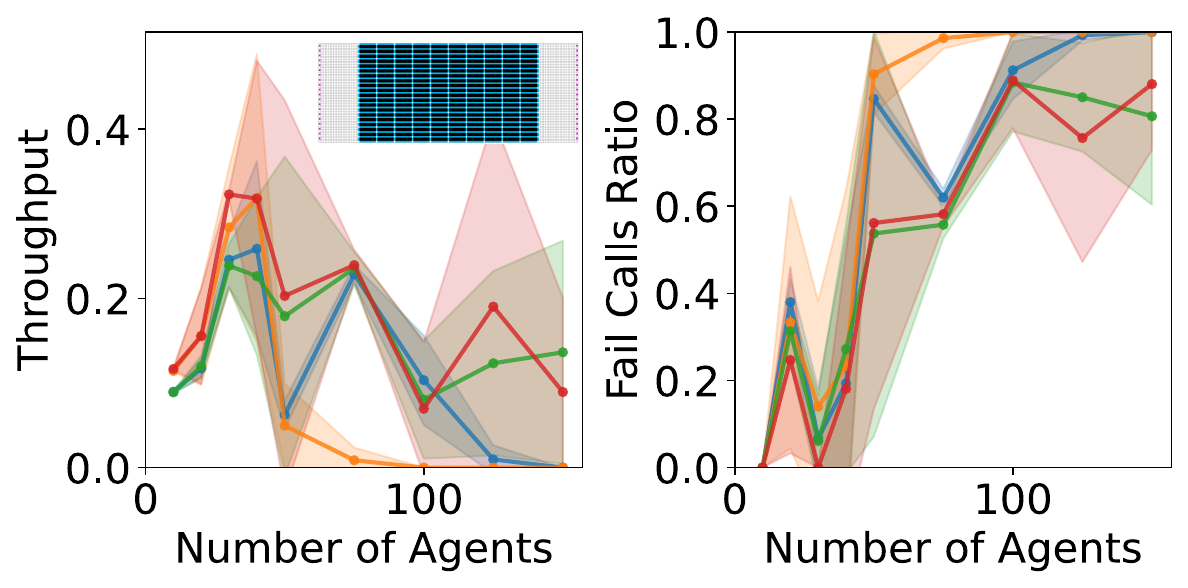}
        \caption{\warehouseLarge}
        \label{fig:planner-modelAccOptimality-ab:warehouse-10-20-10-2-1}
    \end{subfigure}

    \caption{Experiment results of agent model accuracy and planner optimality (Experiment 6 in \Cref{tab:exp}).}
    \label{fig:planner-modelAccOptimality-ab}
\end{figure*}

\subsubsection{Experiment Result}
\Cref{fig:fail-policy-ab} shows the results of experiment 5 in \Cref{tab:exp}. \lrgw and PIBT achieve the best performance on different maps and at different values of $W$. Compared to PIBT, \lrgw is advantageous when the conflicting paths returned by windowed PBS are valuable. For example, in \randomMid, for both values of $W$, \lrgw achieves higher throughput than the other two fail policies when they are invoked with a large number of agents.
This is because using \lrgw to instruct agents to follow such paths is better than replanning from scratch using PIBT, which is known to be a suboptimal and myopic one-step planner. However, when the conflicting paths returned by windowed PBS are of low quality, replanning from scratch using PIBT works better. In \roomMid, for example, we conjecture that the special topology of the map (chunks of empty spaces connected by narrow entry points) results in the conflicting solution of windowed PBS having too many wait actions.
Since \lrgw potentially adds more wait actions to avoid collisions, using it results in traffic congestion. PIBT, on the other hand, tends to move agents instead of having them wait in place, alleviating more congestion than \lrgw. Surprisingly, Guided PIBT fails to be the best fail policy in all setups, meaning that the conflicting paths returned by windowed PBS are low-quality guide paths for it. \Cref{fig:fail-policy-ab-add} in Appendix~\ref{appen:exp-add:FP} shows additional results.

\subsection{Optimality and Robot Model Accuracy} \label{sec:exp:subsec:planner}
\subsubsection{Experiment Setup}
Planning with more accurate agent models and more optimal solvers leads to better solution quality, but both lead to a longer planning runtime~\cite{yan2025bridging}.
Therefore, we conduct experiment 6 in \Cref{tab:exp} to study the trade-off between optimality and model accuracy. We pair the pebble motion and rotation motion models with CBS~\cite{LiICAPS20,zhang2023efficient}, a slow optimal solver, and PP~\cite{erdmann1987multiple}, a fast suboptimal solver, to set up the experiment.

\paragraph{Experiment Result}
As shown in \Cref{fig:planner-modelAccOptimality-ab}, if all planners never invoke the fail policy, the planners with more accurate models always achieve better throughput. However, when we compare optimal and suboptimal planners, their solution quality is close when the instances can be solved by both, as prominent in \randomMid. However, both more accurate models and stronger optimality guarantees degrade the planners' scalability. The results in \roomMid and \warehouseLarge\  evidently demonstrate such tradeoffs between planners' solution quality and scalability. \Cref{fig:planner-modelAccOptimality-ab} in Appendix~\ref{appen:exp-add:planner} shows additional results. Additional comparisons of planner optimality under the same agent model are provided in Appendix~\ref{appen:planner-opt}.



\section{Conclusion}

We present \lsmartpro, the first open-source simulator capable of evaluating any MAPF algorithms in FMS that consider kinodynamics constraints, communication delays, and execution uncertainties. \lsmartpro also considers a number of design choices, including (1) MAPF planners, (2) instance generators, (3) planner invocation policies, and (4) fail policies. \lsmartpro synthesizes these design choices as customizable modules, allowing users to conduct experiments to evaluate novel MAPF algorithms in FMS. We conduct empirical comparisons of state-of-the-art solutions in the above design choices, each has been studied in the prior MAPF literature but has not been evaluated under settings as realistic as \lsmartpro. Future work includes adding support for graphs beyond the 4-connected grid in the simulator and for robots with more complex kinodynamics than AGVs.

\section*{Acknowledgments}
This work is in part supported by the National Science Foundation (NSF) under grant numbers \#$2328671$ and \#$2441629$, as well as a gift from Amazon.
This work used Bridge-$2$ at Pittsburgh Supercomputing Center (PSC) through allocation CIS$220115$ from the Advanced Cyberinfrastructure Coordination Ecosystem: Services \& Support (ACCESS) program, which is supported by NSF under grant numbers \#$2138259$, \#$2138286$, \#$2138307$, \#$2137603$, and \#$2138296$.


\bibliography{aaai2026}

\clearpage

\appendix

\section{Logo} \label{appn:logo}

\begin{figure}[!t]
    \centering
    \includegraphics[width=0.8\linewidth]{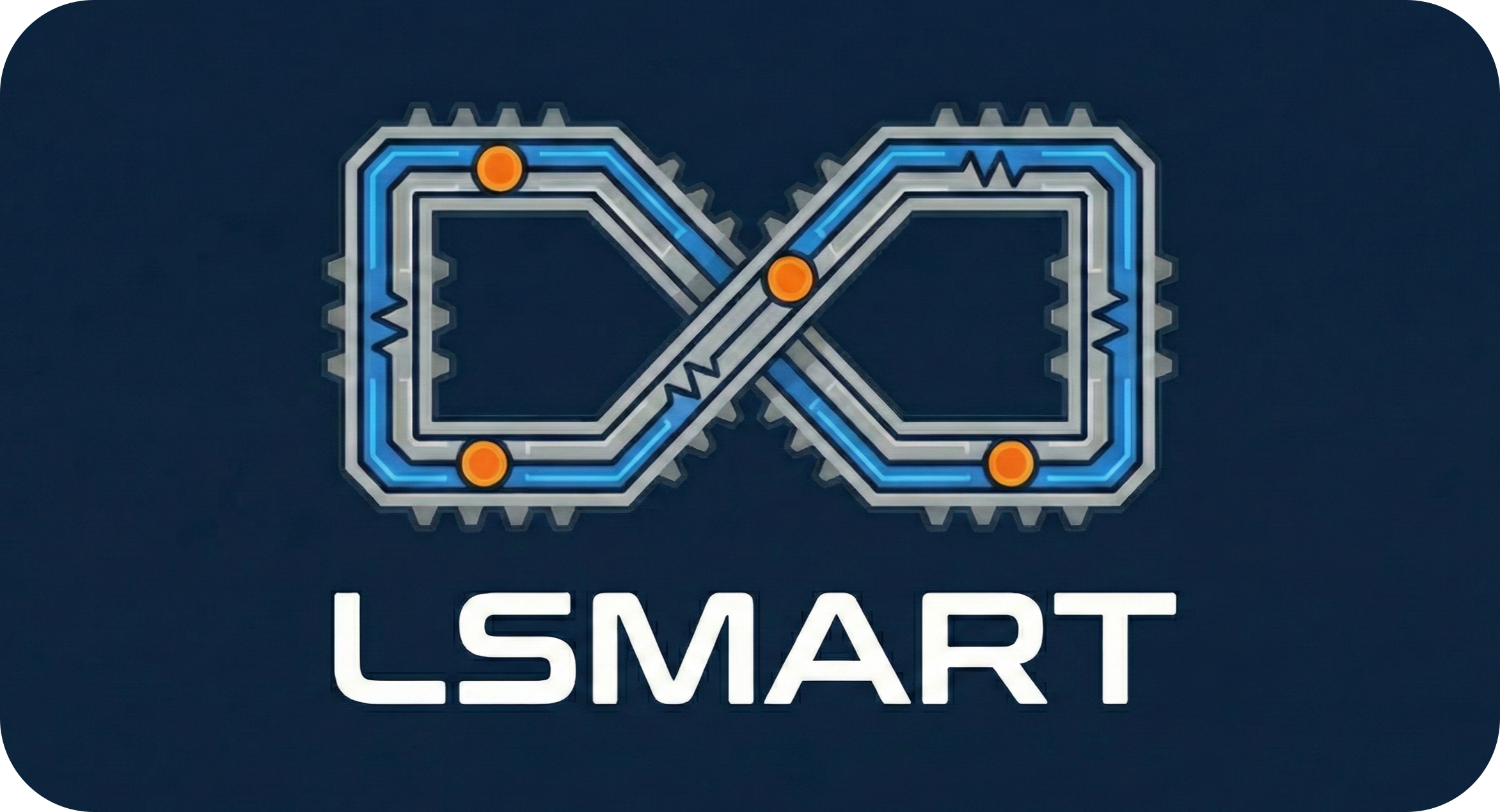}
    \caption{Logo of LSMART.}
    \label{fig:logo}
\end{figure}

\Cref{fig:logo} shows the logo of \lsmartpro\footnote{Generated by Google Gemini 3.}. The logo visualizes the complex nature of a realistic LMAPF system. The central figure—a stylized infinity loop constructed of engineered tracks—represents the \emph{Lifelong} nature of the system, where agents are continuously assigned new goals. The track features mechanical gear teeth and circuit-like resistance lines, symbolizing the realistic kinodynamic constraints, communication delays, and execution uncertainties that \lsmartpro accounts for. The orange agents navigating this rigorous course highlight the core challenge: scalable Multi-Agent coordination within a physically constrained environment.

\section{Additional Results} \label{appen:exp-add}

In this section, we present additional experiment results to augment \Cref{sec:exp}. For experiments outlined in \Cref{tab:exp}, we show results in additional maps, including \warehouseSmall, \mazeSmall, and \emptySmall. 

\subsection{Instance Generator} \label{appen:exp-add:IG}

\begin{figure*}[!t]
    \centering
    \includegraphics[width=0.9\textwidth]{figs/task_assigner_ablation/ta_ablation_legend.pdf}

    \begin{subfigure}{0.33\textwidth}
        \centering
        \includegraphics[width=0.5\textwidth]{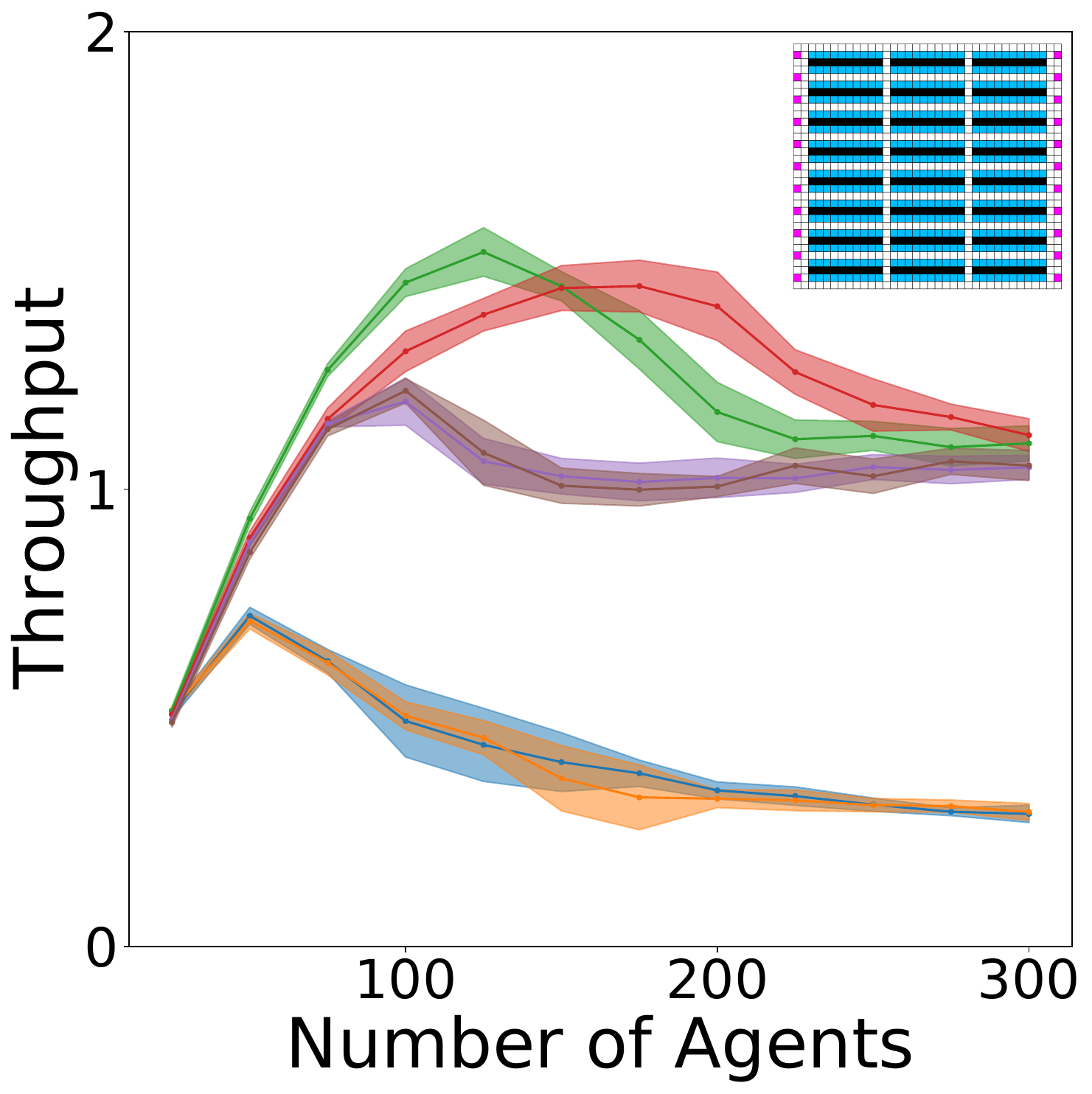}%
        \includegraphics[width=0.5\textwidth]{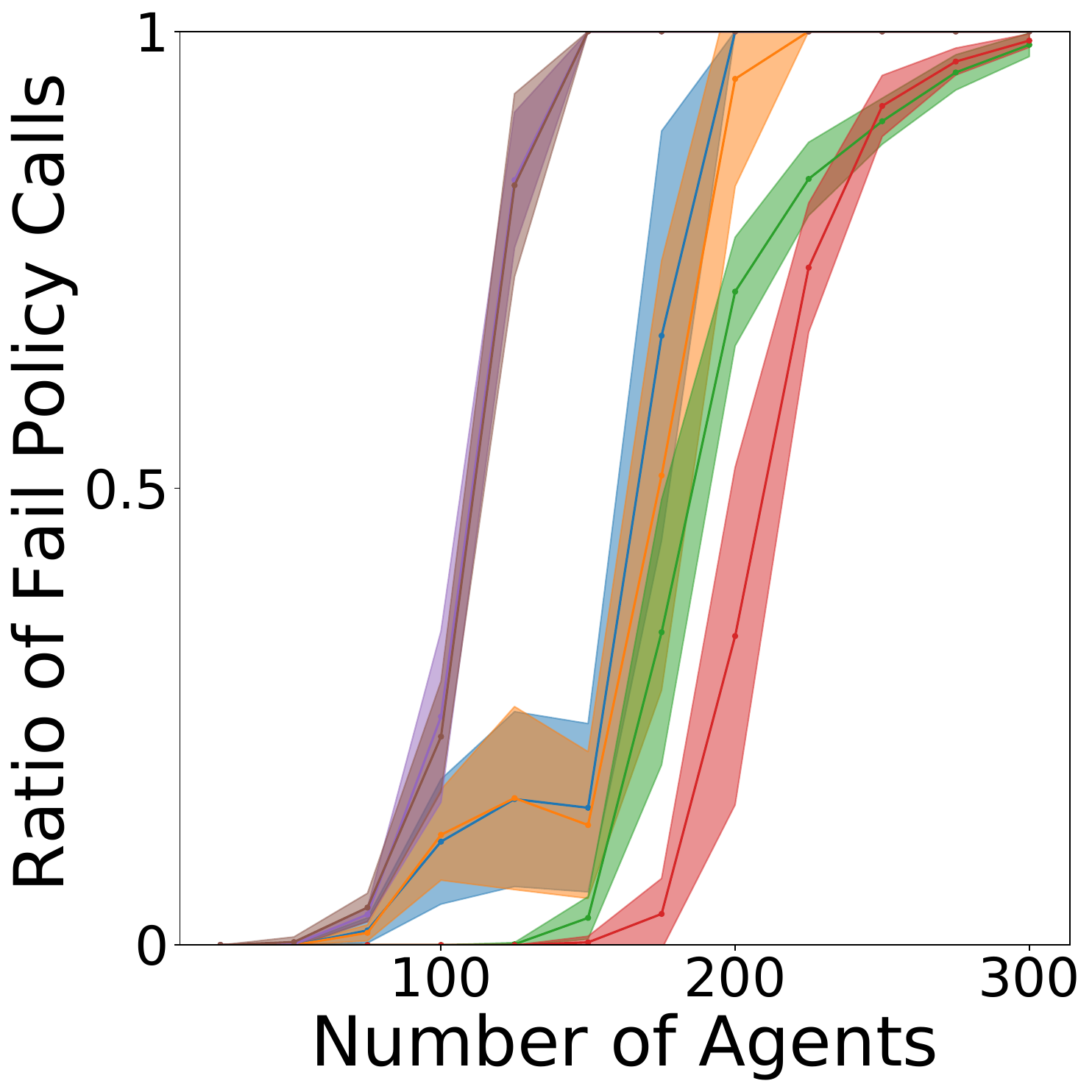}
        \caption{\warehouseSmall}
        \label{fig:ta-ab:warehouse-33-36}
    \end{subfigure}%
    \hfill
    \begin{subfigure}{0.33\textwidth}
        \centering
        \includegraphics[width=0.5\textwidth]{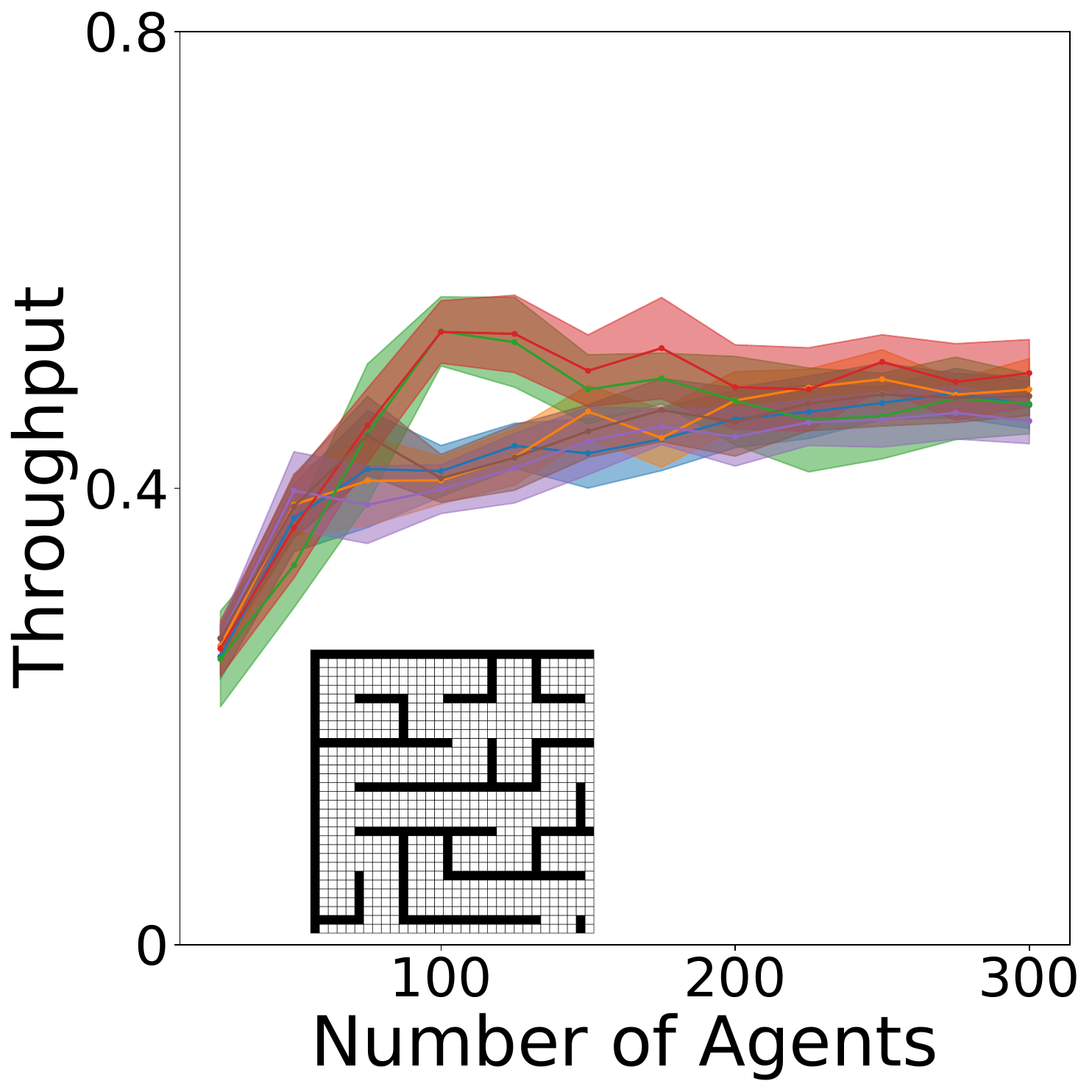}%
        \includegraphics[width=0.5\textwidth]{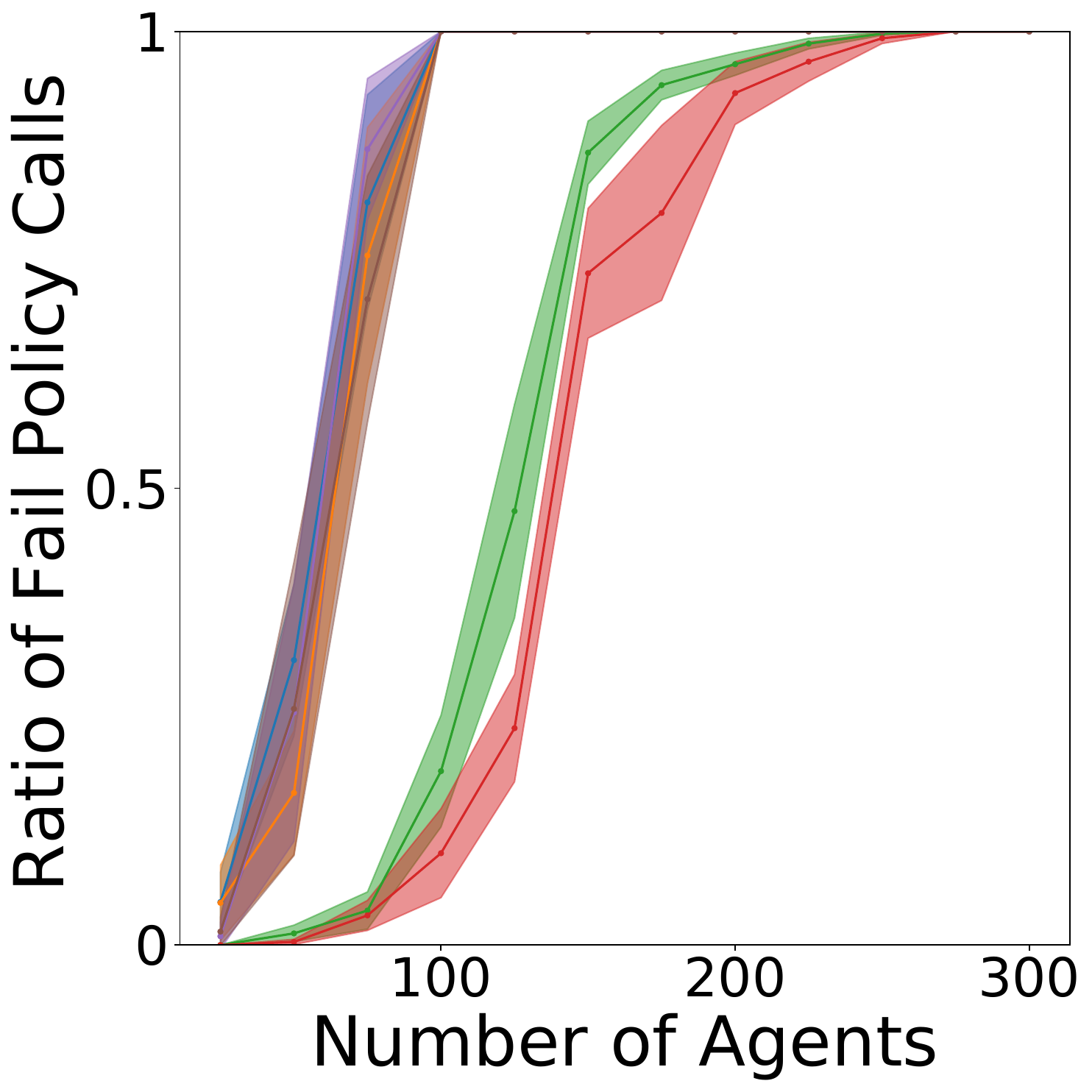}
        \caption{\mazeSmall}
        \label{fig:ta-ab:maze-32-32-4}
    \end{subfigure}%
    \hfill
    \begin{subfigure}{0.33\textwidth}
        \centering
        \includegraphics[width=0.5\textwidth]{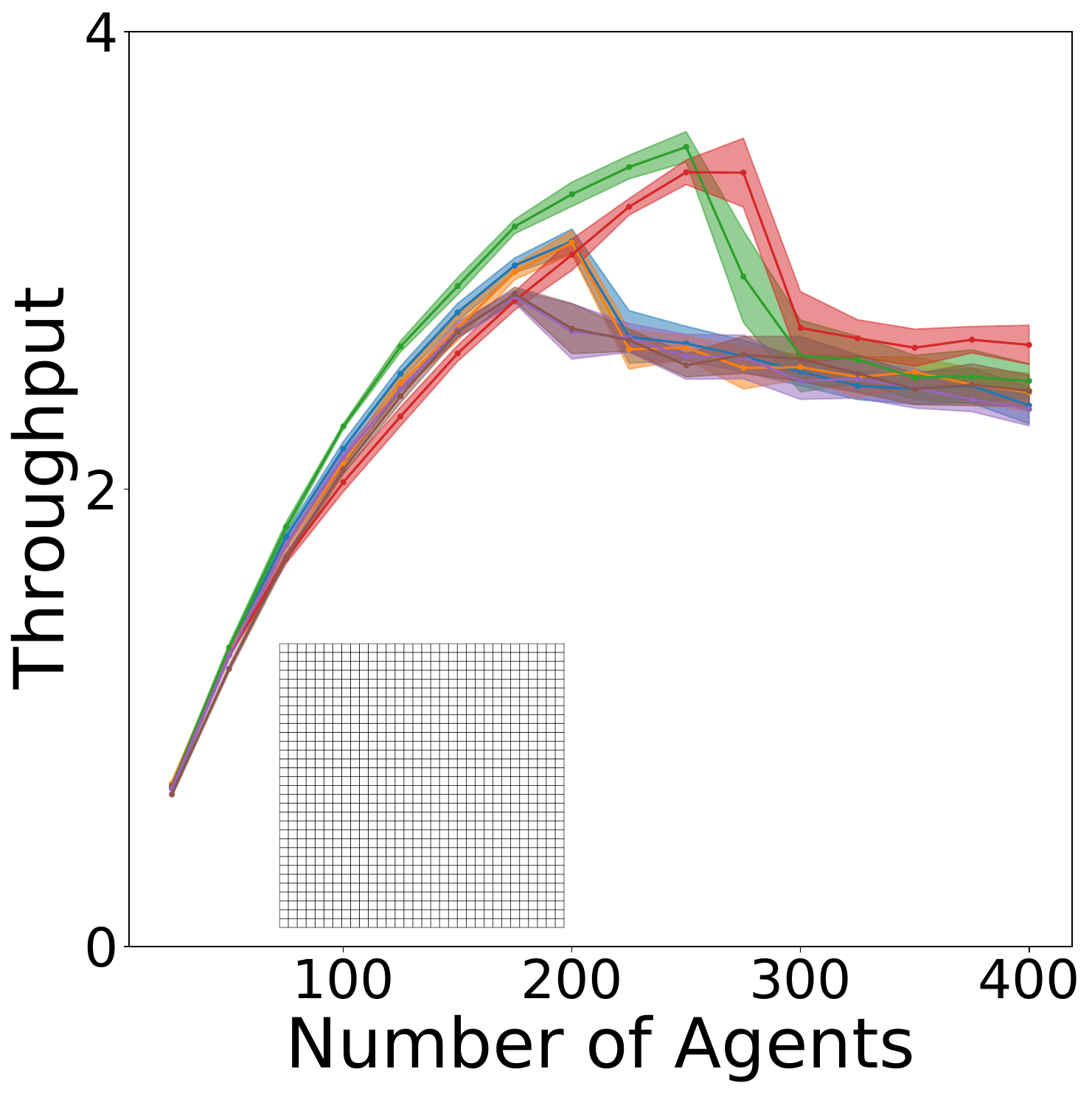}%
        \includegraphics[width=0.5\textwidth]{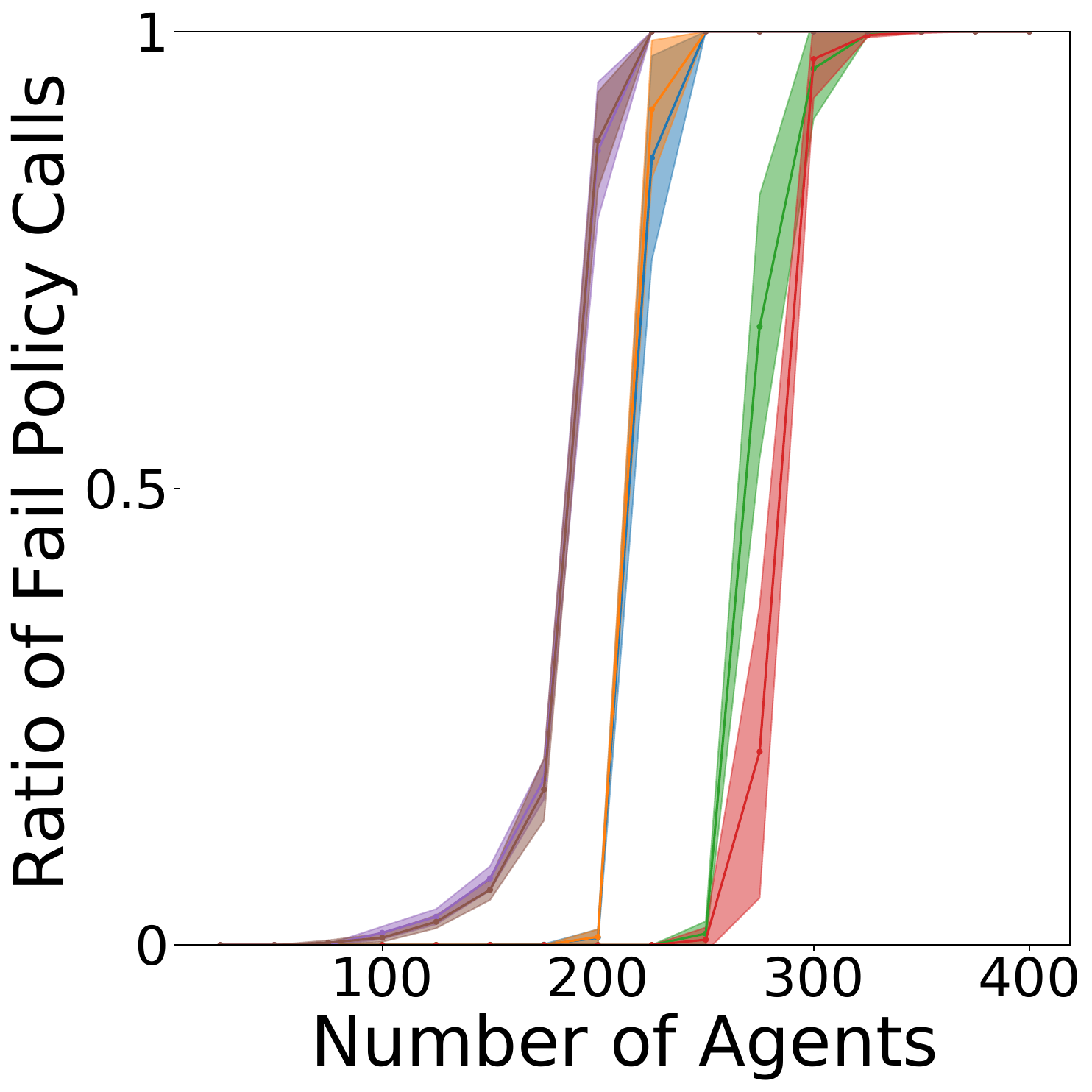}
        \caption{\emptySmall}
        \label{fig:ta-ab:empty-32-32}
    \end{subfigure}%

    \caption{Additional experiment results of instance generators (Experiment 1 in \Cref{tab:exp}).}
    \label{fig:ta-ab-add}
\end{figure*}

\Cref{fig:ta-ab-add} shows the additional experimental results of experiment 1 in \Cref{tab:exp}, comparing different instance generators.

\subsection{Planner Invocation Policies} \label{appen:exp-add:invoke}

\begin{figure*}[!t]
    \centering
    \includegraphics[width=0.8\textwidth]{figs/plan_invoke_ablation/plan_invoke_ablation_legend.pdf}

    \begin{subfigure}{0.33\textwidth}
        \centering
        \includegraphics[width=0.5\textwidth]{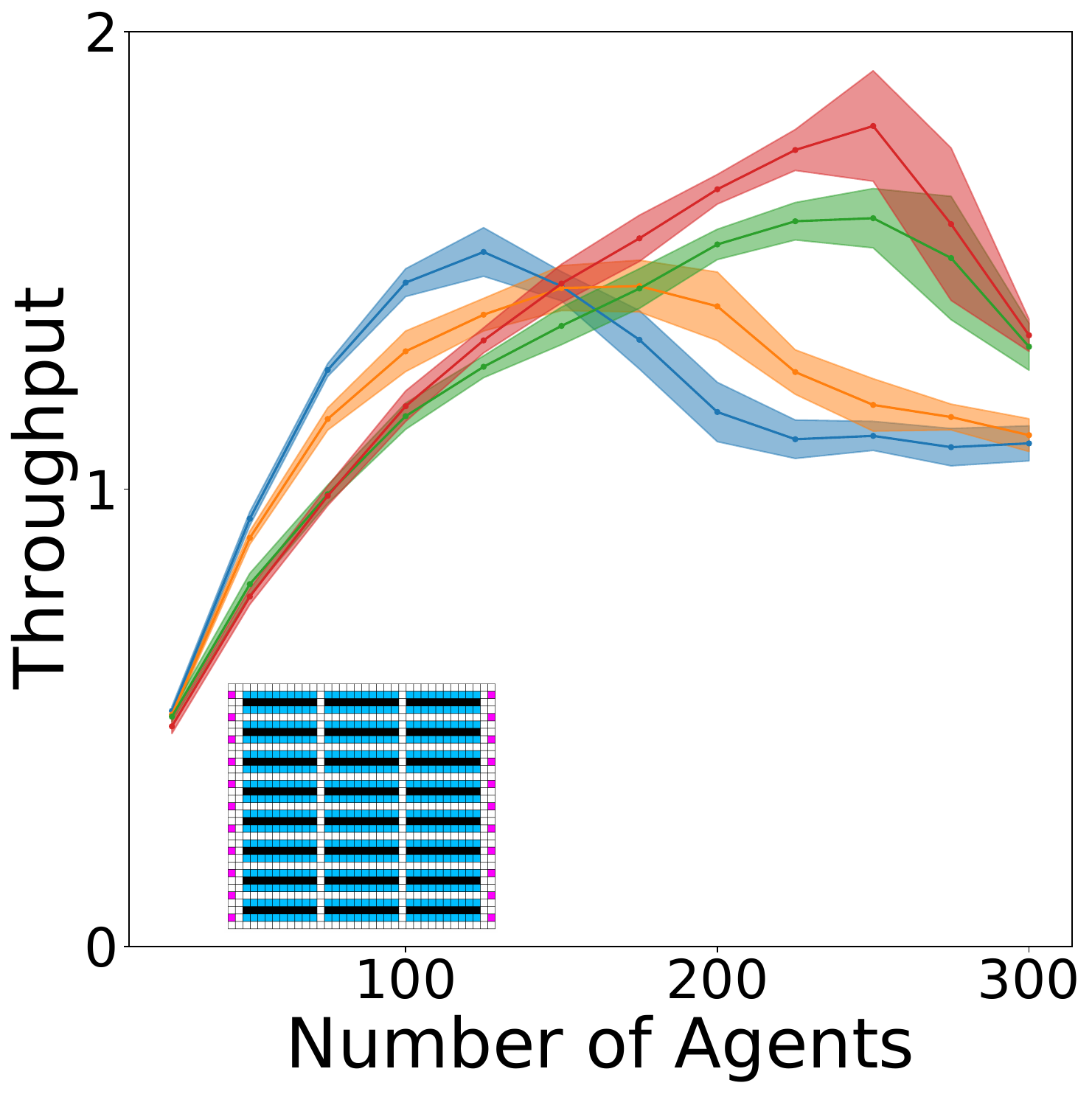}%
        \includegraphics[width=0.5\textwidth]{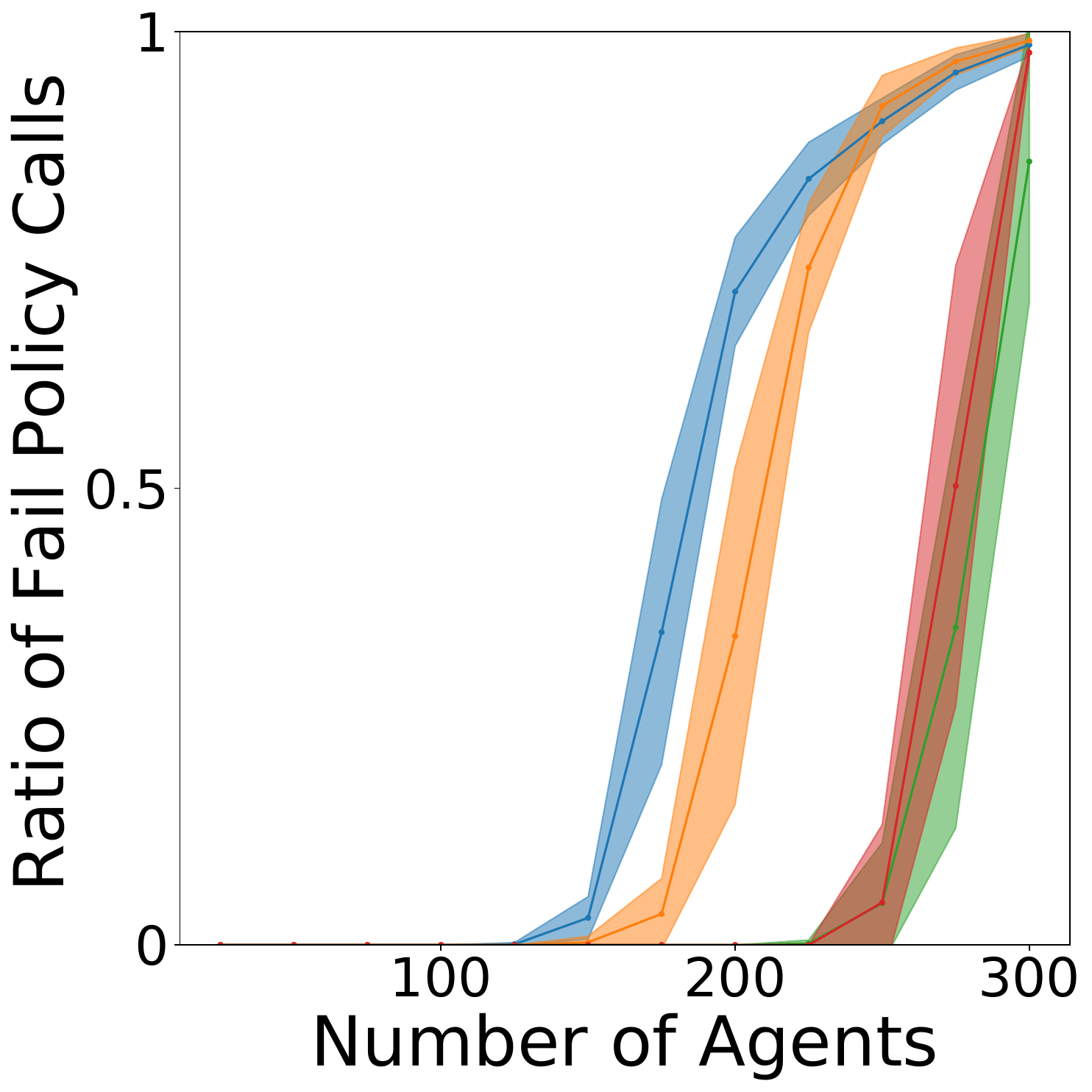}
        \caption{\warehouseSmall}
        \label{fig:plan-invoke-ab:warehouse-33-36}
    \end{subfigure}%
    \hfill
    \begin{subfigure}{0.33\textwidth}
        \centering
        \includegraphics[width=0.5\textwidth]{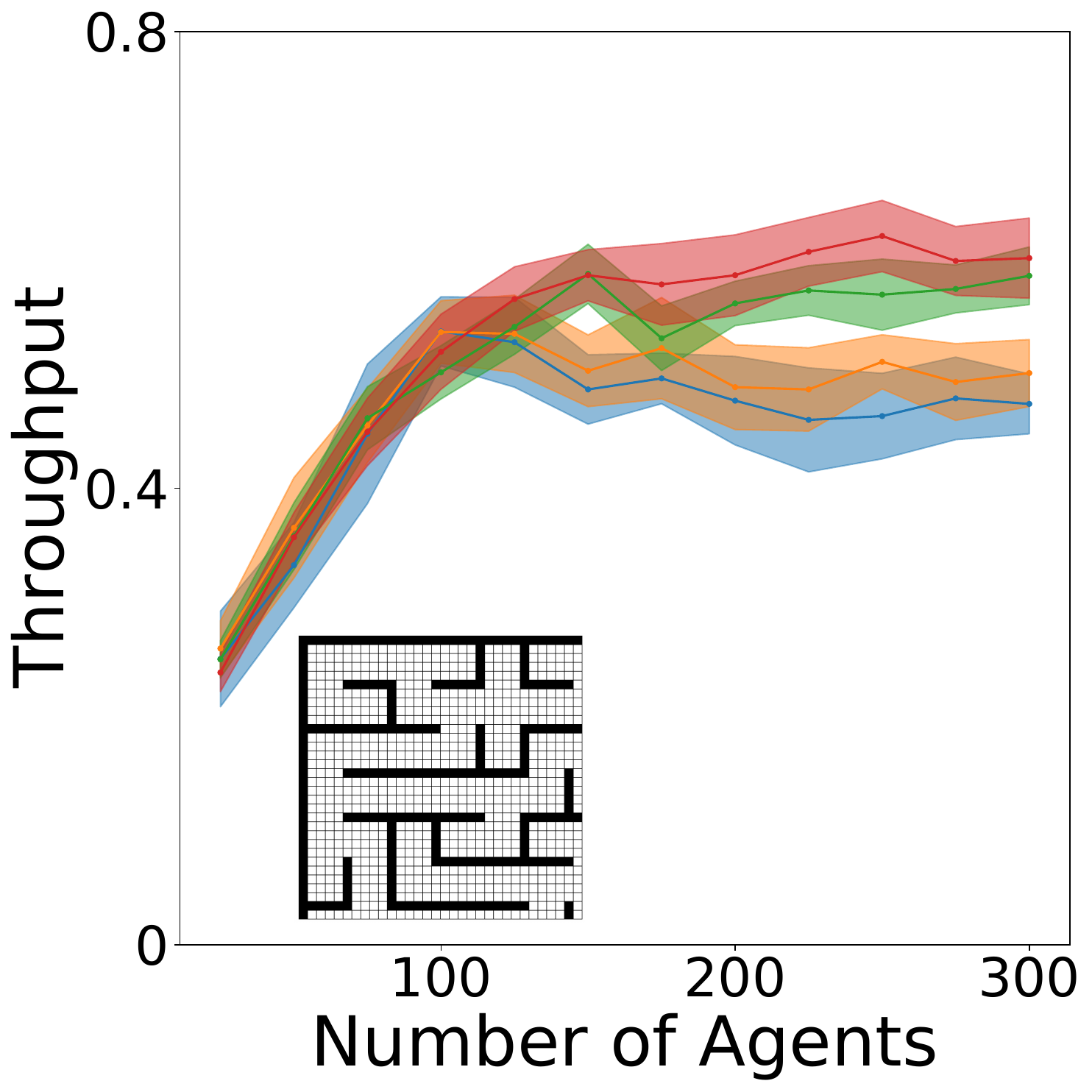}%
        \includegraphics[width=0.5\textwidth]{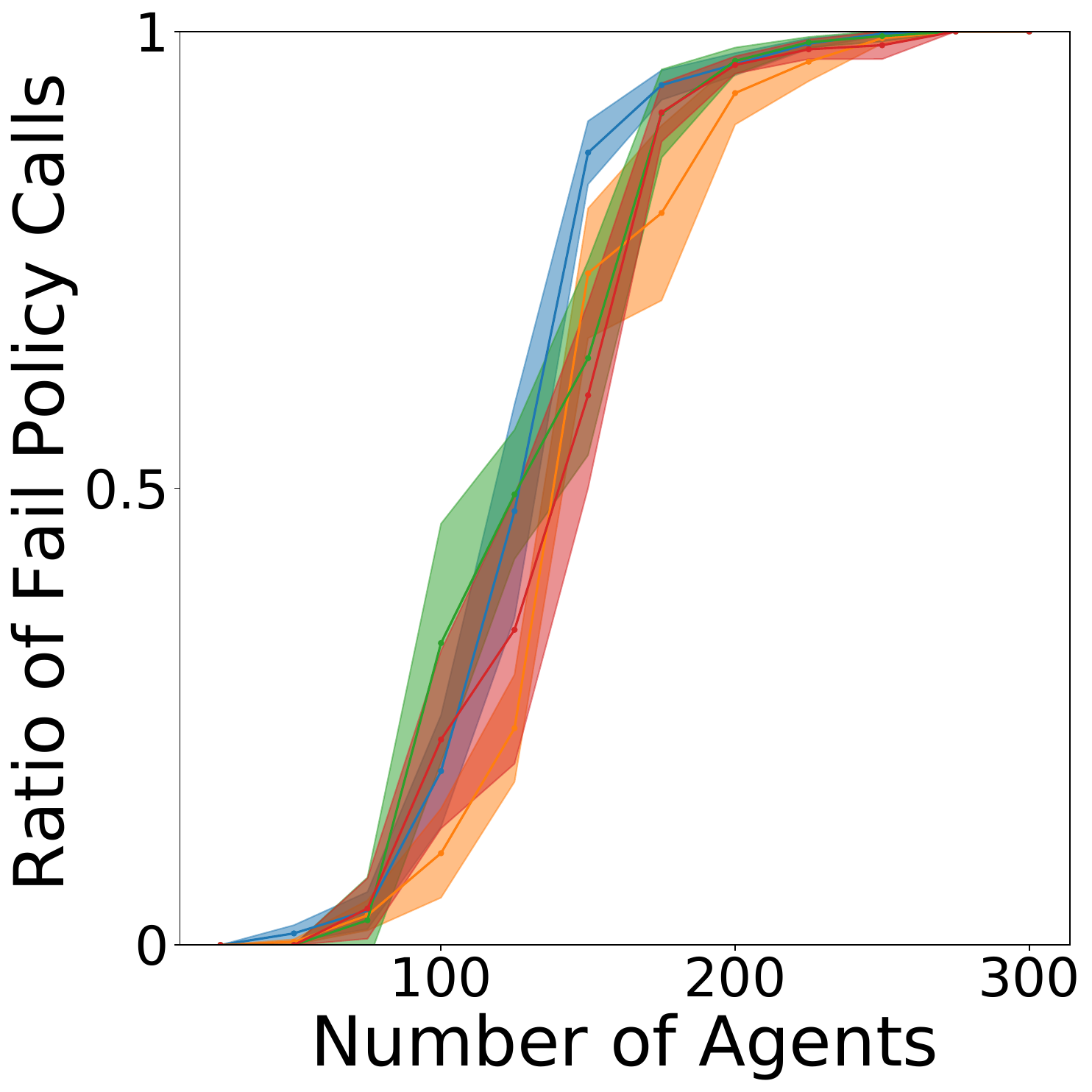}
        \caption{\mazeSmall}
        \label{fig:plan-invoke-ab:maze-32-32-4}
    \end{subfigure}%
    \hfill
    \begin{subfigure}{0.33\textwidth}
        \centering
        \includegraphics[width=0.5\textwidth]{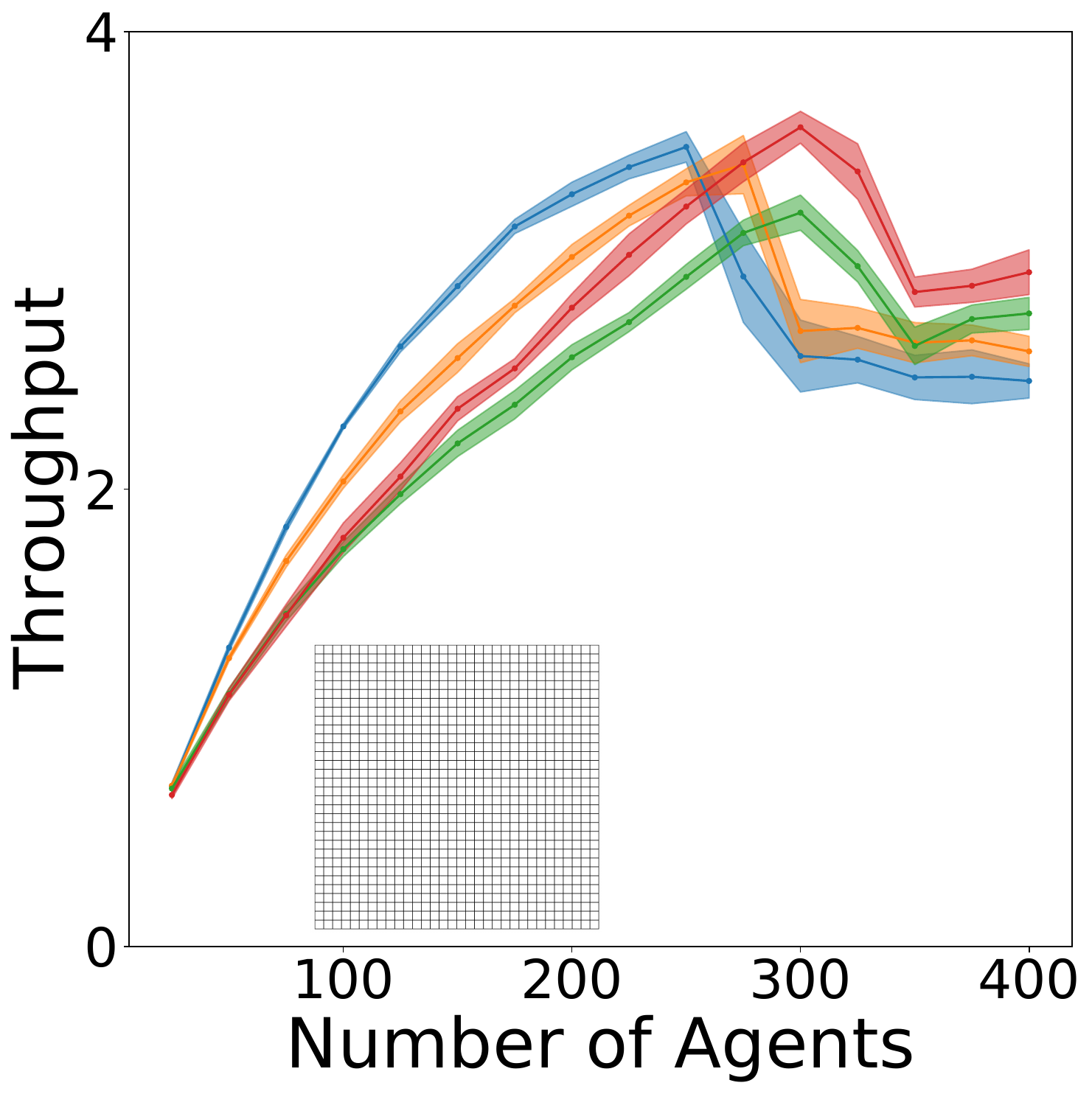}%
        \includegraphics[width=0.5\textwidth]{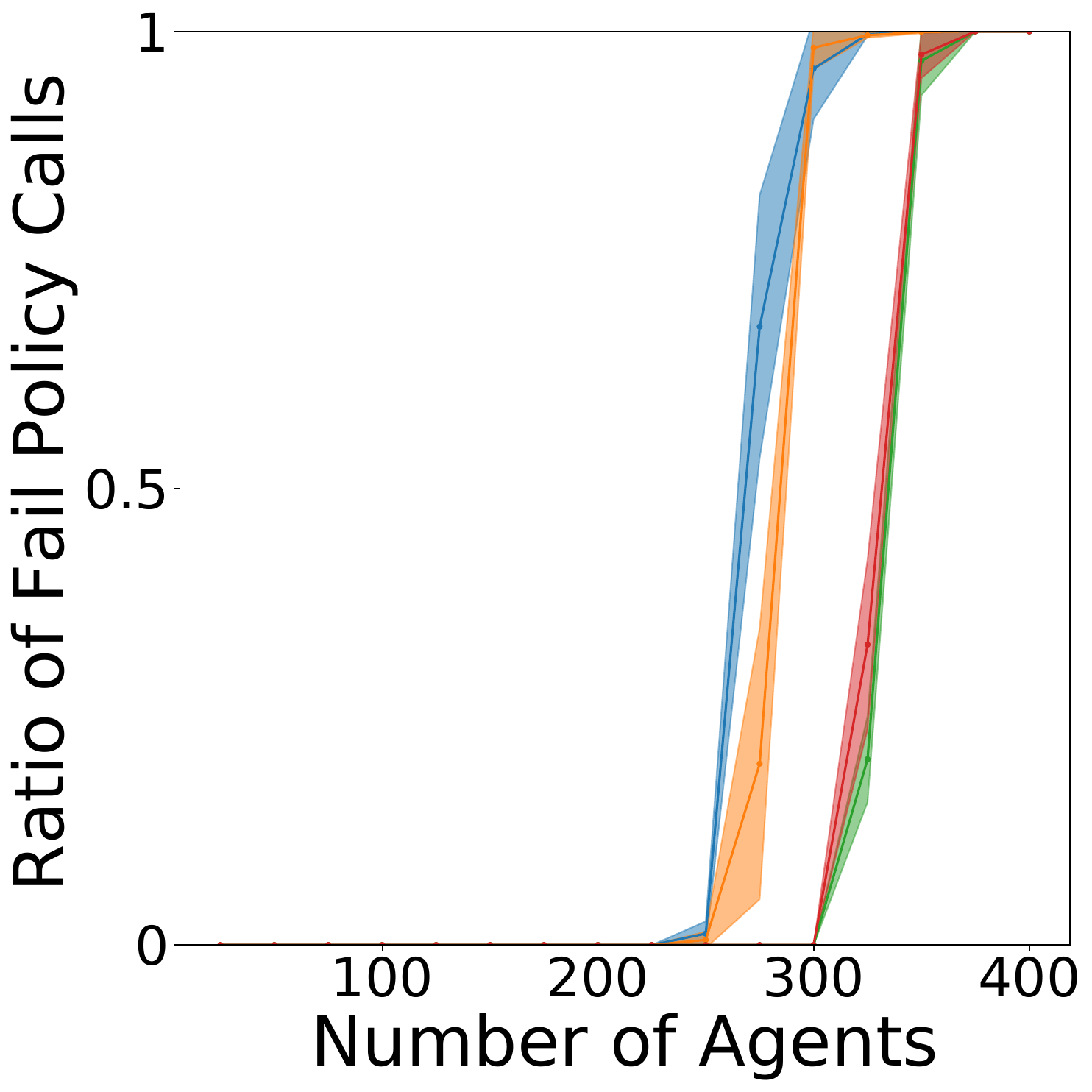}
        \caption{\emptySmall}
        \label{fig:plan-invoke-ab:empty-32-32}
    \end{subfigure}%

    \caption{Experiment results of planner invocation policies (Experiment 2 in \Cref{tab:exp}).}
    \label{fig:plan-invoke-ab-add}
\end{figure*}

\begin{figure*}[!t]
    \centering
    \includegraphics[width=.8\textwidth]{figs/sim_window_ablation/sim-w-ablation-legend.pdf}

    \begin{subfigure}{0.33\textwidth}
        \centering
        \includegraphics[width=0.5\textwidth]{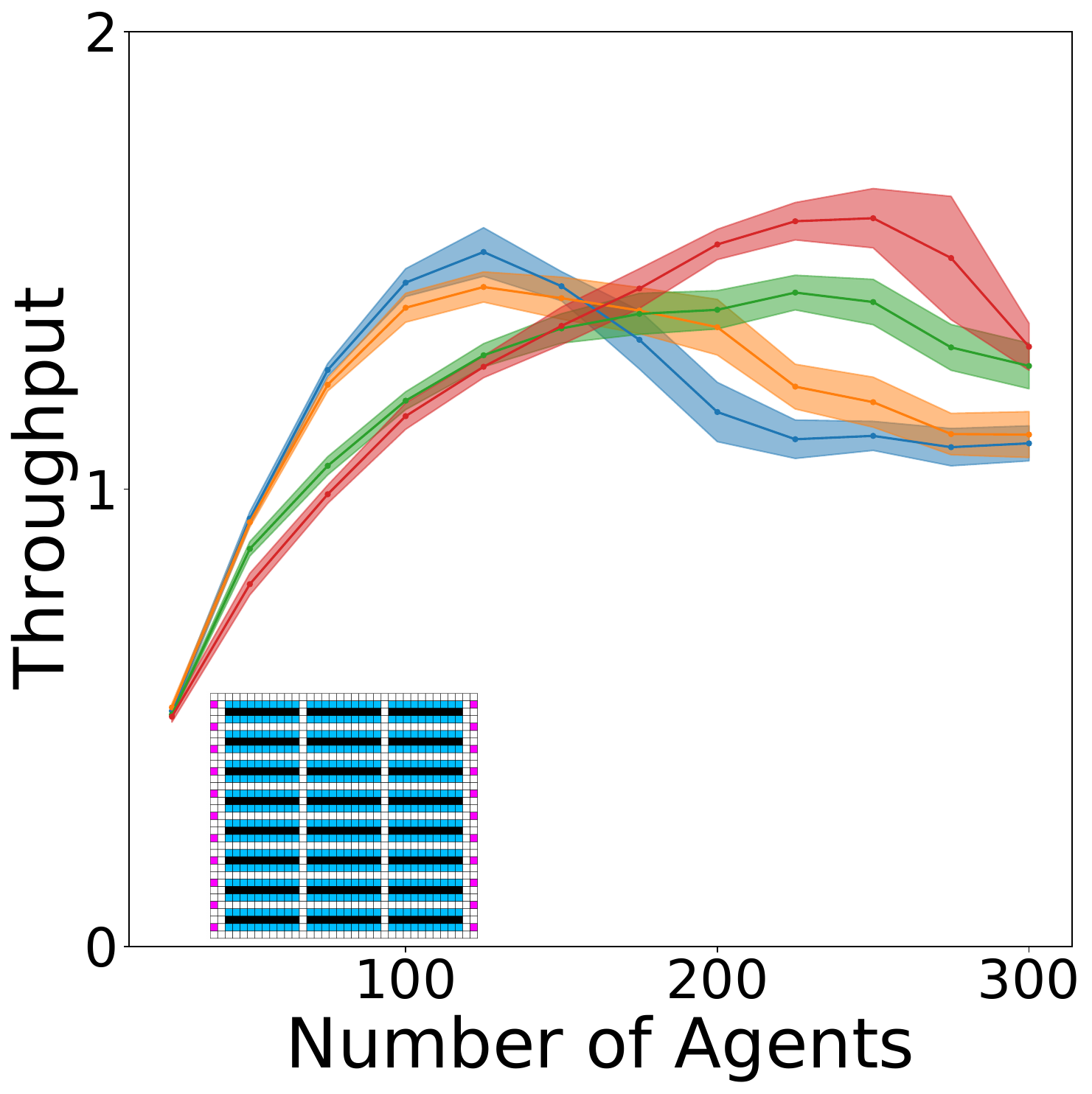}%
        \includegraphics[width=0.5\textwidth]{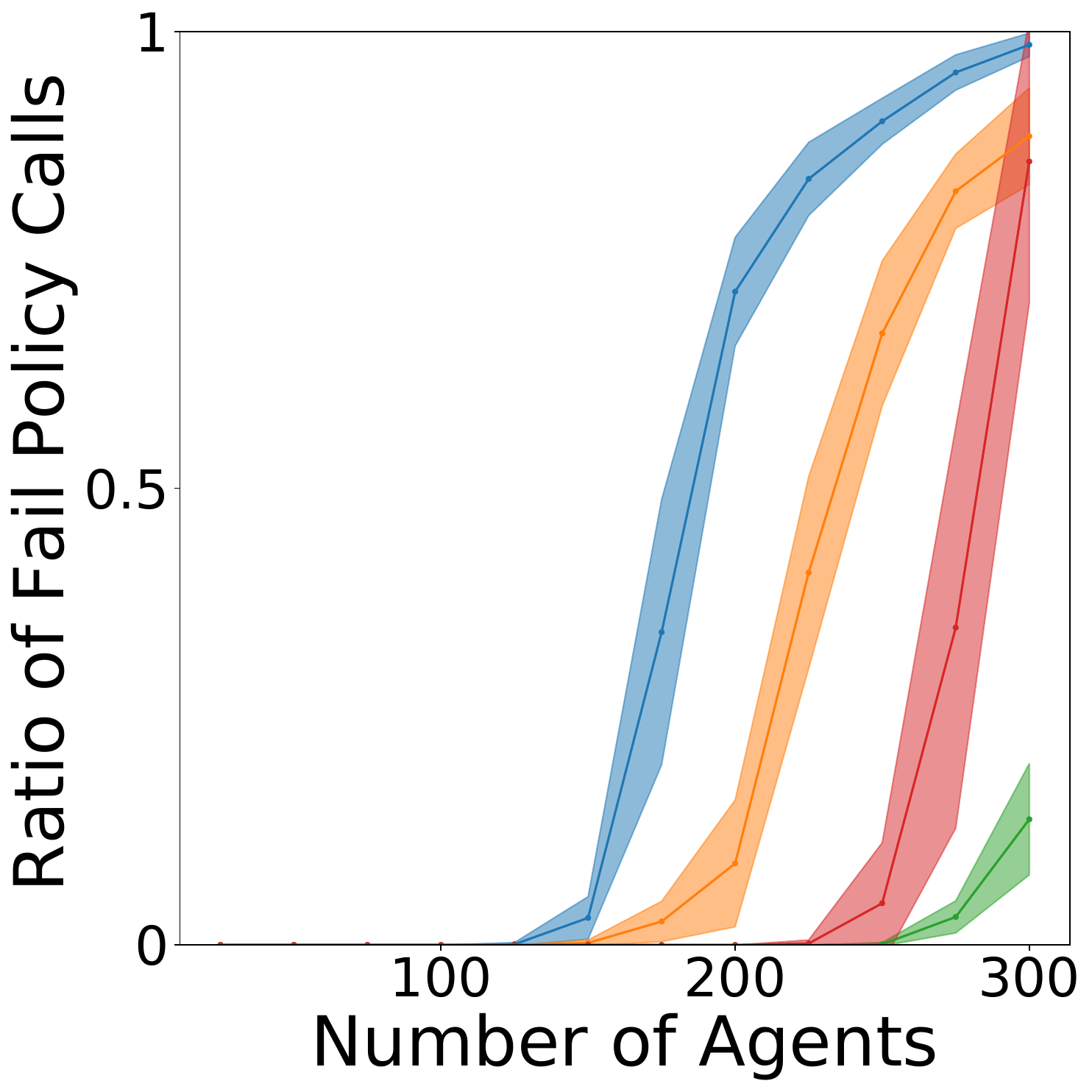}
        \caption{\warehouseSmall}
        \label{fig:sim-w-ab:warehouse-33-36}
    \end{subfigure}%
    \begin{subfigure}{0.33\textwidth}
        \centering
        \includegraphics[width=0.5\textwidth]{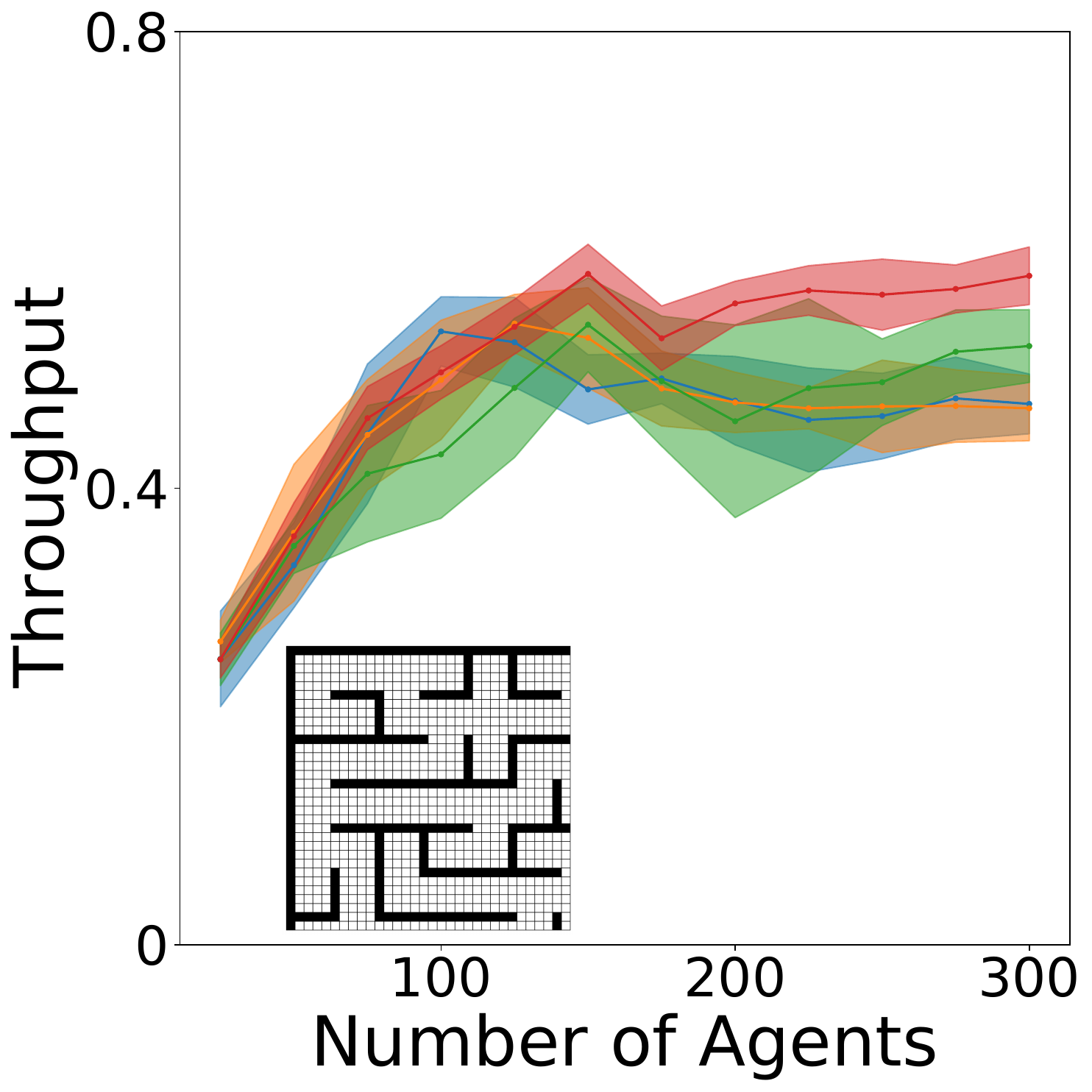}%
        \includegraphics[width=0.5\textwidth]{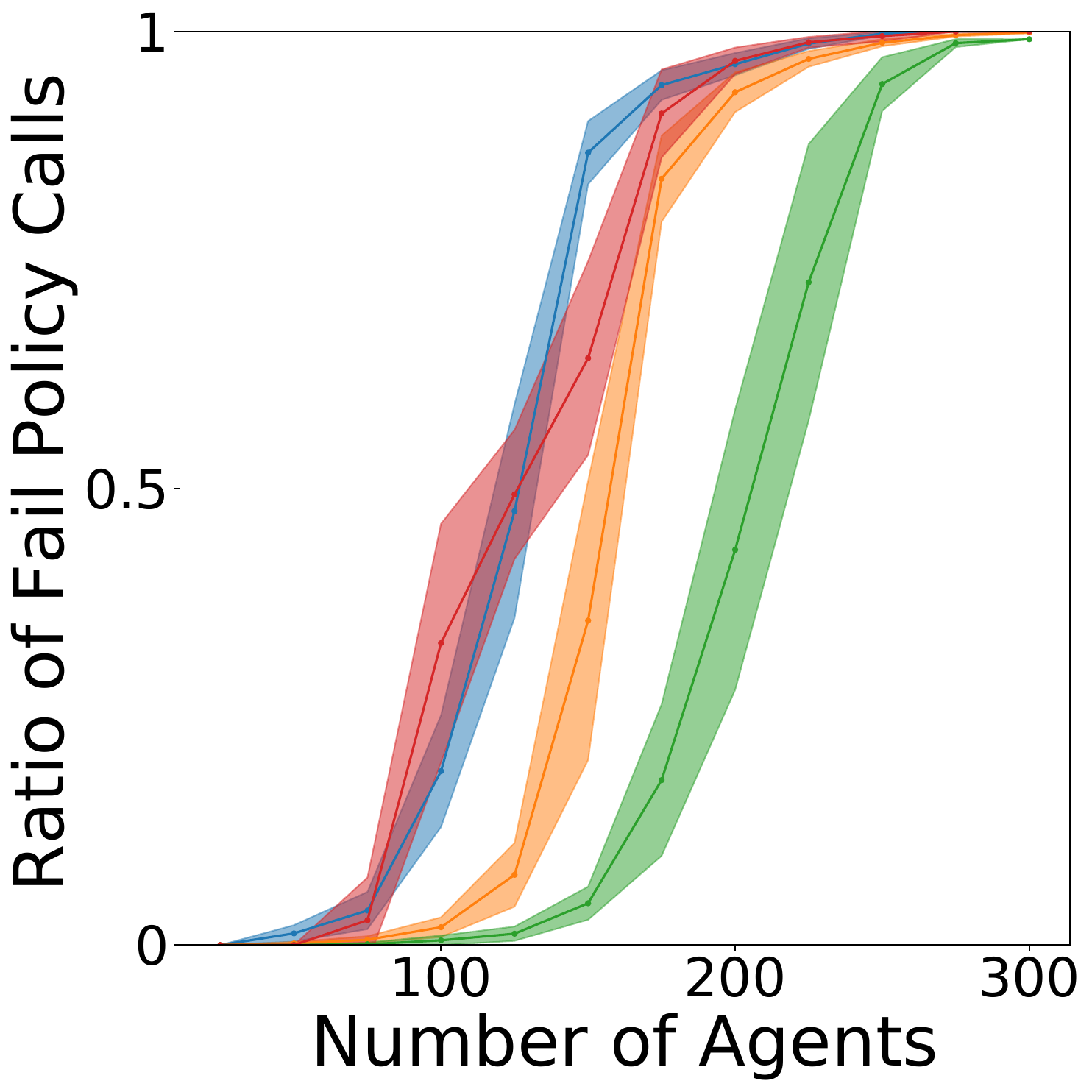}
        \caption{\mazeSmall}
        \label{fig:sim-w-ab:maze-32-32-4}
    \end{subfigure}%
    \hfill
    \begin{subfigure}{0.33\textwidth}
        \centering
        \includegraphics[width=0.5\textwidth]{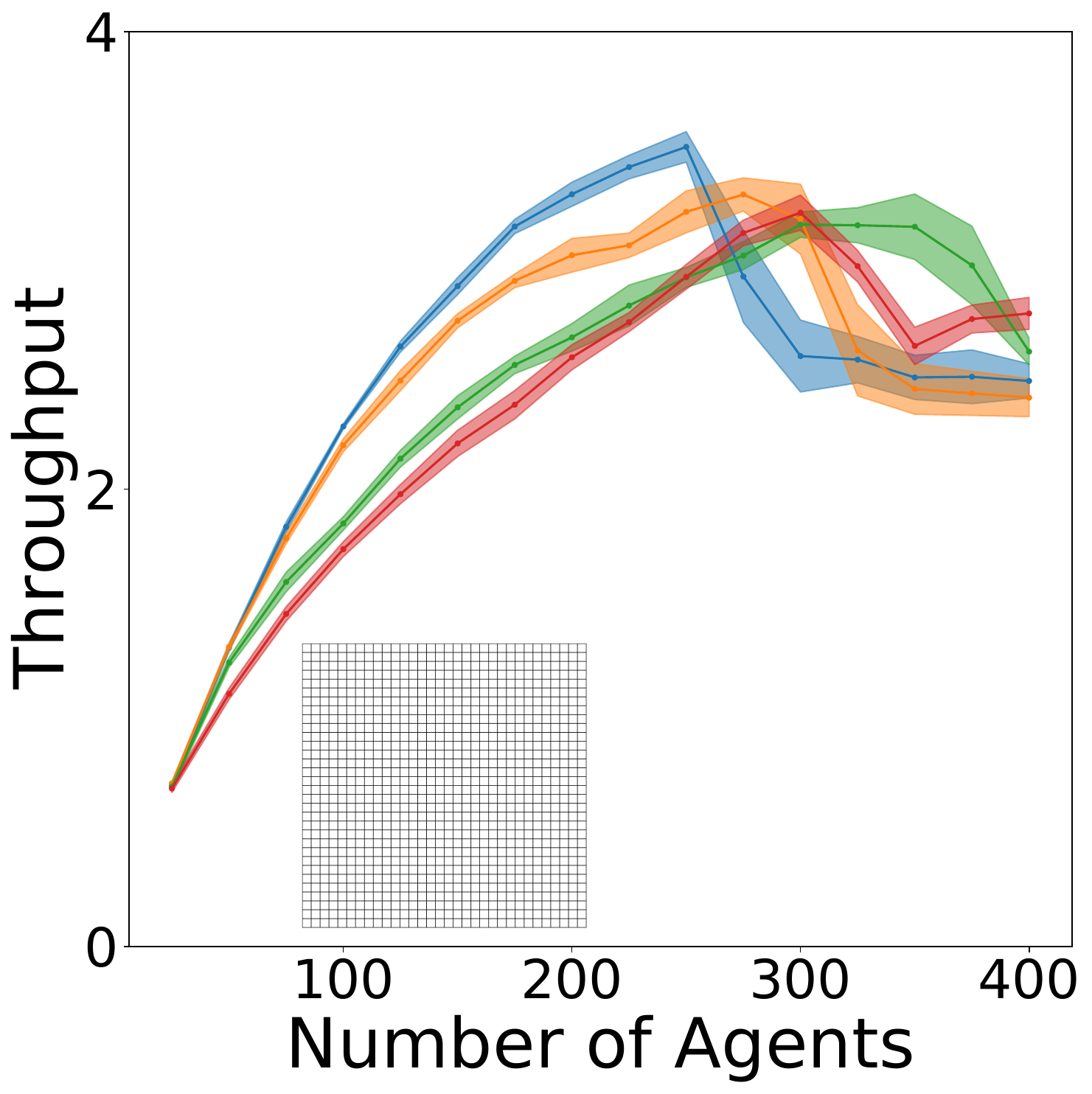}%
        \includegraphics[width=0.5\textwidth]{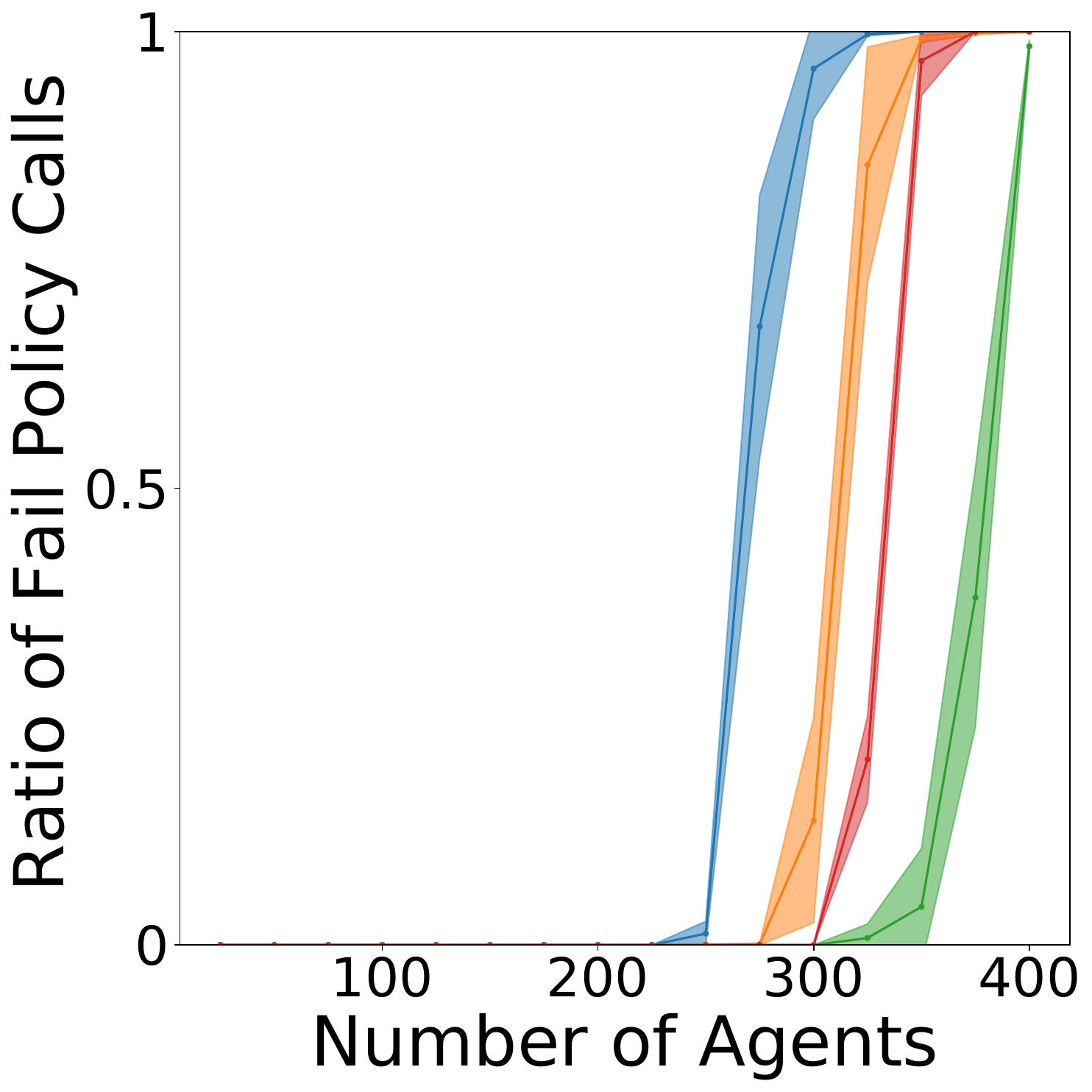}
        \caption{\emptySmall}
        \label{fig:sim-w-ab:empty-32-32}
    \end{subfigure}%

    \caption{Additional Experiment results of $P$ and $W$ (Experiment 3 in \Cref{tab:exp}).}
    \label{fig:sim-w-ab-add}
\end{figure*}

\begin{figure*}[!t]
    \centering
    \includegraphics[width=.4\textwidth]{figs/planner_ablation/replanFreq/legend.pdf}\\

    \begin{subfigure}{0.33\textwidth}
        \centering
        \includegraphics[width=1\textwidth]{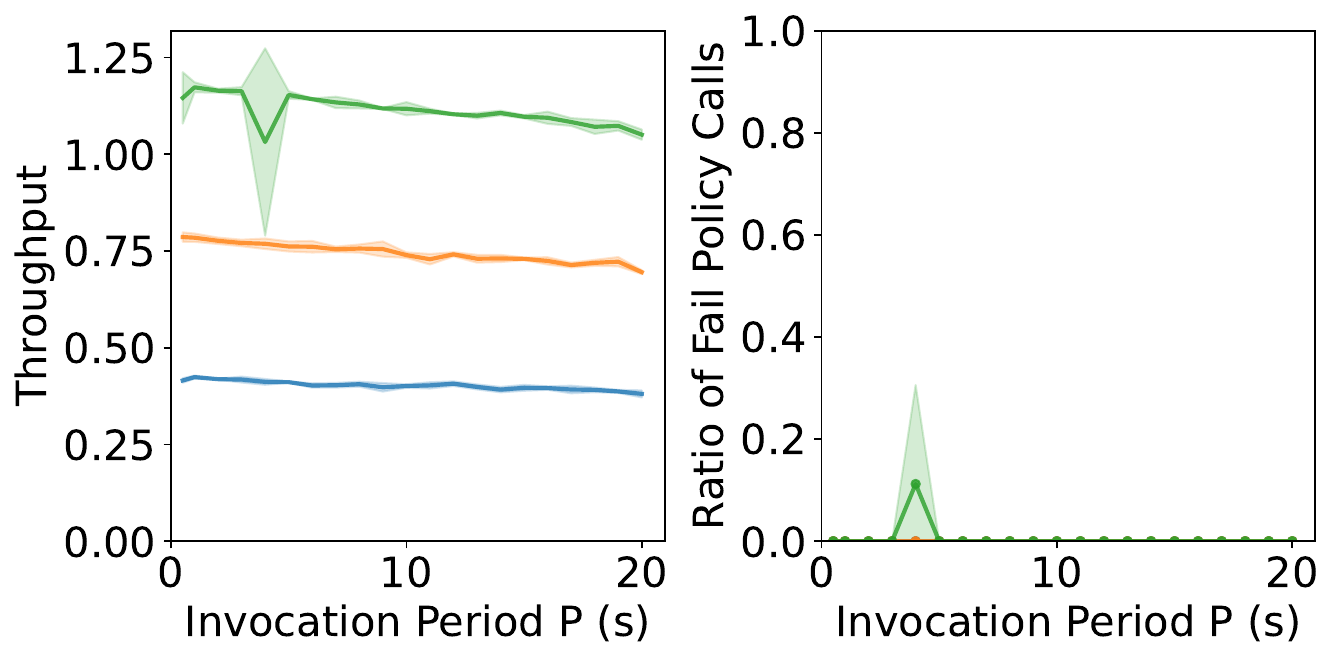}
        \caption{\emptySmall}
        \label{fig:planner-replanFreq-ab:empty-32-32}
    \end{subfigure}%
    \hfill
    \begin{subfigure}{0.33\textwidth}
        \centering
        \includegraphics[width=1\textwidth]{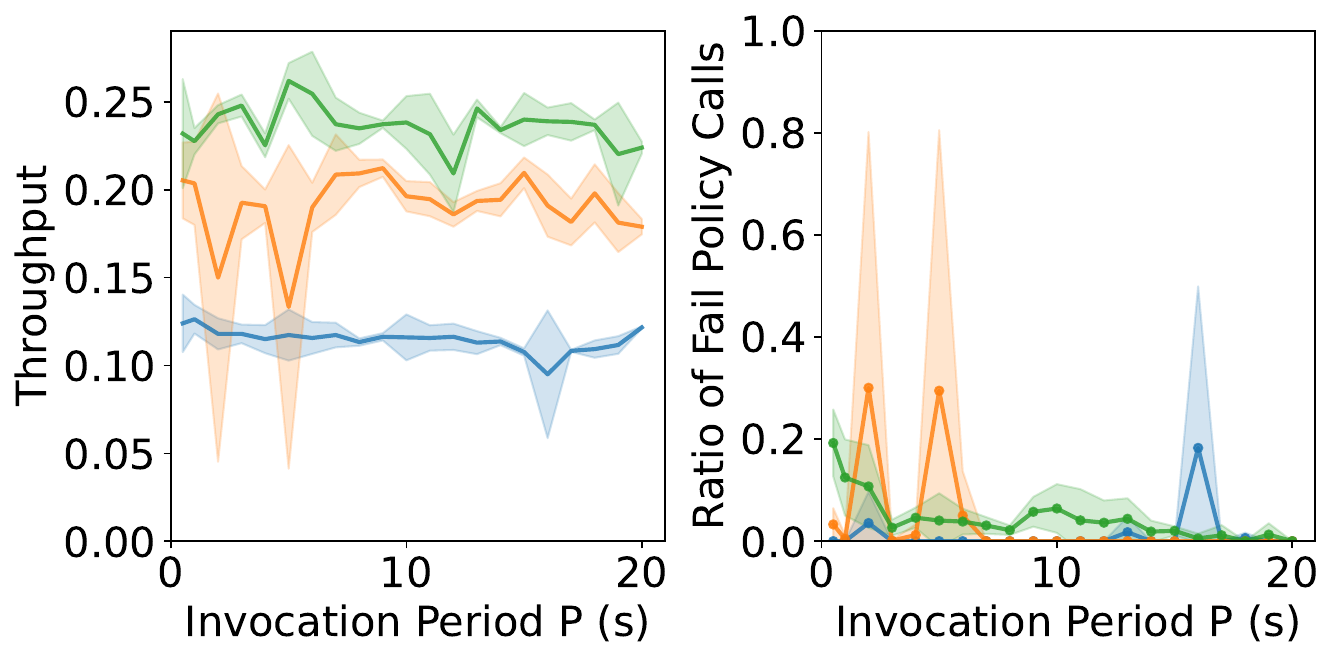}
        \caption{\mazeSmall}
        \label{fig:planner-replanFreq-ab:maze-32-32-4}
    \end{subfigure}
    \hfill
    \begin{subfigure}{0.33\textwidth}
        \centering
        \includegraphics[width=1\textwidth]{figs/planner_ablation/replanFreq/planner_replanFreq_warehouse-10-20-10-2-1.pdf}
        \caption{\warehouseLarge}
        \label{fig:planner-replanFreq-ab:warehouse-10-20-10-2-1}
    \end{subfigure}

    \caption{Additional experiment results of different invocation frequencies (Experiment 4 in \Cref{tab:exp}).}
    \label{fig:planner-replanFreq-ab-add}
\end{figure*}

\Cref{fig:plan-invoke-ab-add} shows additional experimental results of experiment 2 in \Cref{tab:exp}, comparing different invocation policies.
\Cref{fig:sim-w-ab-add} shows additional experimental results of experiment 3 in \Cref{tab:exp}, comparing different planning windows $W$ and invocation policy frequencies $P$ for the periodic invocation policy. \Cref{fig:planner-replanFreq-ab-add} shows additional experimental results of experiment 4 in \Cref{tab:exp}, comparing periodic policy of different $P$ values with standard MAPF planners.

\subsection{Fail policy} \label{appen:exp-add:FP}

\begin{figure*}[!t]
    \centering
    \includegraphics[width=1\textwidth]{figs/fail_policy_ablation/fail_policy_ablation_legend.pdf}

    \begin{subfigure}{0.33\textwidth}
        \centering
        \includegraphics[width=0.5\textwidth]{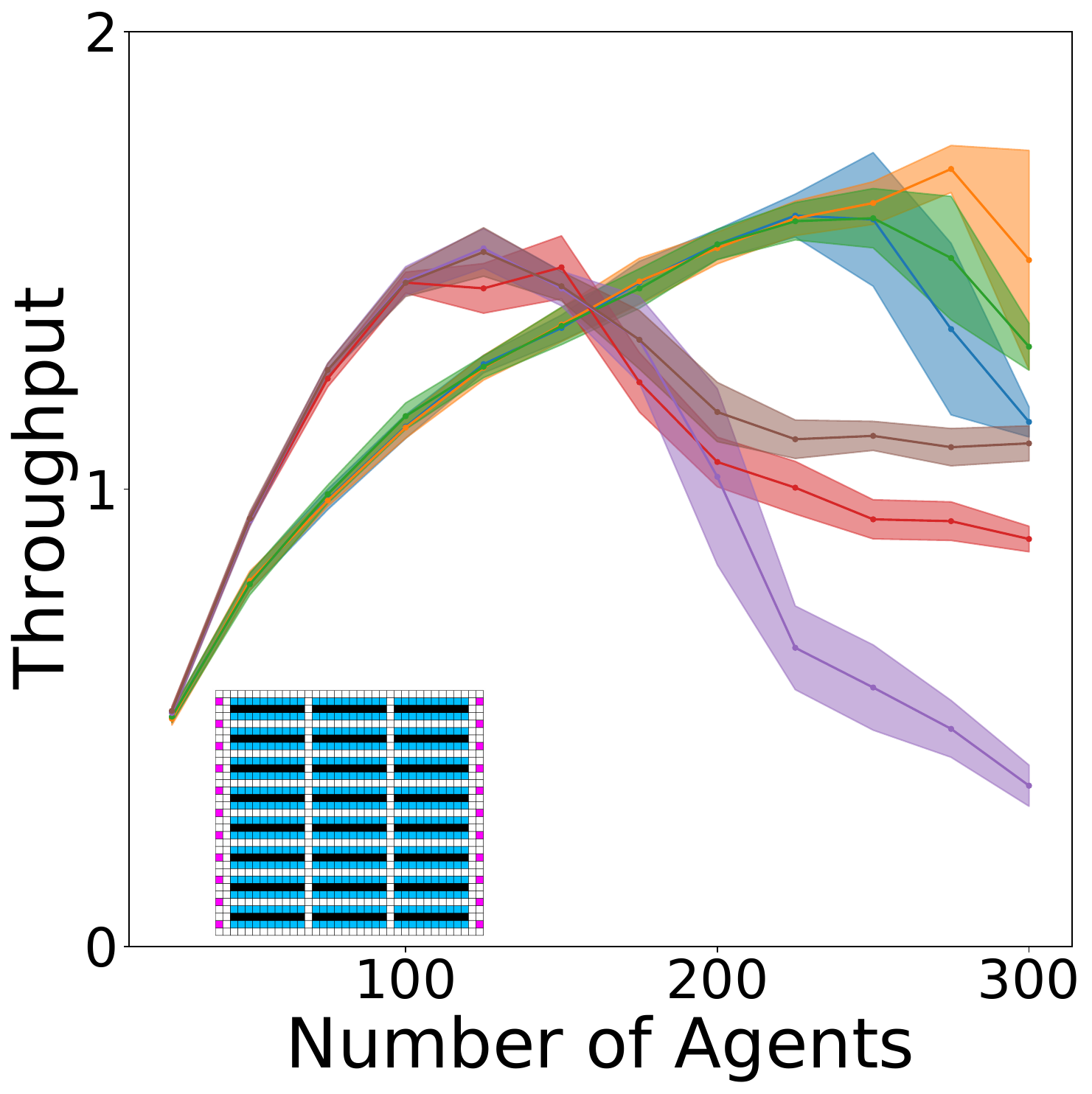}%
        \includegraphics[width=0.5\textwidth]{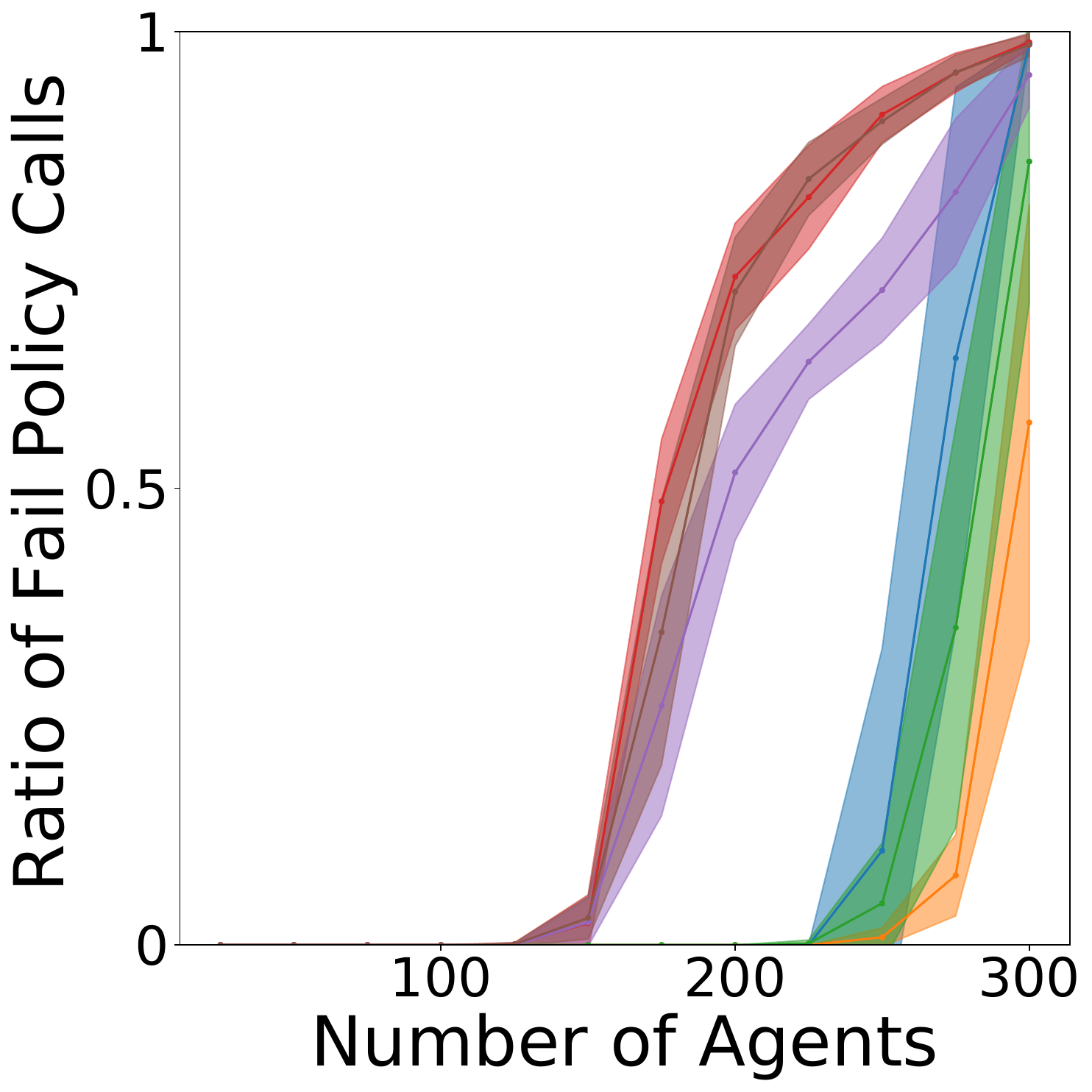}
        \caption{\warehouseSmall}
        \label{fig:fail-policy-ab:warehouse-33-36}
    \end{subfigure}%
    \hfill
    \begin{subfigure}{0.33\textwidth}
        \centering
        \includegraphics[width=0.5\textwidth]{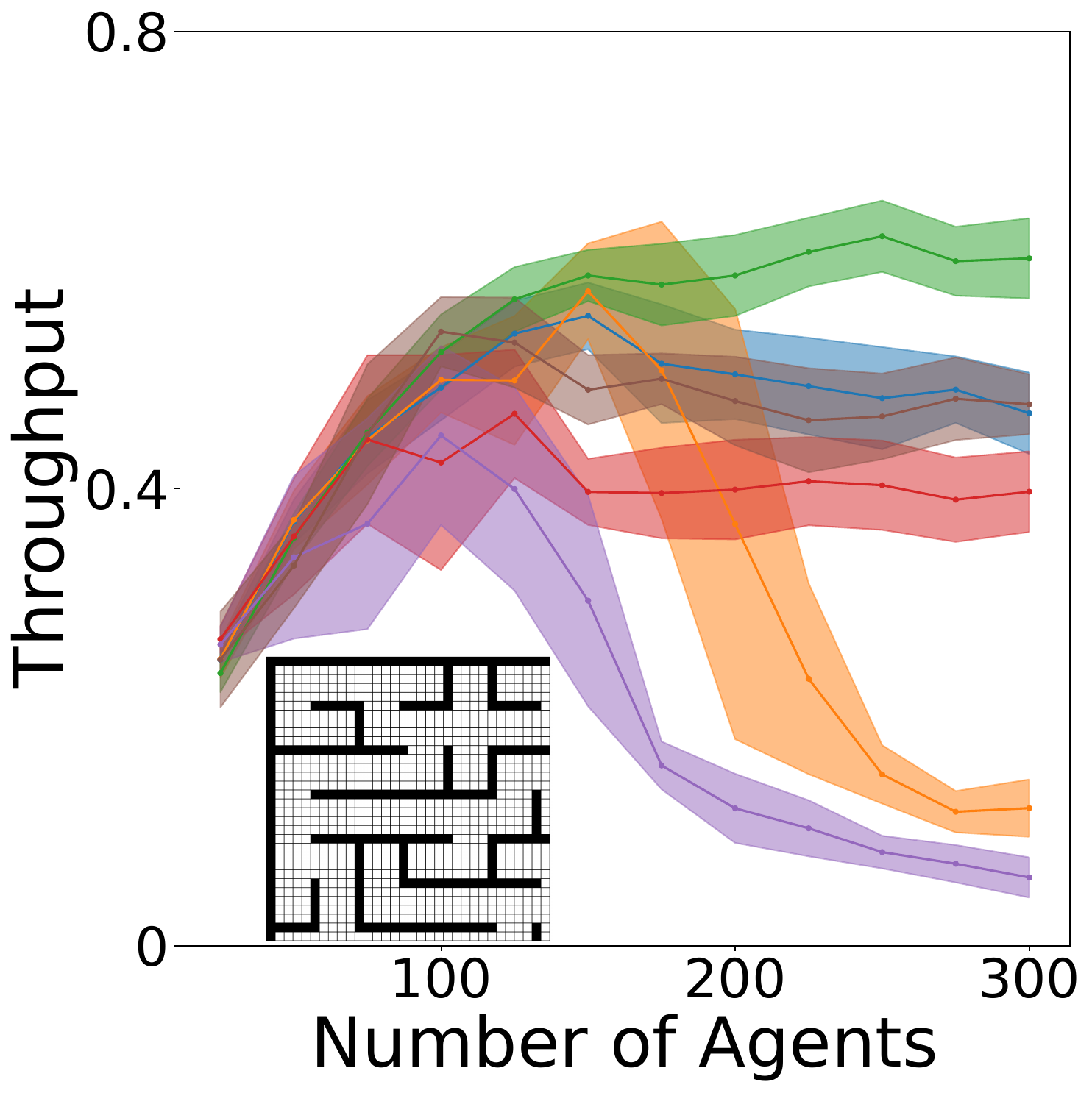}%
        \includegraphics[width=0.5\textwidth]{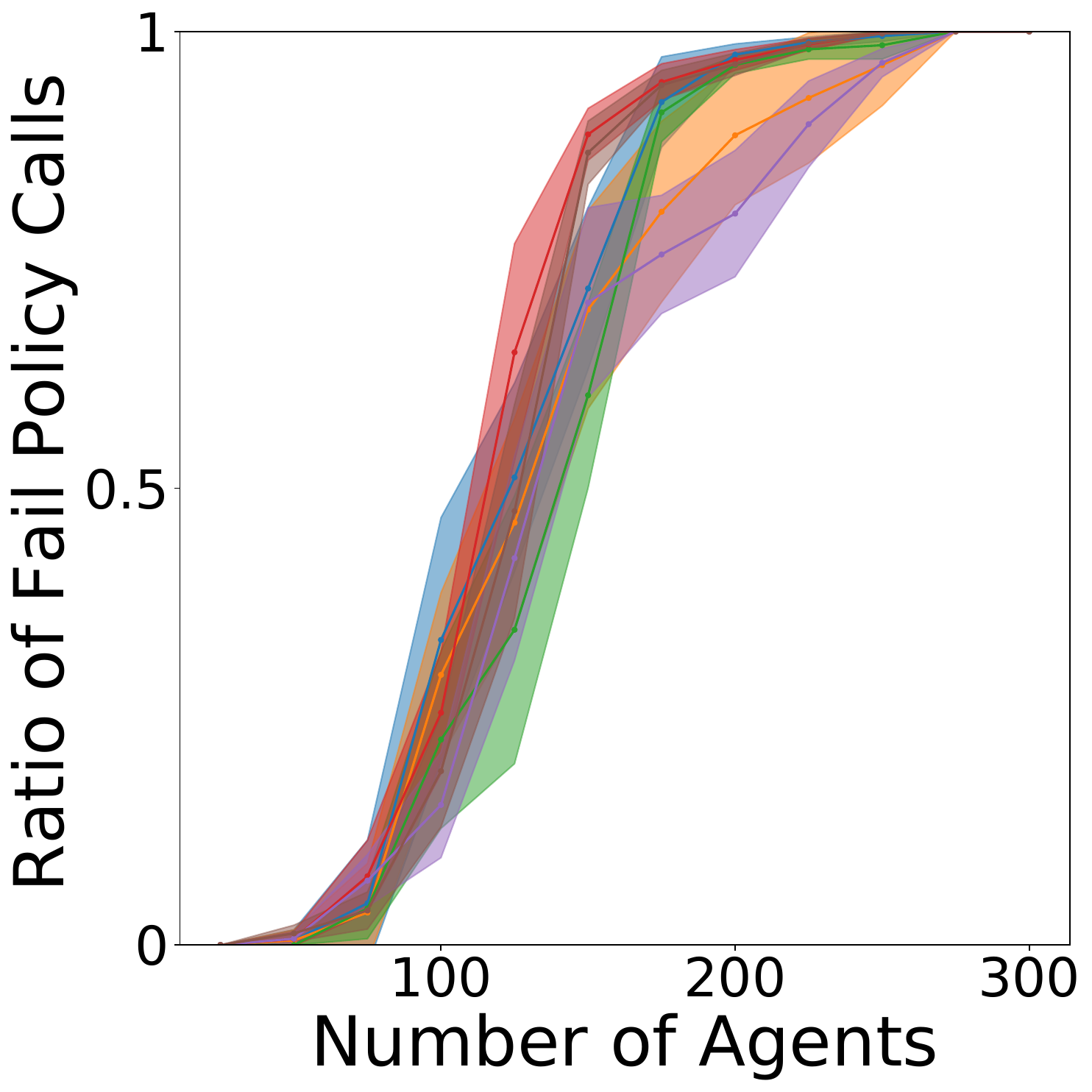}
        \caption{\mazeSmall}
        \label{fig:fail-policy-ab:maze-32-32-4}
    \end{subfigure}%
    \hfill
    \begin{subfigure}{0.33\textwidth}
        \centering
        \includegraphics[width=0.5\textwidth]{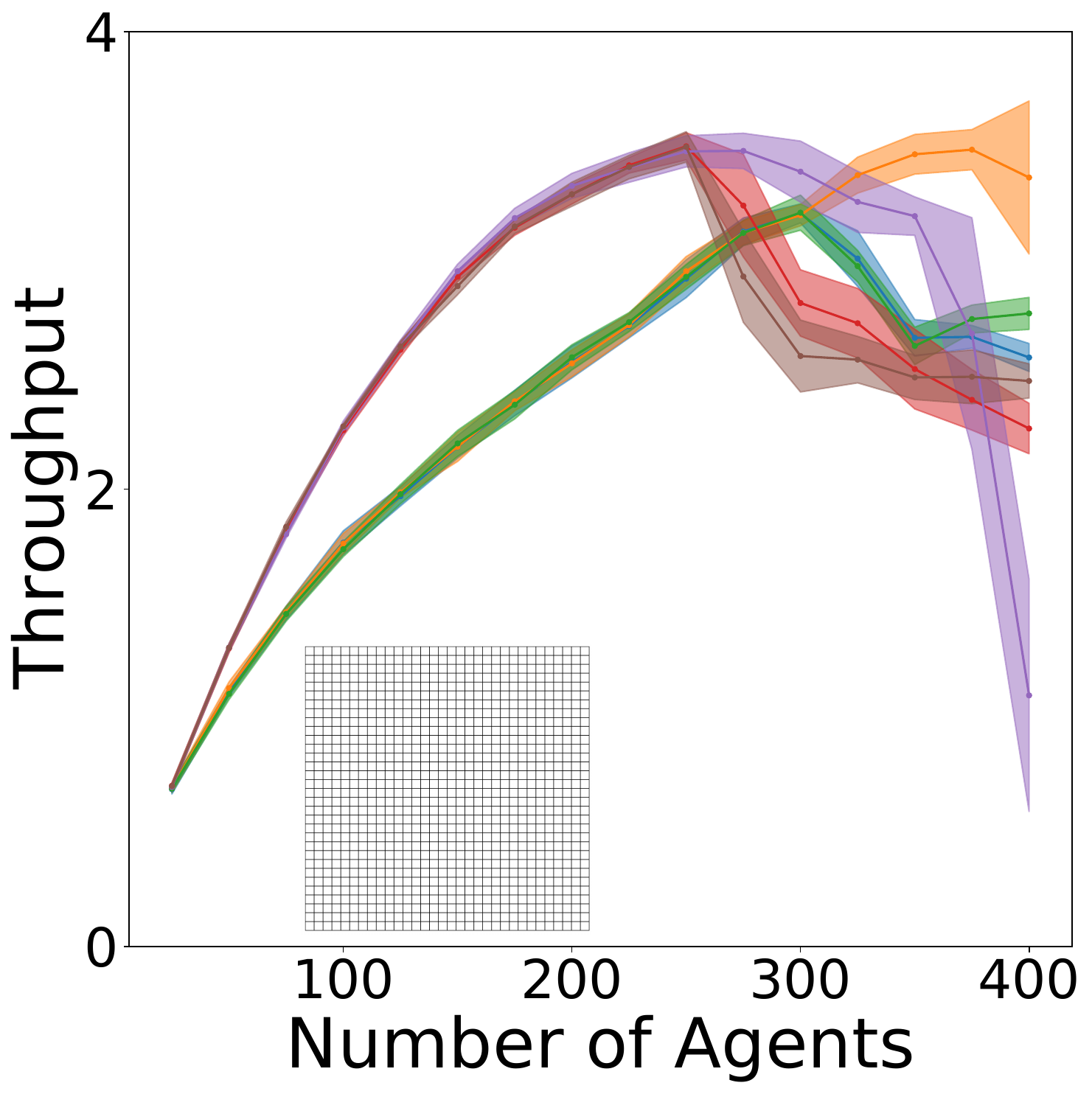}%
        \includegraphics[width=0.5\textwidth]{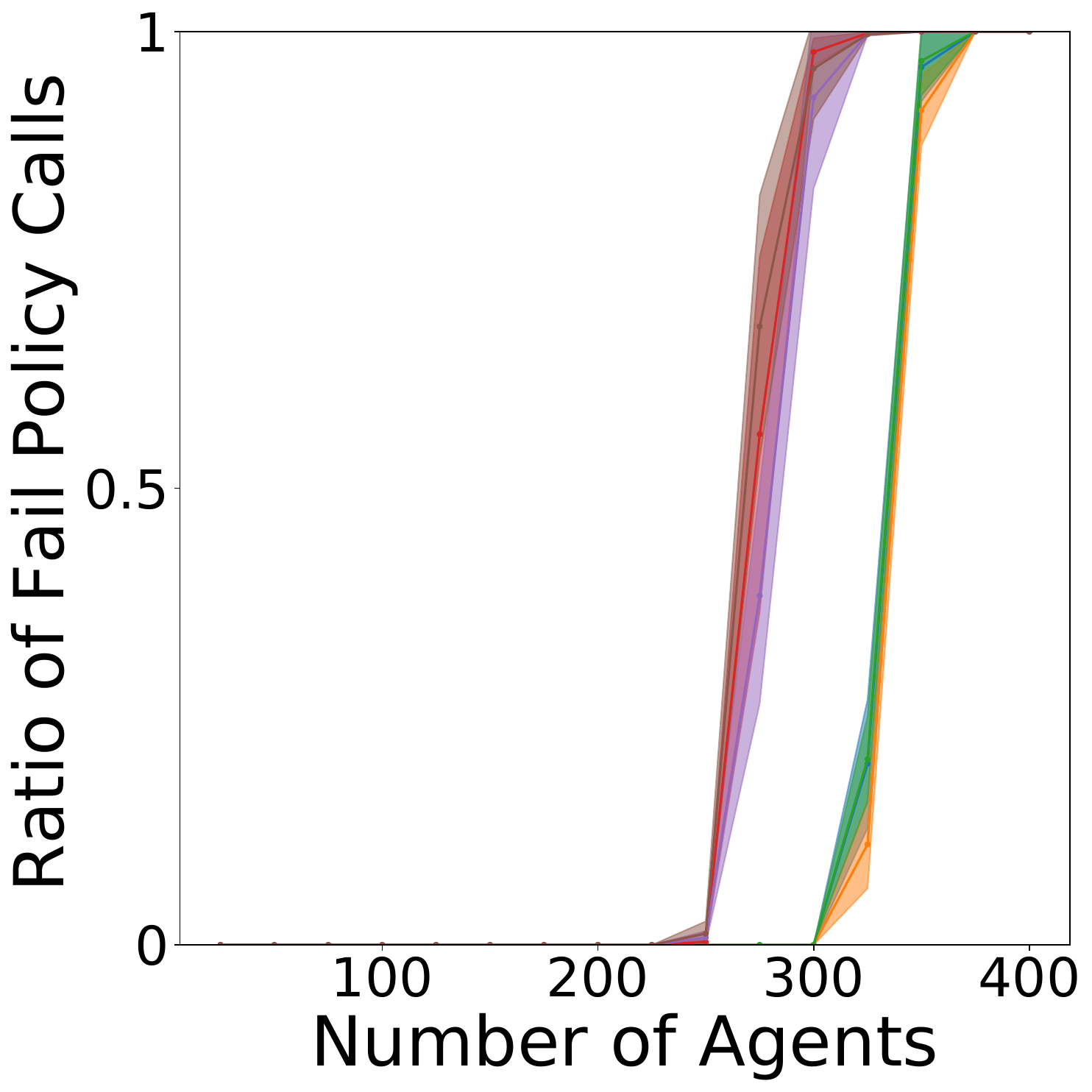}
        \caption{\emptySmall}
        \label{fig:fail-policy-ab:empty-32-32}
    \end{subfigure}%

    \caption{Additional experiment results of fail policies (Experiment 5 in \Cref{tab:exp}).}
    \label{fig:fail-policy-ab-add}
\end{figure*}

\Cref{fig:fail-policy-ab-add} shows additional experimental results of experiment 5 in \Cref{tab:exp}, comparing different fail policies.

\subsection{Optimality and Robot Model Accuracy} \label{appen:exp-add:planner}

\begin{figure*}[!t]
    \centering
    \includegraphics[width=.5\textwidth]{figs/planner_ablation/modelAccOptimality/modelAccOptimality_legend.pdf}

    \begin{subfigure}{0.33\textwidth}
        \centering
        \includegraphics[width=\textwidth]{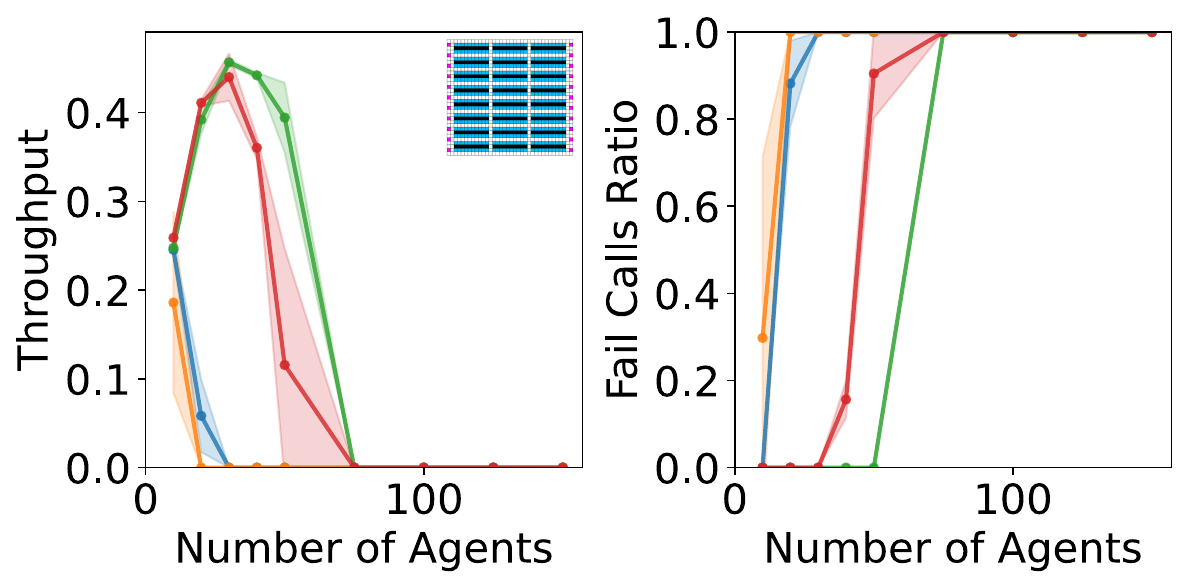}
        \caption{\warehouseSmall}
        \label{fig:planner-modelAccOptimality-ab:kiva}
    \end{subfigure}%
    \hfill
    \begin{subfigure}{0.33\textwidth}
        \centering
        \includegraphics[width=1\textwidth]{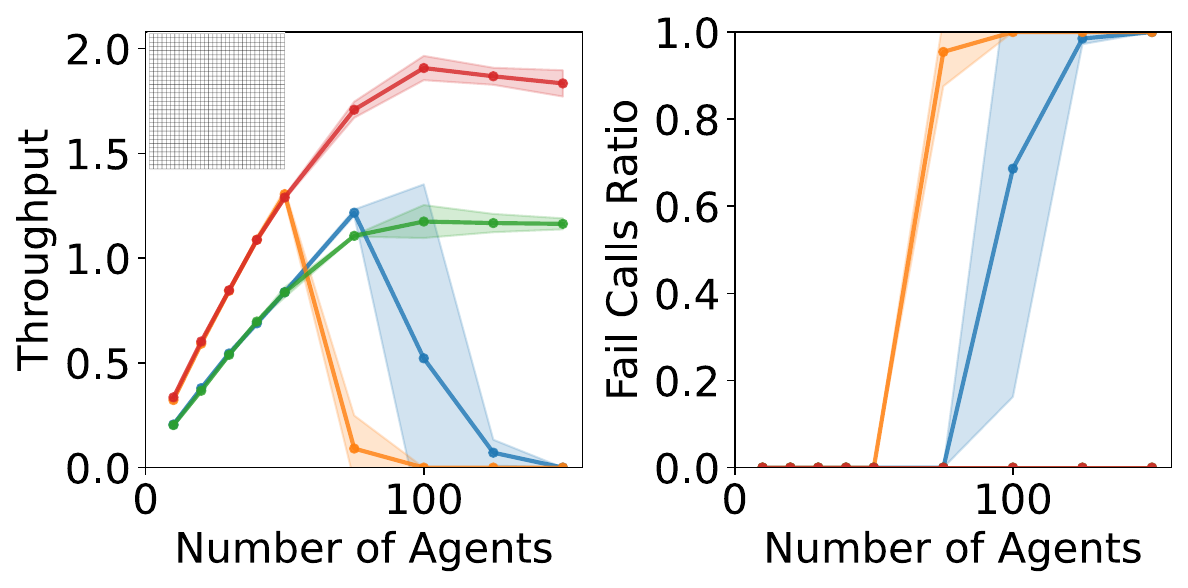}
        \caption{\emptySmall}
        \label{fig:planner-modelAccOptimality-ab:empty-32-32}
    \end{subfigure}%
    \hfill
    \begin{subfigure}{0.33\textwidth}
        \centering
        \includegraphics[width=1\textwidth]{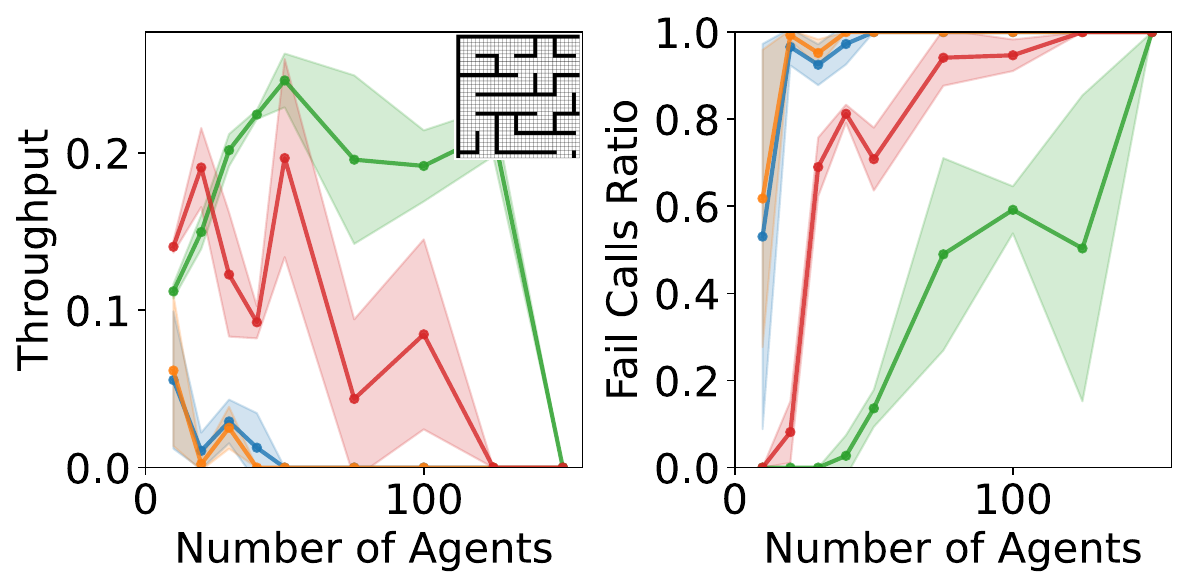}
        \caption{\mazeSmall}
        \label{fig:planner-modelAccOptimality-ab:maze-32-32-4}
    \end{subfigure}


    \caption{Additional experiment results of MAPF model accuracy and planner optimality (Experiment 6 in \Cref{tab:exp}).}
    \label{fig:planner-modelAccOptimality-ab-add}
\end{figure*}

\Cref{fig:planner-modelAccOptimality-ab-add} shows additional experimental results of experiment 6 in \Cref{tab:exp}, comparing planners of different optimalities and agent model accuracies.

\section{Planner Optimality} \label{appen:planner-opt}
In this section, we carry out an additional experiment to compare MAPF planners of different optimality.
\subsubsection{Experiment Setup}
In this section, given the trade-off between plan optimality and runtime of MAPF algorithms, we evaluate the impact of different planner optimality on throughput.
We plan paths of different optimality using MAPF-LNS~\cite{li2021anytime}, an anytime MAPF algorithm that progressively improves solution quality given a runtime limit seconds. By increasing the runtime limit $T$, MAPF-LNS returns better solutions.
We use the standard MAPF model with the \distinctOneGoal refiner and periodic invocation policy with $P=T$, comparing the runtime limits of $T \in \{0.1, 0.5, 1, 2, 3, ..., 20\}$ seconds.

\subsubsection{Experiment Result}
\begin{figure*}[!t]
    \centering

    \begin{subfigure}{0.33\textwidth}
        \centering
        \includegraphics[width=1\textwidth]{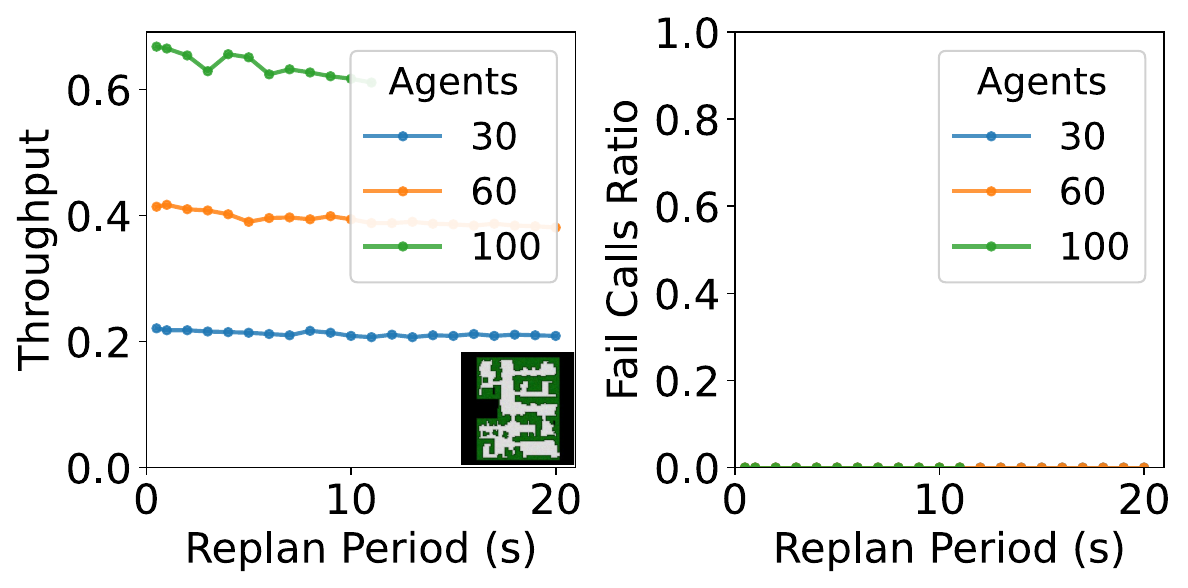}
        \caption{\denMid}
        \label{fig:planner-optimality-ab:den312d}
    \end{subfigure}%
    \hfill
    \begin{subfigure}{0.33\textwidth}
        \centering
        \includegraphics[width=1\textwidth]{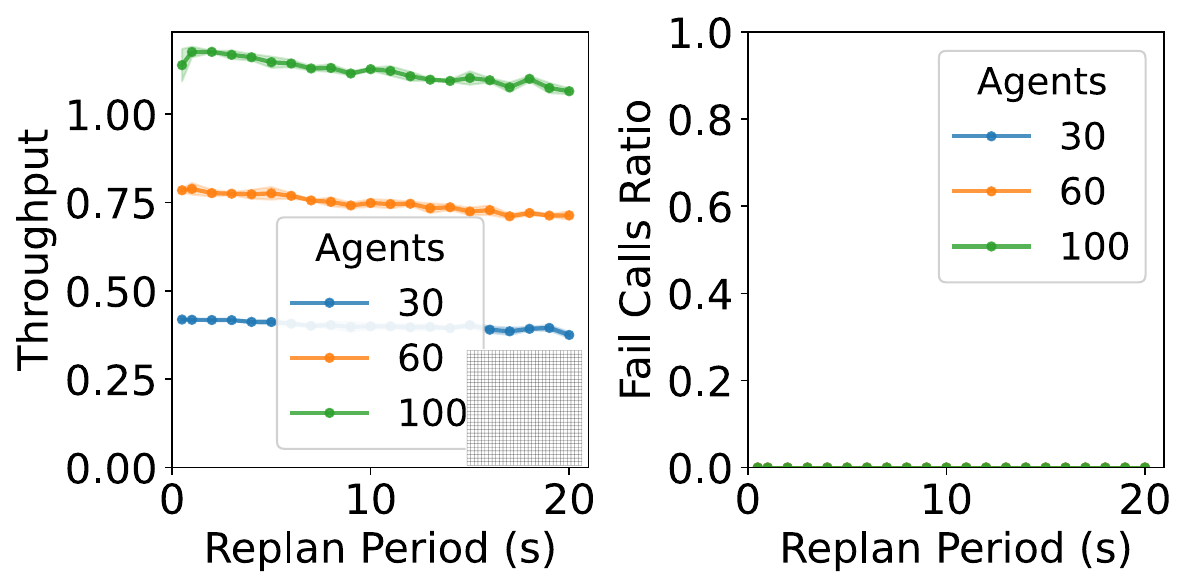}
        \caption{\emptySmall}
        \label{fig:planner-optimality-ab:empty-32-32}
    \end{subfigure}%
    \hfill
    \begin{subfigure}{0.33\textwidth}
        \centering
        \includegraphics[width=1\textwidth]{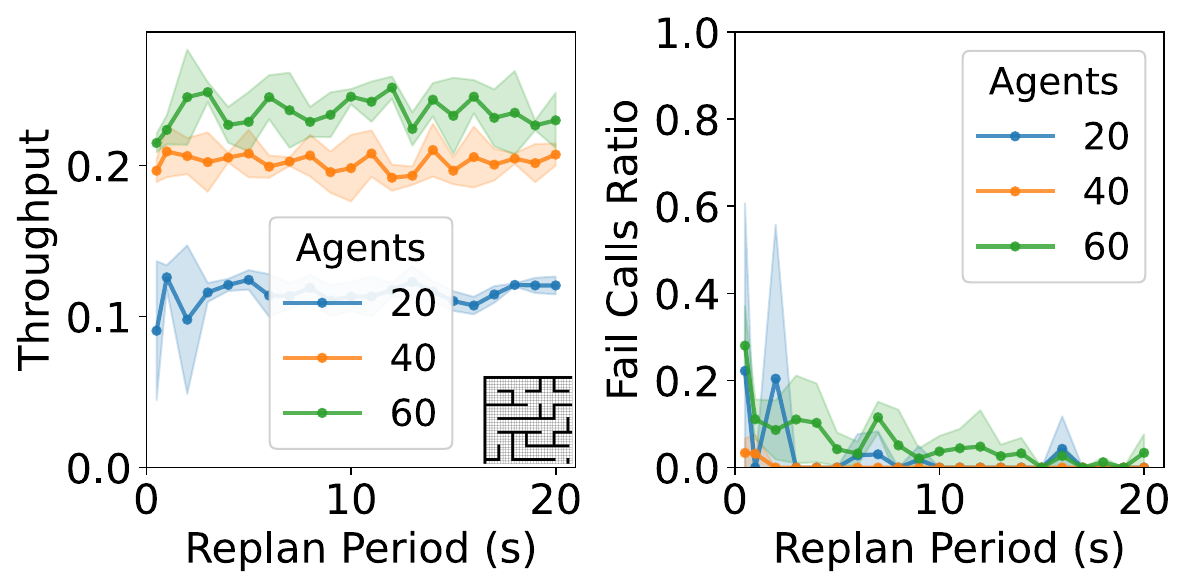}
        \caption{\mazeSmall}
        \label{fig:planner-optimality-ab:maze-32-32-4}
    \end{subfigure}

    \vspace{1em}
    \begin{subfigure}{0.33\textwidth}
        \centering
        \includegraphics[width=1\textwidth]{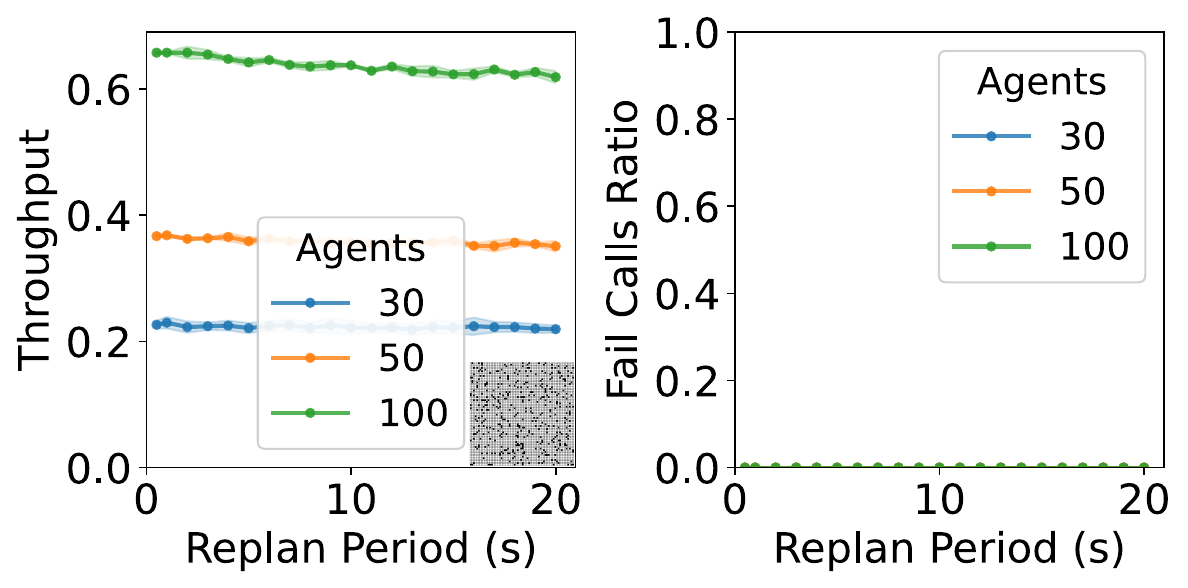}
        \caption{\randomMid}
        \label{fig:planner-optimality-ab:random-64-64-10}
    \end{subfigure}%
    \hfill
    \begin{subfigure}{0.33\textwidth}
        \centering
        \includegraphics[width=1\textwidth]{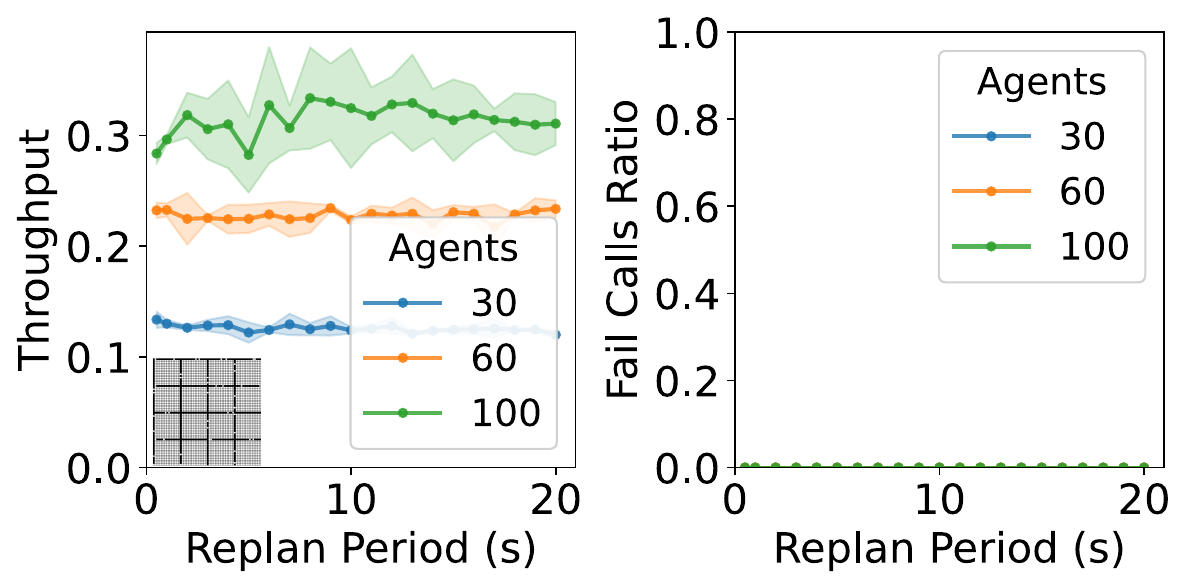}
        \caption{\roomMid}
        \label{fig:planner-optimality-ab:room-64-64-16}
    \end{subfigure}%
    \hfill
    \begin{subfigure}{0.33\textwidth}
        \centering
        \includegraphics[width=1\textwidth]{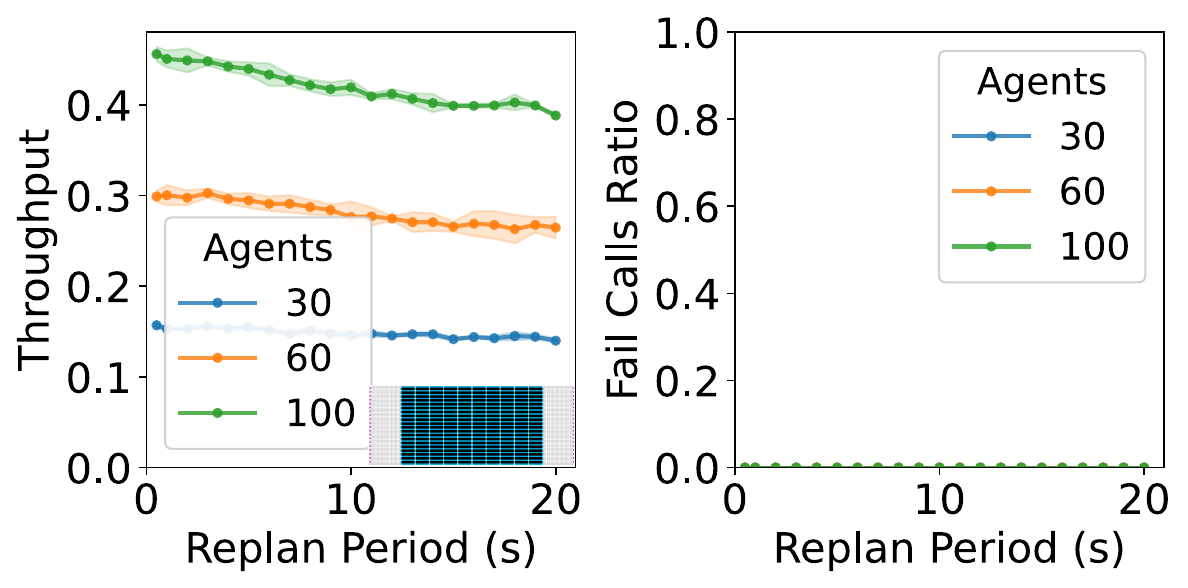}
        \caption{\warehouseLarge}
        \label{fig:planner-optimality-ab:warehouse-10-20-10-2-1}
    \end{subfigure}

    \caption{Experiment results of MAPF planner optimality.}
    \label{fig:planner-optimality-ab}
\end{figure*}

As shown in~\Cref{fig:planner-optimality-ab}, the throughput curves remain relatively flat across environments and robot densities, indicating that more frequent replanning does not necessarily yield better execution performance.
This result arises from two competing effects:
Higher replanning frequency should help mitigate execution asynchrony caused by delays and controller variability.
However, because we use an anytime planner (MAPF-LNS2), a longer replanning period enables more computation, producing higher-quality plans with fewer collisions and thus higher throughput.
As these effects counteract each other, the net influence of replanning frequency becomes small.

\end{document}